\documentclass[journal]{IEEEtran}   % Comment this line out if you need a4paper

\IEEEoverridecommandlockouts        % This command is only needed if you want to use the \thanks command
\usepackage[pdftex]{graphicx} % for pdf, bitmapped graphics files
\usepackage{mathptmx} % assumes new font selection scheme installed
\usepackage{times}    % assumes new font selection scheme installed
\usepackage{amsmath}  % assumes amsmath packzage installed
\usepackage{amssymb}  % assumes amsmath package installed
\usepackage{url}
\usepackage{cite}
\usepackage{tikz}
\usepackage{xargs}
\usepackage{ifthen}
\usepackage{dsfont}
\usepackage{lipsum}
\usepackage{calrsfs}
\usepackage{hyperref}
\usepackage{tabularx}
\usepackage{amsfonts}
\usepackage{booktabs}
\usepackage{epstopdf}
\usepackage{mathtools}
\usepackage{subcaption}
\usepackage{dblfloatfix}
\usepackage[T1]{fontenc}
\usepackage[stable]{footmisc}
\usepackage[framemethod=tikz]{mdframed}

\renewcommand{\baselinestretch}{1.00}

\definecolor{lightblue}{rgb}{0.145,0.6666,1.0000} % color for video sequences
\definecolor{modcolor}{rgb}{0.2745,0.4156,0.6705} % color for manuscript updates

\newcommand{\filename}[1]{#1.pdf}

\DeclareMathAlphabet{\pazocal}{OMS}{zplm}{m}{n}

\DeclareFontFamily{OT1}{pzc}{}
\DeclareFontShape{OT1}{pzc}{m}{it}{<-> s * [1.10] pzcmi7t}{}
\DeclareMathAlphabet{\mathpzc}{OT1}{pzc}{m}{it}

\newcommand{\ctxt}[1]{{\color{black} #1}}

\newcommand{ \iid        }{ \mathrm{i}            } % index i
\newcommand{ \jid        }{ \mathrm{j}            } % index j
\newcommand{ \xid        }{ \mathrm{x}            } % index x
\newcommand{ \yid        }{ \mathrm{y}            } % index y
\newcommand{ \zid        }{ \mathrm{z}            } % index z
\newcommand{ \tid        }{ \mathrm{t}            } % index t
\newcommand{ \cid        }{ \mathrm{m}            } % index m (number of contacts)
\newcommand{ \effid      }{ \mathrm{e}            } % endeffector index
\newcommand{ \hid        }{ \mathfrak{h}          } % index h
\newcommand{ \rid        }{ \mathfrak{r}          } % index r
\newcommand{ \numiters   }{ \pazocal{I}           } % number of iterations
\newcommand{ \coptrq     }{ \tau                  } % torque at center of pressure
\newcommand{ \efftrq     }{ \gamma                } % torque at endeffector frame
\newcommand{ \effcross   }{ \kappa                } % endeffector cross product
\newcommand{ \timeopt    }{ \Delta                } % time opt variable
\newcommand{ \coms       }{ \mathbf{r}            } % center of mass
\newcommand{ \moms       }{ \mathbf{h}            } % momenta
\newcommand{ \lmoms      }{ \mathbf{l}            } % linear momentum
\newcommand{ \amoms      }{ \mathbf{k}            } % angular momentum
\newcommand{ \forcevars  }{ \mathbf{f}            } % endeffector force
\newcommand{ \lmomds     }{ \mathbf{\dot{l}}      } % linear momentum rate
\newcommand{ \amomds     }{ \mathbf{\dot{k}}      } % angular momentum rate
\newcommand{ \lforcevars }{ \mathbf{\mathfrak{f}} } % endeffector force local frame
\newcommand{ \copvars    }{ \mathbf{\mathfrak{z}} } % endeffector center of pressure
\newcommand{ \effcomlen  }{ \ell                  } % distance between CoM and endeffector
\newcommand{ \numjnts    }{ \mathrm{n}            } % number of joints
\newcommand{ \robotpos   }{ \mathbf{q}            } % robot posture
\newcommand{ \robotvel   }{ \mathbf{\dot{q}}      } % robot velocity
\newcommand{ \robotacc   }{ \mathbf{\ddot{q}}     } % robot acceleration
\newcommand{ \jntpos     }{ \mathbf{q}_{\textrm{jnt}}} % joint positions
\newcommand{ \fbpose     }{ \mathbf{x}            } % pose of the floating base
\newcommand{ \effpos     }{ \mathbf{p}            } % endeffector position
\newcommand{ \effrot     }{ \mathbf{R}            } % endeffector rotation
\newcommand{ \gravity    }{ \mathbf{g}            } % gravity vector
\newcommand{ \robotmass  }{ m                     } % robot mass
\newcommand{ \setreals   }{ \mathbb{R}            } % set of real numbers
\newcommand{ \setp       }{ \pazocal{Q}^{+}       } % set of PD functions
\newcommand{ \setpm      }{ \pazocal{Q}^{\pm}     } % set of DPD functions
\newcommand{ \setacteff  }{ \effid_{\mathrm{cnt}} } % set of active endeffectors
\newcommand{ \selmat     }{ \mathbf{S}           } % selection matrix
\newcommand{ \massmat    }{ \mathbf{M}           } % mass matrix
\newcommand{ \nlterms    }{ \mathbf{h}           } % nonlinear terms
\newcommand{ \effjac     }{ \mathbf{J}           } % endeffector jacobian
\newcommand{ \jnttrq     }{ \Lambda              } % joint torques
\newcommand{ \wrench     }{ \lambda              } % endeffector wrench
\newcommand{ \vecspace   }{ V                     } % vector space
\newcommand{ \uvector    }{ \mathbf{u}            } % decomposition vector
\newcommand{ \avector    }{ \mathbf{a}            } % positive term of decomposition
\newcommand{ \bvector    }{ \mathbf{b}            } % negative term of decomposition
\newcommand{ \cvector    }{ \mathbf{c}            } % positive term of decomposition
\newcommand{ \dvector    }{ \mathbf{d}            } % negative term of decomposition
\newcommand{ \pvector    }{ \mathfrak{p}          } % sum vector a_i+b_i
\newcommand{ \ascalar    }{ \bar{a}               } % scalar for + term of decomposition
\newcommand{ \bscalar    }{ \bar{b}               } % scalar for - term of decomposition
\newcommand{ \scpenalty  }{ \eta                  } % penalty for soft-constraint
\newcommand{ \trpenalty  }{ \sigma                } % value for trust region constraint
\newcommand{ \friccoeff  }{ \mu                   } % friction coefficient
\newcommand{ \cnthorizon }{ \pazocal{T}           } % continuous time horizon
\newcommand{ \dishorizon }{ N                     } % discrete time horizon
\newcommand{ \converr    }{ \epsilon              } % convergence error
\newcommand{ \surfnormal }{ \mathfrak{N}          } % surface normal
\newcommand{ \surfpoint  }{ \omega                } % surface point
\newcommand{ \surface    }{ \pazocal{S}           } % surface terrain
\newcommand{ \membership }{ \pazocal{U}           } % membership constraint
\newcommand{ \cntsmap    }{ \varphi               } % map surfaces to plan
\newcommand{ \surfmat    }{ \Xi                   } % matrix for surface membership
\newcommand{ \surfvec    }{ \xi                   } % vector for surface membership
\newcommand{ \numsurf    }{ \mathrm{R}            } % number of terrain surfaces
\newcommand{ \surfselmat }{ \pazocal{H}           } % matrix for surface selection
\newcommand{ \cosmat     }{ \mathfrak{C}          } % binary matrix for cosine
\newcommand{ \sinmat     }{ \mathfrak{S}          } % binary matrix for sine
\newcommand{ \yawlimit   }{ \Theta                } % limit for yaw angle linear approx
\newcommand{ \yawangle   }{ \theta                } % yaw angle within linear approx
\newcommand{ \slopeyaw   }{ \mathfrak{u}          } % slope linear approx
\newcommand{ \interpyaw  }{ \mathfrak{v}          } % intercept linear approx
\newcommand{ \cosapprox  }{ \mathfrak{c}          } % cosine value of linear approx
\newcommand{ \sinapprox  }{ \mathfrak{s}          } % sine value of linear approx
\newcommand{ \numcntsopt }{ \mathrm{M}            } % number of contacts being optimized
\newcommand{ \numpwappr  }{ \mathrm{H}            } % number of piecewise affine approx
\newcommand{ \numsoc     }{ \nu                   } % number soc constraints
\newcommand{ \sizesoc    }{ \iota                 } % size of soc constraints
\newcommand{ \friccone   }{ \pazocal{F}           } % friction cone
\newcommand{ \focuspnts  }{ \pazocal{P}           } % focus points for soc constraints
\newcommand{ \focusdist  }{ \pazocal{D}           } % distance to focus points
\newcommand{ \fcost      }{ \phi                  } % cost function
\newcommand{ \fconsensus }{ \Phi                  } % cost consensus
\newcommand{ \fquad      }{ \mathfrak{Q}          } % quadratic function
\newcommand{ \fbilinear  }{ \mathfrak{B}          } % bilinear function
\newcommand{ \frotation  }{ \mathfrak{R}          } % extract rotation function
\newcommand{ \fdcdecompP }{ \chi                  } % positive dc function
\newcommand{ \fdcdecompN }{ \zeta                 } % negative dc function

\newcommandx{\indexed}[5][1=,2=,4=,5=]{                                     % helper for indices
	\prescript{\mathrm{#1}}{\mathrm{#2}}{#3}^{\mathrm{#4}}_{\mathrm{#5}}
}

\newtheorem{remark}{Remark}
\newtheorem{definition}{Definition}
\mdtheorem[backgroundcolor=white]{mytheorem1}{Trust-Region Approximation of $\setp$ Expressions}
\mdtheorem[backgroundcolor=white]{mytheorem2}{Soft-Constraint Approximation of $\setp$ Expressions}

\newcommand{\norm}[1]{\left\lVert#1\right\rVert} % 2-norm function
\newcommand\defequal{\stackrel{\mathclap{\normalfont\mbox{def}}}{=}}

\newcolumntype{L}[1]{>{\raggedright\arraybackslash}p{#1}} 	% align to the left
\newcolumntype{C}[1]{>{\centering\arraybackslash}p{#1}} 	% align in the center
\newcolumntype{R}[1]{>{\raggedleft\arraybackslash}p{#1}} 	% align to the right

\newlength{\FSZ}
\usetikzlibrary{calc, mindmap,shapes,arrows}
\newcommand{\drawvideo}[3]{
\noindent\pgfmathsetlength{\FSZ}{\linewidth/#2}
\begin{tikzpicture}[outer sep=0pt,inner sep=0pt,x=\FSZ,y=\FSZ]
	\draw[color=lightblue!50!black] (0,0) node[outer sep=0pt,inner sep=0pt,text width=\linewidth,minimum height=0] (video) {\noindent#3};
	\path [fill=lightblue!50!black,line width=0pt] 
	(video.north west) rectangle ([yshift=\FSZ] video.north east) 
	\foreach \x in {1,2,...,#2} {
		{[rounded corners=0.6] ($(video.north west)+(-0.7,0.8)+(\x,0)$) rectangle +(0.4,-0.6)}
	};
	\path [fill=lightblue!50!black,line width=0pt] 
	([yshift=-1\FSZ] video.south west) rectangle (video.south east) 
	\foreach \x in {1,2,...,#2} {
		{[rounded corners=0.6] ($(video.south west)+(-0.7,-0.2)+(\x,0)$) rectangle +(0.4,-0.6)}
	};
	\foreach \x in {1,...,#1} {
		\draw[color=lightblue!50!black] ([xshift=\x\linewidth/#1] video.north west) -- ([xshift=\x\linewidth/#1] video.south west);
	}
	\foreach \x in {0,#1} {
		\draw[color=lightblue!50!black] ([xshift=\x\linewidth/#1,yshift=1\FSZ] video.north west) -- ([xshift=\x\linewidth/#1,yshift=-1\FSZ] video.south west);
	}
\end{tikzpicture}
}

\title{\LARGE \bf Efficient Multi-Contact Pattern Generation with Sequential Convex Approximations of the Centroidal Dynamics}

\author{
	Brahayam Ponton$^{1}$, Majid Khadiv$^{1}$, Avadesh Meduri$^{2}$ and Ludovic Righetti$^{1,2}$%
	\thanks{This research was supported by New York University, the Max-Planck Society, the European Union's Horizon 2020 research and innovation programme (grant agreement No 780684 and European Research Council's grant No 637935), and the US National Science Foundation grant CMMI-1825993.}% <-this % stops a space
	\thanks{$^{1}$ Max Planck Institute for Intelligent Systems, T\"ubingen - Germany}
	\thanks{$^{2}$ New York University, New York - USA}
}

\begin{document}

\maketitle
\thispagestyle{empty}
\pagestyle{empty}

%%%%%%%%%%%%
% Abstract %
%%%%%%%%%%%%
\begin{abstract}
This paper investigates the problem of efficient computation of physically consistent multi-contact behaviors. Recent work showed that under mild assumptions, the problem could be decomposed into simpler kinematic and centroidal dynamic optimization problems. Based on this approach, we propose a general convex relaxation of the centroidal dynamics leading to two computationally efficient algorithms based on iterative resolutions of second order cone programs. They optimize centroidal trajectories, contact forces and, importantly, the timing of the motions. We include the approach in a kino-dynamic optimization method to generate full-body movements. Finally, the approach is embedded in a mixed-integer solver to further find dynamically consistent contact sequences. Extensive numerical experiments demonstrate the computational efficiency of the approach, suggesting that it could be used in a fast receding horizon control loop. Executions of the planned motions on simulated humanoids and quadrupeds and on a real quadruped robot further show the quality of the optimized motions.
\end{abstract}
%
%
%%%%%%%%%%%%%%%%
% Introduction %
%%%%%%%%%%%%%%%%
\section{Introduction}
The computation of multi-contact motions remains a difficult yet important challenge for legged locomotion and manipulation in order to afford more versatile behaviors in complex environments. Of particular interest are methods that can compute such motions in real-time without making restrictive assumptions on the solution set. Indeed, they can provide the necessary adaptive behavior required in uncertain environments without trading-off motion versatility.

Very successful walking pattern generators often rely on simplified linear models of the dynamics \cite{Kajita:2003gj} as they offer important computational advantages that make them suitable for receding horizon control \cite{DBLP:conf/iros/Wieber08, journals/corr/KuindersmaPT13, journals/trob/EnglsbergerOA15}. Unfortunately, these models are fundamentally restricted to locomotion patterns with predefined gaits on quasi-flat grounds. While extensions of such models can enable the use of hands to maintain balance \cite{mason_mpc_2018}, they make substantial assumptions on the admissible gaits and are thus limited by the range of gaits they can generate.

Complete rigid body dynamics models including interaction dynamics, in principle, afford the synthesis of a wider range of behaviors for more complex motion tasks. Despite the inherent computational challenges, very impressive motions can be computed \cite{journals/tog/MordatchTP12, DBLP:conf/iros/TassaET12, DBLP:conf/iros/ErezT12, DBLP:conf/iros/KoenemannPTTSBM15, TimeSwitchedJonas, Neunert2017TrajectoryOT, DBLP:journals/corr/abs-1711-11006, DBLP:conf/wafr/PosaT12, Manchester09stabledynamic, MombaurSomersault, DBLP:conf/syroco/KochMS12}. However such approaches are often limited for receding horizon control as they require the resolution of non-convex, high dimensional optimization problems, often with complex nonlinear constraints such as complementarity constraints for contact dynamics.

Middle-complexity options that decouple the pattern generation problem into simpler sub-problems have also been studied. They typically assume that a sequence of contact configuration be provided first, typically using an efficient search algorithms for contact sequences \cite{Tonneau:2018dm,escande2013planning,lin2017,DBLP:conf/humanoids/DeitsT14, NishiS14}. Of special interests are methods based on the centroidal dynamics of the robot \cite{OrinCentroidalMomentum, Kajita:2003gj, DynamicsAnalysis} which have become very popular recently \cite{Wensing:2013fm, DBLP:conf/humanoids/DaiVT14, Herzog-2016b}. Indeed, under mild assumptions on the kinematic and actuation feasibility, this model provides sufficient conditions to plan dynamically consistent full-body motions with multiple contacts. This model is simple enough to be amenable to online resolution and at the same time expressive to plan complex behaviors \cite{JustinMomentumOptimization, TROCarpentier, winkler18, Audren:2014gl, AlexHumanoidsPaper}. It is then possible to combine momentum dynamics with a full kinematic model to plan highly dynamic motions \cite{DBLP:conf/humanoids/DaiVT14}. This decomposition between centroidal dynamics and kinematics models was, for example, leveraged to create an alternating algorithm that efficiently computes full-body motions in multi-contact by iteratively solving two separate optimization problems until they reach consensus \cite{Herzog-2016b, AlexHumanoidsPaper}.  This connection has then been further explored  in \cite{budhiraja2018dynamics}, which proposed a method to optimize both centroidal and full-body motions using an Alternating Direction Method of Multipliers formulation.

While promising,  approaches based on the centroidal momentum dynamics are inherently non-convex and thus still challenging to solve efficiently. This led researchers to focus on the mathematical structure of the problem to derive more efficient methods. For example, convex bounds on the angular momentum rate (that maximizes the contact wrench cone margin) are used to minimize a worst-case bound on the $l_1$ angular momentum norm via convex optimization \cite{Dai:2016hz}. In \cite{JustinMomentumOptimization, TROCarpentier}, the bilinear terms of the momentum dynamics and timings are handled by a dedicated multiple-shooting solver and, proxy constraints are used for handling whole-body limits based on an offline learning method. \cite{Audren:2014gl} exploits a linear approximation of the momentum dynamics based on a lower dimensional space projection and an adaptive method for timing optimization to control a robot in multi-contact scenarios in a receding horizon fashion. In \cite{TOPP, Caron:2016wt}, the interpretation of friction cones as dual twists allows to compute online cones of feasible CoM accelerations. The resulting bilinear constraints are decoupled into linear pairs via a conservative trajectory-wide contact-stability criterion for online motion generation. \ctxt{Timings between contact switches are optimized online by solving an easy-to-solve nonlinear problem.}

% Recently, \cite{winkler18} presented a novel phase-based parameterization of endeffectors and a smooth terrain description to formulate an optimization problem, online solvable with a general nonlinear solver, able to select gait-sequences, momentum and timings, swing-motions and body poses, thus bringing to mind the question if a problem approximation is actually needed.

In \cite{Herzog-2016b}, we further studied the problem structure and proposed an analytic decomposition of positive and negative definite terms of the problem Lagrangian based on the decomposition of angular momentum non-convex terms. This led to a solver with improved convergence properties. In our previous work \cite{ConvexModelMomentumDynamics}, we proposed a convex relaxation of the problem that suggested the use of a proxy function to minimize angular momentum, namely the sum of the squares of the terms composing the non convex part of the dynamics. While computationally very efficient, this approach was limited as it did not allow the inclusion of an explicit target angular momentum in the cost function, therefore severely limiting the space of solutions. Moreover, the approach could not be used with the alternating full-body motion optimization method discussed above.

In this paper, extending our preliminary work \cite{TimeOptimization}, we study a general convex relaxation of the problem that allows the explicit inclusion of angular momentum objectives and naturally extends to the optimization of timing, a feature missing in most contributions on centroidal dynamics optimization. The main contributions of the paper are\footnote{Part of the material was presented at the 2018 IEEE-RAS International Conference on Robotics and Automation \cite{TimeOptimization}. Contribution 1 is an extension of this work, Contributions 2 and 3 are novel, Contribution 4 extends simulation results to Contribution 2 and 3 and presents real robot experiments.}
\begin{itemize}
	\item[1] Exploiting the structure of the centroidal dynamics optimization problem, we propose two computationally efficient algorithms formulated as a sequence of convex second order cone programs to compute physically consistent center of mass, angular momentum and contact force trajectories and demonstrate how timing optimization can be efficiently included.
	\item[2] We show how our approach can be efficiently used with the kino-dynamic optimization method proposed in \cite{AlexHumanoidsPaper} to generate full-body physically-consistent movements. We further extend the approach to also include actuation limit constraints.
	\item[3] We extend the approach in a mixed-integer program to find dynamically consistent contact sequences and locations.
	\item[4] Finally, we evaluate the capabilities and limitations of our approach in simulation on several multi-contact scenarios for a biped and a quadruped robot, we study the benefits of timing optimization to extend the range of possible behaviors and demonstrate the execution of these movements on a real quadruped robot.
\end{itemize}

The software implementation of the algorithms presented in this paper is open-source and freely available \cite{opensourcelink}. We state the problem and present background material in Section \ref{sec:problem_formulation}. In Section \ref{sec:opt_movement}, we detail the motion optimization approach and in Section \ref{sec:contacts_planning} the contacts planning approach using mixed integer programming. We present simulation and real robot results in Section \ref{sec:experiments} and \ctxt{discuss the features and limitations of our proposed framework in Section \ref{sec:discussion}. Finally, we} conclude in Section \ref{sec:conclusion}.
%

%%%%%%%%%%%%%%%%%%%%%%%%%%%%%%%%%%%%%%%%%
% Preliminaries and problem formulation %
%%%%%%%%%%%%%%%%%%%%%%%%%%%%%%%%%%%%%%%%%
\section{Preliminaries and Problem Formulation} \label{sec:problem_formulation}
%
%============= Control architecture ==============%
\begin{figure}
	\centering
	\includegraphics[width=0.98\linewidth]{\filename{figures/ctrlarch/ControlArchitecture}}
	\caption[]{\small Our architecture maps a high-level task description into functional motions. The initial state $\coms_{0}, \lmoms_{0}, \amoms_{0}$ of the robotic platform (simulated humanoid or a real quadruped robot), a desired CoM motion $\Delta\coms$, a description of the $\numsurf$ surfaces that compose the terrain and a set of costs $\indexed{\fcost}[cnt][\tid](\cdot), \indexed{\fcost}[kin][\tid](\cdot), \indexed{\fcost}[dyn][\tid](\cdot), \indexed{\fcost}[fb][\tid](\cdot)$ are used to select a set of surfaces $\indexed{\surface}[][\{ \rid \}]$ that support a dynamic motion, optimize a kino-dynamic motion over a discrete time horizon $\dishorizon$, and synthesize a set of feedback gains $\mathbf{K}_{\moms}, \mathbf{K}_{\robotpos}, \mathbf{K}_{\wrench_{\effid}}$ that define closed-loop behaviors to be realized by an inverse dynamics controller as in \cite{AlexAuroPaper, compliant_terrain_adaptation, robust_biped_walking}.
	}
	\label{fig:ExecutionArchitecture}
	\vspace{-0.3cm}
\end{figure}
%=================================================%
In this section, we introduce the centroidal dynamics optimization problem for multi-contact locomotion in the larger context of full-body optimization. First, we provide an overview of the larger kino-dynamic optimization problem, present the structured approach used in our architecture to tackle it and outline the centroidal dynamics optimization problem, which is the core focus of this paper.
Our overall approach is summarized in Figure \ref{fig:ExecutionArchitecture}.
From a task description we first select a sequence on physically-feasible
contact sequences using mixed integer programming (Sec. \ref{sec:contacts_planning}). This sequence is used to
optimize a time-optimal full-body movements using our kino-dynamic solver (Fig. \ref{fig:KinDynStructure}).
These movements are then tracked with an instantaneous whole-body feedback controller.

\subsection{Kino-dynamic optimization of multi-contact behaviors}
To synthesize full-body multi-contact behaviors, we seek to efficiently solve an optimal control problem of the form
% \vspace{-0.1cm}
%
% Kino-dynamic optimization of multi-contact behaviors
%
\begin{subequations}
	\begin{align}
		%
		% General objective function
		%
		\min_{\robotpos(\tid), \jnttrq(\tid), \indexed{\wrench}[][\effid](\tid)} & \indexed{\fcost}[][end] \left(\robotpos, \robotvel, \robotacc, \jnttrq, {\indexed{\wrench}[][\effid]} \right) + \int\limits_{0}^{\cnthorizon} \indexed{\fcost}[][\tid]\left(\robotpos, \robotvel, \robotacc, \jnttrq, {\indexed{\wrench}[][\effid]} \right) \mathrm{d}\tid \label{eq_problem_cost} \\
		%
		% Equations of motion
		%
		\textrm{subject to}\;\; &  \indexed{\massmat}(\robotpos)\robotacc + \indexed{\nlterms}(\robotpos,\robotvel) = \selmat^{T} \jnttrq + \hspace{-0.1cm} \sum_{\effid \in \setacteff} \hspace{-0.1cm} \indexed{\effjac}[][\effid](\robotpos)^{T} \indexed{\wrench}[][\effid] \label{eq_of_motion} \\
		%
		% Joint limits
		%
		& \; \jntpos \in [ \indexed[min]{\jntpos}, \indexed[max]{\jntpos} ]  \label{eq_joint_limits} \\
		%
		% Torque limits
		%
		& \; \jnttrq \in [\indexed[min]{\jnttrq}, \indexed[max]{\jnttrq}] \label{eq_torque_limits}\\
		%
		% Contact forces, velocity and acceleration limits
		%
		& \; (\indexed{\wrench}[][\effid], \robotvel, \robotacc) \in \Omega \label{eq:force_vel_constraints}
	\end{align}
	\label{eqns_general_problem}
\end{subequations}
%
%\vspace{-0.4cm}
%
which minimizes a performance cost $\fcost(\cdot)$, composed of a terminal cost $\indexed{\fcost}[][end]$ and the integral of a running cost $\indexed{\fcost}[][\tid]$, over a finite time horizon $\cnthorizon$ under a set of physical constraints. It enforces the equations of motion for a floating-base rigid-body system (Eq. $\eqref{eq_of_motion}$), joint and torque limits (Eqs. \eqref{eq_joint_limits}-\eqref{eq_torque_limits}), as well as contact forces, velocity and acceleration constraints (Eq. \eqref{eq:force_vel_constraints}).
Here, $\robotpos = \begin{bmatrix} \fbpose^{T} \; \jntpos^{T} \end{bmatrix}^{T}$ denotes the robot posture composed of $\fbpose \in SE(3)$, the pose of the floating-base relative to an inertial frame, and $\jntpos \in \setreals^\numjnts$, the joint positions, where $\numjnts$ is the number of joints. $\jnttrq(\tid) \in \setreals^\numjnts$ are joint torques and $\indexed{\wrench}[][\effid](\tid) \in \setreals^{6}$ is the contact wrench of endeffector $\effid \in \setacteff$ (where $\setacteff$ is the set of endeffectors in contact with the environment at the time in question).
$\indexed{\massmat}(\robotpos) \in \setreals^{(\numjnts+6) \times (\numjnts+6)}$ is the inertia matrix; $\indexed{\nlterms}(\robotpos,\robotvel) \in \setreals^{\numjnts+6}$ a vector of generalized forces including Coriolis, centrifugal, gravity and joint friction forces. $\selmat = \begin{bmatrix} \mathbf{0}^{\numjnts \times 6} \; \mathbf{I}^{\numjnts \times \numjnts} \end{bmatrix}$ is a selection matrix reflecting the system under-actuation and $\indexed{\effjac}[][\effid](\robotpos) \in \setreals^{6 \times (\numjnts+6)}$ is the contact Jacobian of endeffector $\effid$. The pre-superscripts $\mathrm{min}$ and $\mathrm{max}$ for joint positions $\jntpos$ and joint torques $\jnttrq$ denote their minimum and maximum limits. The set $\Omega$ denotes constraints such as friction or non-sliding contacts, that will be explicitly defined within the next subsection. \ctxt{Note that additional kinematic constraints could also be added to the problem without changing the reasoning below.}

The problem described in Eq. \eqref{eqns_general_problem} is nonlinear, non-convex and computationally intensive and we seek to formulate a more tractable approximation without sacrificing the versatility of motion synthesis. The equations of motion can be decomposed into actuated (superscript $\mathrm{a}$) and unactuated parts (superscript $\mathrm{u}$)
%
% Decomposition on Equations of Motion
%
\begin{subequations}
	\begin{align}
	%
	% Unactuated part EoM
	%
	\indexed{\massmat}[u](\robotpos)\robotacc + \indexed{\nlterms}[u](\robotpos,\robotvel) &= \sum_{\effid \in \setacteff} \indexed{\effjac}[u][\effid](\robotpos)^{T} \indexed{\wrench}[][\effid] \label{eq_unactuated_part} \\
	%
	% Actuated part EoM
	%
	\indexed{\massmat}[a](\robotpos)\robotacc + \indexed{\nlterms}[a](\robotpos,\robotvel) &= \sum_{\effid \in \setacteff} \indexed{\effjac}[a][\effid](\robotpos)^{T} \indexed{\wrench}[][\effid] + \jnttrq \label{eq_actuated_part}
	\end{align}
\end{subequations}
As shown in \cite{AlexAuroPaper}, the actuated part of the dynamics provides the necessary actuation torques needed to achieve any combination of desired acceleration $\robotacc$ and contact forces $\indexed{\wrench}[][\effid]$. Thus, assuming sufficient actuation $\jnttrq$, it is possible to ignore the actuated part of the equations of motion (Eq. \eqref{eq_actuated_part}) and base the synthesis of multi-contact behaviors only on the unactuated part (Eq. \eqref{eq_unactuated_part}). 
\ctxt{As we will later in the paper, it is nevertheless possible to add torque limits in the decoupled optimization problems.}
In \cite{robust_biped_walking, WieberNonholonomy}, it has been shown that the right-hand side of the unactuated part and the gravitational effects of the vector of nonlinear terms $\indexed{\nlterms}[u](\robotpos,\robotvel)$ that relate the acceleration of the floating-base to external contact forces, are equivalent to the robot centroidal momentum dynamics
%
% Centroidal momentum dynamics
%
\begin{equation}
	%
	% Robot momenta
	%
	\begin{bmatrix}
		\lmomds \\[0.5em]
		\amomds \\[0.5em]
	\end{bmatrix} =
	%
	% Definition in terms of force and torques
	%
	\underbrace{
		\begin{bmatrix}
			\robotmass \gravity + \sum\limits_{\effid \in \setacteff} \indexed{\forcevars}[][\effid] \\[0.0em]
			\sum\limits_{\effid \in \setacteff} ( (\indexed{\effpos}[][\effid] + \indexed{\effrot}[\xid,\yid][\effid] \indexed{\copvars}[][\effid] - \coms) \times \indexed{\forcevars}[][\effid] + \indexed{\effrot}[\zid][\effid] \indexed{\coptrq}[][\effid] ) \\[0.0em]
		\end{bmatrix}
	}_{\textrm{From \ctxt{Newton-Euler} dynamics}}\label{eqns_momentum_dynamics}
\end{equation}
%
%=============== Notation Summary ================%
\begin{figure}
	\centering
	\includegraphics[width=0.68\linewidth]{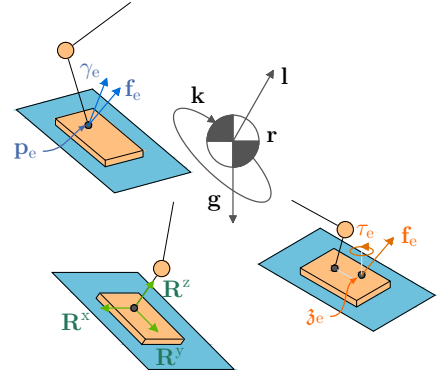}
	\caption[]{\small The figure illustrates the representation used in the paper. }
	\label{fig1:schematic}
\end{figure}
%=================================================%
%
%============= Kino-dynamic approach =============%
\begin{figure}
	\centering
	\includegraphics[width=0.36\textwidth]{\filename{figures/kinodyn/KinoDynStructure}}
	\caption[]{\small Schematic of the kino-dynamic optimization approach that iteratively computes contact force $\indexed{\wrench}[][\effid]$ and whole-body trajectories $\robotpos, \robotvel, \robotacc$ until convergence of the common set of variables: CoM $\indexed{\coms}[][\tid]$, robot momenta $\indexed{\lmoms}[][\tid], \indexed{\amoms}[][\tid]$ and endeffector poses $\indexed{\effpos}[][\effid,\tid]$. The vector $\indexed{\moms}[][\tid]$ is built by vertically stacking CoM and robot momenta. The pre superscripts $\mathrm{kin}$ and $\mathrm{dyn}$ relate the variables to the problem they are a solution for. The optimization objective $\fcost$ is assumed to be separable and composed by $\indexed{\fcost}[dyn][\tid] + \indexed{\fcost}[kin][\tid]$. Finally, the cost penalties $\indexed{\fconsensus}[dyn][\tid], \indexed{\fconsensus}[kin][\tid]$ ensure the consensus of the solutions at convergence.}
	\label{fig:KinDynStructure}
\end{figure}
%=================================================%
%

The center of mass position is denoted $\coms$ and the linear and angular momentum expressed at the CoM are written as $\lmoms$ and $\amoms$. $\robotmass$ is the robot mass and $\gravity$ the gravity vector. The endeffector frame is located at the endeffector position $\indexed{\effpos}[][\effid]$, and it is oriented so that $\indexed{\effrot}[\zid][\effid] \in \setreals^{3 \times 1}$ is normal to the contact surface, and $\indexed{\effrot}[\xid][\effid],\indexed{\effrot}[\yid][\effid] \in \setreals^{3 \times 1}$ are aligned with the rectangular shape of the endeffector support surface in the desired motion direction.
\ctxt{The rotation matrix $\indexed{\effrot}[][\effid] = \begin{bmatrix} \indexed{\effrot}[\xid][\effid] & \hspace{-0.2cm} \indexed{\effrot}[\yid][\effid] & \hspace{-0.2cm} \indexed{\effrot}[\zid][\effid] \end{bmatrix}   \in \setreals^{3 \times 3}$ rotates quantities from endeffector to inertial frame. For instance, the endeffector force $\indexed{\forcevars}[][\effid]$, expressed in the inertial frame, is equivalent in local endeffector coordinates to $\indexed{\lforcevars}[][\effid] = {\indexed{\effrot}[][\effid]}^{T} \indexed{\forcevars}[][\effid]$. The center of pressure (CoP) $\indexed{\copvars}[][\effid] \in \setreals^{2}$ expressed in local endeffector frame and scalar torque $\indexed{\coptrq}[][\effid]$ at the CoP complete the description of the endeffector wrench. They can be equivalently described by a torque at $\indexed{\effpos}[][\effid]$ as $\indexed{\efftrq}[][\effid] = (\indexed{\effrot}[\xid,\yid][\effid] \indexed{\copvars}[][\effid]) \times \indexed{\forcevars}[][\effid] + \indexed{\effrot}[\zid][\effid] \indexed{\coptrq}[][\effid]$. The endeffector wrench can now be defined as $\indexed{\wrench}[][\effid] = \begin{bmatrix} {\forcevars}^{T}_{\effid} & \hspace{-0.2cm} {\efftrq}^{T}_{\effid} \end{bmatrix}^{T}$. } \ctxt{Figure \ref{fig1:schematic} depicts} coordinate frames and the notation. 

It has been shown \ctxt{\cite{OrinCentroidalMomentum}} that the left-hand side of the unactuated part in Eq. \eqref{eq_unactuated_part}, \ctxt{under an appropriate coordinate transformation from the floating base to the robot's CoM, relates the robot rate of momenta expressed at the robot's center of mass ($\lmomds, \amomds$)} to the robot velocity $\robotvel$ and acceleration $\robotacc$ via the centroidal momentum matrix \ctxt{$\indexed{\massmat}[u][CoM](\robotpos) \in \setreals^{6 \times (\numjnts+6)}$}.

%
% Alternating optimization algorithm
%
\begin{equation}
%
% Definition in terms of kinematics
%
\underbrace{
	\frac{d}{d\tid} \left[ {\indexed{\massmat}[u][CoM](\robotpos) \robotvel} \right]
}_{\textrm{From \ctxt{full-body} kinematics}} =
\indexed{\massmat}[u][CoM](\robotpos) \robotacc + \indexed{\dot{\massmat}}[u][CoM](\robotpos) \robotvel =
%
% Robot momenta
%
\begin{bmatrix}
	\lmomds \\[0.2em]
	\amomds \\[0.2em]
\end{bmatrix}
\label{eqns_kindyn_momentum_dynamics}
\end{equation}

At this point, it becomes clear that the problem of finding feasible multi-contact motions can be reduced to the optimization of centroidal dynamics (Eq. \eqref{eqns_momentum_dynamics}) and the optimization of full-body kinematics (Eq. \eqref{eqns_kindyn_momentum_dynamics}) as long as the motion-induced momentum agrees with the dynamic optimization. In \cite{AlexHumanoidsPaper}, an alternating algorithm to solve the optimal control problem \eqref{eqns_general_problem} using this idea was proposed (see Fig. \ref{fig:KinDynStructure}). It optimized centroidal dynamic motions and full-body kinematics separately, but ensured through added cost penalties that both optimization problems come to an agreement on their common variables: CoM, momentum and contact locations.

In this paper, we use the complete architecture shown in Figure \ref{fig:ExecutionArchitecture} to evaluate our contributions, but our work mostly focuses on the centroidal dynamics optimization problem, which is sufficient to synthesize physically consistent motion behaviors.

\subsection{Dynamic optimization with the centroidal dynamics}
We now present in detail the centroidal dynamics optimization problem we are interested in, that synthesizes a motion plan (timing, contact wrenches and momentum trajectories) under the momentum dynamics (Eq. \eqref{eqns_momentum_dynamics}) and is optimal in terms of a desired quadratic performance objective. First, we discretize the dynamics equations using Euler's methods and then seek a local solution for the following problem:
%
% Dynamics optimization problem with centroidal dynamics
%
\begin{subequations}
	\begin{align}
		\phantom{abcdefghi}
		&\begin{aligned}
			%
			% Objective function
			%
			\mathllap{\min_{\substack{\indexed{\moms}[][\tid], \indexed{\timeopt}[][\tid], \indexed{\effpos}[][\effid,\tid] \\ \indexed{\forcevars}[][\effid,\tid], \indexed{\copvars}[][\effid,\tid], \indexed{\coptrq}[][\effid,\tid]}}}
			&\sum\limits_{\tid=1}^{\dishorizon} \hspace{-0.075cm}
			%
			% Dynamics part of the cost
			%
			\left[ \hspace{-0.05cm}\indexed{\fcost}[dyn][\tid] \hspace{-0.1cm}\left(
			\begin{matrix}
				\indexed{\moms}[][\tid], \indexed{\timeopt}[][\tid], \indexed{\effpos}[][\effid,\tid],\\
				\indexed{\copvars}[][\effid,\tid], \indexed{\forcevars}[][\effid,\tid], \indexed{\coptrq}[][\effid,\tid]
			\end{matrix} \right) \hspace{-0.05cm}+\hspace{-0.025cm}
			%
			% Consensus part of the cost
			%
			\indexed{\fconsensus}[dyn][\tid] \hspace{-0.1cm} \left(
			\begin{matrix}
				\indexed{\moms}[][\tid] - \indexed[kin]{\moms}[][\tid] \\
				\indexed{\effpos}[][\effid,\tid] - \indexed[kin]{\effpos}[][\effid,\tid]
			\end{matrix} \right) \hspace{-0.05cm} \right]
		\end{aligned} \hspace{-0.5cm} \label{eq_dynopt_cost} \\[0.0cm]
		&\begin{aligned}
			%
			% Centroidal dynamics
			%
			\mathllap{\textrm{subject to}}\;
			& \; \moms_{\tid} =
			\begin{bmatrix}
				\indexed{\coms}[][\tid]  \\[0.5em]
				\indexed{\amoms}[][\tid] \\[0.5em]
				\indexed{\lmoms}[][\tid] \\[0.5em]
			\end{bmatrix} = 
			\begin{bmatrix}
				\indexed{\coms}[][\tid-1] + \frac{1}{\robotmass} \indexed{\lmoms}[][\tid] \indexed{\timeopt}[][\tid] \\[0.3em]
				\indexed{\amoms}[][\tid-1] + \sum\limits_{\effid \in \setacteff} \indexed{\effcross}[][\effid,\tid] \indexed{\timeopt}[][\tid] \\[0.6em]
				\indexed{\lmoms}[][\tid-1] + \robotmass \gravity \indexed{\timeopt}[][\tid] + \sum\limits_{\effid \in \setacteff} \indexed{\forcevars}[][\effid,\tid] \indexed{\timeopt}[][\tid] \\[0.0em]
			\end{bmatrix}
		\end{aligned} \hspace{-0.5cm} \label{eq_dynopt_momentum} \\
		&\begin{aligned}
			%
			% Torque at endeffector position
			%
			\mathllap{}
			&\; \indexed{\effcross}[][\effid,\tid] = (\indexed{\effpos}[][\effid,\tid] - \indexed{\coms}[][\tid]) \times \indexed{\forcevars}[][\effid,\tid] + \indexed{\efftrq}[][\effid,\tid]
		\end{aligned}  \label{eq_dynopt_kappa} \\
		&\begin{aligned}
			%
			% Torque at center of pressure
			%
			\mathllap{}
			&\; \indexed{\efftrq}[][\effid,\tid] = ( \indexed{\effrot}[\xid,\yid][\effid,\tid] \indexed{\copvars}[][\effid,\tid] )  \times \indexed{\forcevars}[][\effid,\tid] + \indexed{\effrot}[\zid][\effid,\tid] \indexed{\coptrq}[][\effid,\tid]
		\end{aligned} \label{eq_dynopt_gamma} \\
		&\begin{aligned}
			%
			% Membership to contact surfaces
			%
			\mathllap{}
			&\; \indexed{\effpos}[][\effid,\tid] \in \membership(\indexed{\surface}[][\rid = \cntsmap(\effid,\tid)])
		\end{aligned} \label{eq_dynopt_effpos} \\[0.0em]
		&\begin{aligned}
			%
			% Constraints on time discretization variable
			%
			\mathllap{}
			&\; \indexed{\timeopt}[][\tid] \in [ \indexed[min]{\timeopt}[][\tid], \indexed[max]{\timeopt}[][\tid] ]
		\end{aligned} \label{eq_dynopt_time}\\[0.0em]
		&\begin{aligned}
			%
			% Constraints on center of pressure limits
			%
			\mathllap{}
			&\; \indexed{\copvars}[\xid,\yid][\effid,\tid] \in [ \indexed[min]{\copvars}[\xid,\yid], \indexed[max]{\copvars}[\xid,\yid] ]
		\end{aligned} \label{eq_dynopt_cop} \\[0.0em]
		&\begin{aligned}
			%
			% Friction cone constraints
			%
			\mathllap{}
			&\; \norm{ \indexed{\lforcevars}[\xid,\yid][\effid,\tid] }_{2} \le \friccoeff \indexed{\lforcevars}[\zid][\effid,\tid], \hspace{0.25cm} \indexed{\lforcevars}[\zid][\effid,\tid] > 0
		\end{aligned} \label{eq_dynopt_frccone} \\
		&\begin{aligned}
			%
			% Heuristic on distance between center of mass and contacts
			%
			\mathllap{}
			&\; \norm{\indexed{\effpos}[][\effid,\tid] - \coms_{\tid}}_{2} \le \indexed[max]{\pazocal{L}}[][\effid]
		\end{aligned} \label{eq_dynopt_eff_length} \\[0.0em]
		&\begin{aligned}
			%
			% Contraint on torque limits
			%
			\mathllap{}
			&\; \indexed{\jnttrq}[][\tid] = \big( \indexed{\massmat}[a](\indexed[*]{\robotpos}) \indexed[*]{\robotacc} + \indexed{\nlterms}[a](\indexed[*]{\robotpos},\indexed[*]{\robotvel})  \big. \\[0.1cm]
			& \big. \hspace{1.1cm} - \sum_{\effid \in \setacteff} \indexed{\effjac}[a][\effid](\indexed[*]{\robotpos})^{T} \indexed{\wrench}[][\effid,\tid] \big) \; \in [{\indexed[min]{\jnttrq}},{\indexed[max]{\jnttrq}}]  \label{eq_dynopt_joint_torques}
		\end{aligned}
	\end{align}
	\label{dynopt_problem}
\end{subequations}

We minimize a quadratic cost \eqref{eq_dynopt_cost} that includes a running cost $\indexed{\fcost}[dyn][\tid]$ composed by user-defined task costs (such as reaching a CoM position or moving through a way-point) and regularization of control variables (such as contact wrenches or Euler discretization of time $\indexed{\timeopt}[][\tid]$). When the problem is solved in the context of the alternating kino-dynamic optimization procedure, it also includes a consensus cost $\indexed{\fconsensus}[dyn][\tid]$ penalizing momentum trajectories and contact locations deviating from the solution of the kinematic optimization step. The problem is optimized over a discrete time horizon $\dishorizon \approx \cnthorizon / \indexed[0]{\timeopt}[][\tid]$ computed using the initial guess for the timestep variable $\indexed{\timeopt}[][\tid]$, that corresponds to the difference between time at step $\tid$ and $\tid-1$.

The constraints (defined for all active endeffectors $\effid \in \setacteff$ and timesteps $ \tid$) include consistency with the centroidal dynamics \eqref{eq_dynopt_momentum}-\eqref{eq_dynopt_gamma}. Here, we have formulated the dynamics using torques at each contact's center of pressure and added an extra variable $\indexed{\effcross}[][\effid,\tid]$ which will facilitate the formulation of the time optimization algorithm. Other constraints include: constraints on the endeffector locations to remain on the assigned contact surface \eqref{eq_dynopt_effpos} modeled as linear inequality constraints (cf. Section \ref{sec:contact_membership} for a detailed explanation of $\indexed{\effpos}[][\effid,\tid] \in \membership(\indexed{\surface}[][\rid = \cntsmap(\effid,\tid)])$), box constraints to restrict the timestep variable \eqref{eq_dynopt_time} between a lower $\indexed[min]{\timeopt}[][\tid]$ and upper $\indexed[max]{\timeopt}[][\tid]$ limits, constraints to maintain the CoP of the endeffectors (assumed to be rectangular) within the  support region \eqref{eq_dynopt_cop} defined by the lower $\indexed[min]{\copvars}[\xid,\yid]$ and $\indexed[max]{\copvars}[\xid,\yid]$ upper limits, friction cone constraints \eqref{eq_dynopt_frccone} (with friction coefficient $\friccoeff$) and a heuristic constraint to ensure that the contact locations remain reachable expressed as a distance from the CoM \eqref{eq_dynopt_eff_length} that cannot exceed a predefined value $\indexed[max]{\pazocal{L}}[][\effid]$. A linear time-varying approximation of the torque limits constraint \eqref{eq_dynopt_joint_torques} along the motion trajectory $\indexed[*]{\robotpos}, \indexed[*]{\robotvel}, \indexed[*]{\robotacc}$ optimized in the previous kinematics optimization problem can also be considered and provides the ability to adapt contact wrenches to satisfy torque limits.

In its general form, the optimization problem defined in Eq. \eqref{dynopt_problem} is non-convex. Its non-convexities are due to the cross products from the angular momentum dynamics and the bilinear terms from the timestep variable. In the next section, we leverage the structure of the problem and propose two algorithms based on convex relaxations to efficiently solve it. We then extend the approach to also optimally select contact surfaces that support a dynamic motion by embedding the dynamics model within a custom mixed-integer solver.

\begin{remark}
	In general, we can write down the relation between the contact forces and the CoM motion  in two ways, 1) using the contact wrench sum (CWS) at the CoM and imposing contact wrench cone (CWC) constraints \cite{Caron:2016wt, Dai:2016hz, AdiosZMP, compliant_terrain_adaptation} 2) using the contact forces (or wrench) at each end-effector and imposing directly contact force constraints \cite{Herzog-2016b, JustinMomentumOptimization, winkler18, TimeOptimization}. In this paper, we use the second approach. The main advantage of this approach is the capability of adapting contact location of the end-effectors. The main caveat is that for more than one end-effector in contact (i.e. $\effid \geq 2$), the number of decision variables (i.e. $6 \times \effid$) is more than the minimal representation of the centroidal wrench (i.e. 6). However, the cross product term between decision variables is inherent in the centroidal dynamics and our approach to dealing with the cross-product (and bilinear terms in general) is also applicable to a CWC formulation.
\end{remark}

%
%%%%%%%%%%%%%%%%%%%%%%%%%%%%%%%%%%%%%%%%%%%%%
% Centroidal Momentum Dynamics Optimization %
%%%%%%%%%%%%%%%%%%%%%%%%%%%%%%%%%%%%%%%%%%%%%
\section{Centroidal Momentum Dynamics Optimization} \label{sec:opt_movement}
This section presents our approach to solve the centroidal dynamics optimization based on an analytical decomposition of non-convex bilinear expressions as a difference of quadratic functions, whose known curvature is exploited to design efficient iterative convex approximations. In the following, we analyze the nature of nonconvexities of problem \eqref{dynopt_problem}, propose two convex relaxations to approximate them and, detail the optimization procedures and their convergence criteria.
\subsection{Bilinear terms as difference of \texorpdfstring{quadratic}{quadratic} functions}
Some constraints in problem \eqref{dynopt_problem} are affine \eqref{eq_dynopt_effpos}-\eqref{eq_dynopt_cop}, \eqref{eq_dynopt_joint_torques} or second-order cones (SOC) \eqref{eq_dynopt_frccone}-\eqref{eq_dynopt_eff_length} and thus convex; others however describe nonconvex constraints such as the momentum dynamics evolution when considering the timestep variable $\indexed{\timeopt}[][\tid]$ as an optimization variable \eqref{eq_dynopt_momentum} or torque cross products \eqref{eq_dynopt_kappa}-\eqref{eq_dynopt_gamma}. Next, we show the common nature of all the nonlinearities and reformulate them in a way amenable to efficient approximations using iterative convex models.

The torque cross product $\effcomlen \times \forcevars$ between a length ($\indexed{\effpos}[][\effid] - \coms)$ in \eqref{eq_dynopt_kappa} or $\indexed{\effrot}[\xid,\yid][\effid] \indexed{\copvars}[][\effid]$ in \eqref{eq_dynopt_gamma}) and the force $\indexed{\forcevars}[][\effid,\tid]$ can be written as
%
% Cross product length and force
%
\begin{subequations}
\begin{align}
	%
	% General product
	%
	\effcomlen \times \forcevars & =\hspace{-0.1cm}
	\begin{bmatrix}
		\begin{array}{lll}
			\phantom{-}0                          & -\indexed{\effcomlen}[\zid]           & \phantom{-}\indexed{\effcomlen}[\yid] \\
			\phantom{-}\indexed{\effcomlen}[\zid] & \phantom{-}0                          & -\indexed{\effcomlen}[\xid]           \\
			-\indexed{\effcomlen}[\yid]           & \phantom{-}\indexed{\effcomlen}[\xid] & \phantom{-}0
		\end{array}
	\end{bmatrix}
	\begin{bmatrix}
		\indexed{\forcevars}[\xid] \\ \indexed{\forcevars}[\yid] \\ \indexed{\forcevars}[\zid]
    \end{bmatrix} \label{eq_len_cross_force_1} \\
    %
    % Decomposed product
    %
    & =\hspace{-0.1cm} \Bigg[ {
    	% ax
    	\overbrace{
    		\begin{bmatrix}
    			-\indexed{\effcomlen}[\zid] &\hspace{-0.18cm}  \indexed{\effcomlen}[\yid]
    		\end{bmatrix}}^{ \indexed{\avector}[\xid] }
    	% bx
    	\overbrace{
    		\begin{bmatrix}
    		     \indexed{\forcevars}[\yid] \\ \indexed{\forcevars}[\zid]
    		\end{bmatrix}}^{ \indexed{\bvector}[\xid] },
    	% ay
    	\overbrace{
    		\begin{bmatrix}
    		     \indexed{\effcomlen}[\zid] &\hspace{-0.18cm} -\indexed{\effcomlen}[\xid]
    		\end{bmatrix}}^{ \indexed{\avector}[\yid] }
    	% by
    	\overbrace{
    		\begin{bmatrix}
    			 \indexed{\forcevars}[\xid] \\ \indexed{\forcevars}[\zid]
    		\end{bmatrix}}^{ \indexed{\bvector}[\yid] },
    	% az
    	\overbrace{
    		\begin{bmatrix}
    			-\indexed{\effcomlen}[\yid] &\hspace{-0.18cm}  \indexed{\effcomlen}[\xid]
    		\end{bmatrix}}^{ \indexed{\avector}[\zid] }
    	% bz
    	\overbrace{
    		\begin{bmatrix}
    			 \indexed{\forcevars}[\xid] \\ \indexed{\forcevars}[\yid]
    		\end{bmatrix}}^{ \indexed{\bvector}[\zid] }
	} \Bigg] \label{eq_len_cross_force_2}
\end{align}
\end{subequations}
\noindent where the superscripts $\xid, \yid, \zid$ reference to the components of the vectors $\effcomlen, \forcevars \in \setreals^{3\times1}$, but then they also identify the vectors $\indexed{\avector}[\iid], \indexed{\bvector}[\iid] \in \setreals^{2\times1}$ for $\iid \in \{ \xid, \yid, \zid \}$, whose scalar product $\indexed{\avector}[\iid] \cdot \indexed{\bvector}[\iid]$ is equivalent to the corresponding element of the cross product vector $(\effcomlen \times \forcevars)^{\iid}$. Similarly, we notice that the nonconvexity in \eqref{eq_dynopt_momentum} can be written as a scalar product between the timestep variable $\indexed{\timeopt}[][\tid]$ and linear momentum $\indexed{\lmoms}[][\tid]$, contact forces $\indexed{\forcevars}[][\effid,\tid]$ and torque $\indexed{\effcross}[][\effid,\tid]$ variables. It means that all nonconvex constraints solely include equality constraints with bilinear terms.

Noticing that $\indexed{\avector}[\iid] \cdot \indexed{\bvector}[\iid] = \frac{1}{4}\norm{\indexed{\avector}[\iid] + \indexed{\bvector}[\iid]}^{2}_{2} - \frac{1}{4}\norm{\indexed{\avector}[\iid] - \indexed{\bvector}[\iid]}^{2}_{2}$,
we reformulate all the bilinear expressions as differences of convex quadratic functions with known positive curvature, as was done in \cite{Herzog-2016b} and in the spirit of \cite{DBLP:conf/cdc/ShenDGB16}. In other words, we can now decompose a bilinear expression with an indefinite curvature into quadratic terms with known curvature, which is key for the efficiency of our algorithm. To simplify the subsequent presentation, we define the following sets
\begin{definition}
Given a real vector space $\vecspace$, we define $\setp$ as the set of quadratic functions $\vecspace \rightarrow \setreals$ with a positive semi-definite Hessian matrix.
\end{definition}
\begin{definition} \label{def:decomposition}
	Given a real vector space $\vecspace$, the set $\setpm$ is
	%
	% Difference of quadratic functions
	%
	\begin{align}
		&\setpm \hspace{-0.025cm}=\hspace{-0.025cm} \bigg\{ \hspace{-0.025cm}\fbilinear{\cdot} : \vecspace \hspace{-0.025cm}\rightarrow\hspace{-0.025cm} \setreals \; | \; \fbilinear(\hspace{-0.025cm}\uvector\hspace{-0.025cm}) \hspace{-0.025cm}=\hspace{-0.025cm} \fdcdecompP(\hspace{-0.025cm}\uvector\hspace{-0.025cm}) \hspace{-0.025cm}-\hspace{-0.025cm} \fdcdecompN(\hspace{-0.025cm}\uvector\hspace{-0.025cm}) \textrm{ for } \fdcdecompP, \fdcdecompN \hspace{-0.025cm}\in\hspace{-0.025cm} \setp \hspace{-0.025cm} \bigg\}
	\end{align}
\end{definition}
\noindent where Figure \ref{fig2:dc_picture} graphically illustrates this decomposition.
%
%============= Function Decomposition ============%
\begin{figure}
\begin{minipage}{1.00\linewidth}
	\centering
	\includegraphics[width=1.00\linewidth]{\filename{figures/decomposition/Decomp}}
	\caption[]{\small Decomposition (as shown in \textit{Definition} \ref{def:decomposition}) of the bilinear form $\fbilinear(\uvector) = \fbilinear([{\indexed{\uvector}[][1]}^{T},{\indexed{\uvector}[][2]}^{T}]^{T}) = \indexed{\uvector}[][1] \cdot \indexed{\uvector}[][2]$ into a difference of quadratic expressions $\fbilinear(\uvector) = \fdcdecompP(\uvector) - \fdcdecompN(\uvector)$ with $\fdcdecompP(\uvector) = \frac{1}{4} \fquad (\indexed{\uvector}[][1]+\indexed{\uvector}[][2])$ and $\fdcdecompN(\uvector) = \frac{1}{4}\fquad(\indexed{\uvector}[][1]-\indexed{\uvector}[][2])$, where $\fquad(\cdot)$ is the quadratic function $\norm{\cdot}^{2}_{2}$.}
	\label{fig2:dc_picture}
\end{minipage}
\end{figure}
%=================================================%
%
\noindent In particular, the set $\setpm$ is closed under scalar multiplication, addition and composition with affine functions,
%
% Group closed under scalar multiplication, addition and composition
%
\begin{align}
	\alpha (\mathbf{\bar{v}} \circ \mathbf{v}) + \beta (\mathbf{\bar{w}} \circ \mathbf{w}) \in \setpm \quad
% 	\begin{matrix} \ctxt{\forall \mathbf{\bar{v}}(\cdot), \mathbf{\bar{w}}(\cdot)} \in \setpm;\\
% 	\mathbf{v}(\cdot), \mathbf{w}(\cdot) \in \pazocal{A}; \quad \ctxt{\alpha}, \beta \in \setreals \end{matrix}
\end{align}
\noindent for any $\alpha, \beta \in \setreals$, affine functions $\mathbf{v}(\cdot),\mathbf{w}(\cdot)$ and $\mathbf{\bar{v}}(\cdot), \mathbf{\bar{w}}(\cdot) \in \setpm$.
Consider for example Equation \eqref{eq_dynopt_gamma} and assume for simplicity that $(\indexed{\effrot}[\xid,\yid][\effid,\tid]   \indexed{\copvars}[][\effid,\tid]) \times \indexed{\forcevars}[][\effid,\tid]$ is represented by the decomposition $\effcomlen \times \forcevars$, then each endeffector torque component becomes $\indexed{\efftrq}[\iid][\effid,\tid] = \indexed{\avector}[\iid] \cdot \indexed{\bvector}[\iid] + ( \indexed{\effrot}[\zid][\effid,\tid] \indexed{\coptrq}[][\effid,\tid] )^{\iid}$. The torque component $\indexed{\coptrq}[][\effid,\tid]$ could also be formulated as a difference of positive components $\indexed{\coptrq} = \indexed[+]{\coptrq} - \indexed[-]{\coptrq}, \textrm{ where } \indexed[+]{\coptrq},\indexed[-]{\coptrq} \ge 0 $, as in \cite{DBLP:conf/wafr/PosaT12} to embed them into the decomposition; however, this is not required. Then
%
% Decomposition of endeffector torque
%
\begin{equation}
\indexed{\efftrq}[\iid][\effid,\tid] =
\overbrace{
	%
	% Positive term of endeffector torque decomposition
	%
	\underbrace{\left[
		\frac{1}{4} \norm{\indexed{\avector}[\iid] + \indexed{\bvector}[\iid]}^{2}_{2}
		\right]}_{\in \setp}
	-
	%
	% Negative term of endeffector torque decomposition
	%
	\underbrace{ \left[
		\frac{1}{4} \norm{\indexed{\avector}[\iid] - \indexed{\bvector}[\iid]}^{2}_{2}
		\right] }_{\in \setp}
}^{\in \setpm}
	%
	% Rotational part
	%
	+ (\indexed{\effrot}[\zid][\effid,\tid])^{\iid} \indexed[]{\coptrq}[][\effid,\tid]
\label{eq_gamma_decomposition}
\end{equation}
In a similar manner, each endeffector torque component $\indexed{\effcross}[\iid][\effid,\tid]$ \eqref{eq_dynopt_kappa}, can be decomposed parameterizing its cross product $(\indexed{\effpos}[][\effid] - \coms) \times \indexed{\forcevars}[][\effid]$ with vectors $\indexed{\cvector}[\iid]$ and $\indexed{\dvector}[\iid]$ as $\indexed{\effcross}[\iid][\effid,\tid] = \indexed{\cvector}[\iid] \cdot \indexed{\dvector}[\iid] + \indexed{\efftrq}[\iid][\effid,\tid]$.
%
% Decomposition of vector ci and di
%
\begin{equation}
\indexed{\effcross}[\iid][\effid,\tid] =
\overbrace{
	\overbrace{
		%
		% Positive term of endeffector torque decomposition
		%
		\underbrace{\left[
			\frac{1}{4} \norm{\indexed{\cvector}[\iid] + \indexed{\dvector}[\iid]}^{2}_{2}
			\right]}_{\in \setp}
		-
		%
		% Negative term of endeffector torque decomposition
		%
		\underbrace{ \left[
			\frac{1}{4} \norm{\indexed{\cvector}[\iid] - \indexed{\dvector}[\iid]}^{2}_{2}
			\right] }_{\in \setp}
	}^{\in \setpm} +
	\overbrace{ \indexed{\efftrq}[\iid][\effid,\tid] }^{\in \setpm}
}^{\in \setpm}
\label{eq_kappa_decomposition}
\end{equation}
\noindent where the vectors $\indexed{\cvector}[\iid], \indexed{\dvector}[\iid] \in \setreals^{2\times1}$ for $\iid \in \{ \xid, \yid, \zid \}$ have been introduced in a similar fashion to Eq. \eqref{eq_len_cross_force_2} to refer to the vectors whose scalar product $\indexed{\cvector}[\iid] \cdot \indexed{\dvector}[\iid]$ is equivalent to the corresponding component of the cross product $((\indexed{\effpos}[][\effid] - \coms) \times \indexed{\forcevars}[][\effid])^{\iid}$. A similar analysis holds for each of the Cartesian components of the bilinear expressions within the dynamic constraints \eqref{eq_dynopt_momentum}, which can be decomposed into elements of $\setpm$ as given by
%
% Decomposition of other terms in centroidal dynamics
%
\begin{subequations}
	\begin{align}
		%
		% linear momentum
		%
		\indexed{\lmoms}[\iid][\tid] \indexed{\timeopt}[][\tid] =&
		 \frac{1}{4} \norm{\indexed{\lmoms}[\iid][\tid] + \indexed{\timeopt}[][\tid]}^{2}_{2}
		-\frac{1}{4} \norm{\indexed{\lmoms}[\iid][\tid] - \indexed{\timeopt}[][\tid]}^{2}_{2} \\
		%
		% wrenches
		%
		\sum_{\effid \in \setacteff} \hspace{-0.15cm}\indexed{\effcross}[\iid][\effid,\tid] \indexed{\timeopt}[][\tid] =&
		\frac{1}{4} \norm{ \sum_{\effid \in \setacteff} \hspace{-0.15cm}\indexed{\effcross}[\iid][\effid,\tid] + \indexed{\timeopt}[][\tid]}^{2}_{2}
		-\frac{1}{4} \norm{ \sum_{\effid \in \setacteff} \hspace{-0.15cm}\indexed{\effcross}[\iid][\effid,\tid] - \indexed{\timeopt}[][\tid]}^{2}_{2} \\
		%
		% forces
		%
		\sum_{\effid \in \setacteff} \hspace{-0.15cm}\indexed{\forcevars}[\iid][\effid,\tid] \indexed{\timeopt}[][\tid] =&
		\frac{1}{4} \norm{ \sum_{\effid \in \setacteff} \hspace{-0.15cm}\indexed{\forcevars}[\iid][\effid,\tid] + \indexed{\timeopt}[][\tid]}^{2}_{2}
		-\frac{1}{4} \norm{ \sum_{\effid \in \setacteff} \hspace{-0.15cm}\indexed{\forcevars}[\iid][\effid,\tid] - \indexed{\timeopt}[][\tid]}^{2}_{2}
	\end{align}
	\label{eq_dc_dynamics}
\end{subequations}
In the next section, we show how we can use this structure to approximate the problem using iterative convex relaxations.
%
%%%%%%%%%%%%%%%%%%%%%%%%%%%%%%%%%%%%%%%%%%%%%%%%%%
% Optimization with iterative convex relaxations %
%%%%%%%%%%%%%%%%%%%%%%%%%%%%%%%%%%%%%%%%%%%%%%%%%%
\subsection{Optimization with iterative convex relaxations} \label{subsubsec:iterative_methods}
We now use the known curvature of the quadratic terms $\setp$ to build a convex approximation. We start by isolating the quadratic expressions into quadratic constraints by introducing scalar variables $\indexed{\ascalar}[\iid], \indexed{\bscalar}[\iid] \in \setreals$. For example, Eq. \eqref{eq_gamma_decomposition} would become
%
% Bilinear terms as quadratic equality constraints
%
\begin{subequations}
	\begin{align}
		%
		% Quadratic equalities
		%
		&\indexed{\ascalar}[\iid] = \norm{\indexed{\avector}[\iid] + \indexed{\bvector}[\iid]}^{2}_{2} \; ,  \quad \indexed{\bscalar}[\iid] = \norm{\indexed{\avector}[\iid] - \indexed{\bvector}[\iid]}^{2}_{2} \label{eq_chg_variables} \\
		%
		% Additional linear constraints
		%
		&\quad\indexed{\efftrq}[\iid][\effid,\tid] =
			\frac{1}{4} \left(
			 \indexed{\ascalar}[\iid] - \indexed{\bscalar}[\iid]
			\right)
			+ (\indexed{\effrot}[\zid][\effid,\tid])^{\iid} \indexed[]{\coptrq}[][\effid,\tid]
			\label{eq_linear_part}
	\end{align}
\end{subequations}
\noindent where the introduction of the additional scalar variables $\indexed{\ascalar}[\iid], \indexed{\bscalar}[\iid]$ renders the original equation \eqref{eq_linear_part} linear and isolates the quadratic nonconvex expressions with known curvature into a pair of additional quadratic constraints \eqref{eq_chg_variables}, whose very simple form will benefit the search of efficient convex approximations.%\ctxt{, and thus makes up for the problem size increase}.

%============ Function Approximations ============%
\begin{figure}
	\centering
	\begin{subfigure}[b]{0.48\textwidth}   
		\centering
		\includegraphics[width=0.90\linewidth]{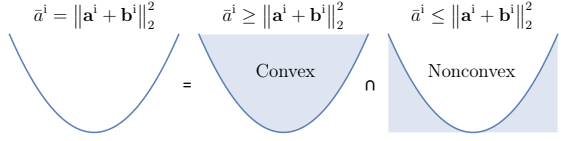}
		\caption[]{\small Non-convex quadratic equality constraint as the intersection of a convex and a nonconvex quadratic inequality constraints. In this work, we use only the convex space of this constraint and a heuristic to guide solutions towards its boundary.}
		\label{fig3:intersection_approx}
		\includegraphics[width=0.90\linewidth]{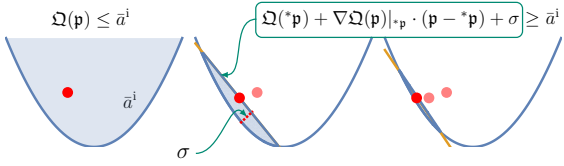}
		\caption[]{\small Trust-region method: We first find a solution within the convex space $\fquad(\pvector) \leq \indexed{\ascalar}[\iid]$ (our approximation variable $\indexed{\ascalar}[\iid]$ can take any value within the blue region). Then based on this solution, we iteratively build a trust-region that limits the search space to the boundaries. The parameter $\trpenalty$ controls the distance between trust-region and quadratic constraint, thus the amount of constraint violation.}
		\label{fig3:trust_region_approx}
		\includegraphics[width=0.90\linewidth]{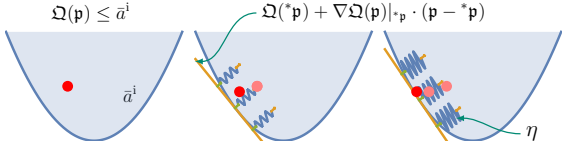}
		\caption[]{\small Soft-constraint method: We first find a solution within the convex space $\fquad(\pvector) \leq \indexed{\ascalar}[\iid]$ and based on this solution, we iteratively build a function underestimator, that allows us to include a cost that rewards selecting values close to it and thus close to the constraint boundary. Parameter $\scpenalty$ controls the desirability of selecting solutions close to the underestimator of the convex quadratic inequality constraint.}
		\label{fig3:soft_constraint_approx}
	\end{subfigure}
	\caption[]
	{\small Sequential approximation of quadratic expressions $\setp$ within its convex space using iterative convex relaxation methods.} 
	\label{fig3:quadractic_equality_constraint}
\end{figure}
%=================================================%
%
Figure \ref{fig3:intersection_approx} sketches the hyperplane defined by the nonconvex constraint \eqref{eq_chg_variables}, conceived as the intersection of two inequalities, a convex $\indexed{\ascalar}[\iid] \geq \norm{\indexed{\avector}[\iid] + \indexed{\bvector}[\iid]}^{2}_{2}$ and a concave one $\indexed{\ascalar}[\iid] \leq \norm{\indexed{\avector}[\iid] + \indexed{\bvector}[\iid]}^{2}_{2}$. While it is difficult to search a solution in a high dimensional nonconvex space, it is easier to search within the space defined by the convex inequality and guide the optimization towards the constraint boundary, approaching in this way towards solutions with practical feasibility for the original nonconvex quadratic equality constraint.

To summarize, we systematically isolate all the quadratic expressions present in the optimization problem and replace them with new scalar optimization variables in order to render the original constraints linear. We then add simple equality constraints between the new variables and the quadratic terms. This allows us to move all the nonconvex elements of the problem into simpler terms in the form of quadratic equality constraints. We now propose two iterative methods based on SOC programs to deal with each of the quadratic equalities.
\subsubsection{Trust-region method}
In this approach the main idea is to use a primal constraint to limit the convex search space to values close to the boundaries. In mathematical terms, the trust-region should constrain the problem to values of $\indexed{\ascalar}[\iid]$ near $\fquad(\pvector)$  (for simplicity of notation, we define $\pvector = \indexed{\avector}[\iid] + \indexed{\bvector}[\iid]$ and $\fquad(\cdot) = \norm{\cdot}^{2}_{2}$). During the first iteration, an initial guess of the optimal problem values is obtained by searching over the entire relaxed convex search space. From there on, the trust-region is built based on the optimal problem values from the previous iteration $\indexed[*]{\pvector}$ and by reducing the desired allowed amount of constraint violation $\trpenalty$, as shown in Figure \ref{fig3:trust_region_approx}.
\begin{mytheorem1*}
	In the case of $\setp$ expressions, thanks to the positive curvature of the constraint's hessian, a linear inequality constraint suffices to constrain the problem as desired.
	\begin{equation}
		\hspace{-0.05cm}\fquad(\pvector) \hspace{-0.05cm}=\hspace{-0.05cm} \indexed{\ascalar}[\iid] \hspace{-0.05cm}\rightarrow\hspace{-0.075cm}
		\begin{cases}
			\hspace{-0.025cm}\fquad(\indexed[*]{\pvector}\hspace{-0.025cm}) \hspace{-0.025cm}+\hspace{-0.05cm} \nabla \fquad(\hspace{-0.025cm}\pvector\hspace{-0.025cm}) |_{\indexed[*]{\pvector}} \hspace{-0.025cm}\cdot\hspace{-0.025cm} (\pvector \hspace{-0.025cm}-\hspace{-0.05cm} \indexed[*]{\pvector} \hspace{-0.025cm}) \hspace{-0.025cm}+\hspace{-0.05cm} \trpenalty \hspace{-0.05cm}\geq\hspace{-0.025cm} \indexed{\ascalar}[\iid]
		\end{cases}
	\end{equation}
	The linear constraint is built based on the optimal values of $\pvector$ found in the previous iteration $\indexed[*]{\pvector}$ and $\trpenalty$ is a positive threshold, big enough to provide a feasible interior to the intersection of the constraints, but also small enough so as to achieve the desired precision at convergence.
\end{mytheorem1*}
The benefits of constraining the problem in this way are twofold: firstly, we can easily refine the solution with values of $\pvector$ around $\indexed[*]{\pvector}$ that satisfy the amount of desired constraint violation $\trpenalty$, and secondly, it provides a method to iteratively increase the approximation accuracy by reducing the value of $\trpenalty$, as required by convergence tolerances. We further note that if the hessian of this constraint were an indefinite matrix, this trust-region would lead to unbounded regions instead of constraining the problem as desired.
\subsubsection{Soft-constraint method}
Alternatively, a hard restriction of the search space could be replaced with a cost that biases the optimizer towards finding solutions close to the boundary of the constraint by pulling optimization variables towards a function underestimator, as shown in Figure  \ref{fig3:soft_constraint_approx}.
\begin{mytheorem2*}
	A cost heuristic is used to reward the selection of values for the variable $\indexed{\ascalar}[\iid]$ close to the function underestimator ($ \fquad(\indexed[*]{\pvector}) + \nabla \fquad(\pvector) |_{\indexed[*]{\pvector}} \cdot (\pvector - \indexed[*]{\pvector})$), hyperplane that supports the function and was built based the optimal values of $\pvector$ found in the previous iteration $\indexed[*]{\pvector}$.
	\begin{equation}
		\fquad(\pvector) \hspace{-0.05cm}=\hspace{-0.05cm} \indexed{\ascalar}[\iid] \hspace{-0.05cm}\rightarrow\hspace{-0.05cm}
		\begin{cases}
			\fquad(\pvector) \leq \indexed{\ascalar}[\iid] \\
			\scpenalty \hspace{-0.025cm}\norm{ 
				\fquad(\hspace{-0.025cm}\indexed[*]{\pvector}\hspace{-0.025cm}) \hspace{-0.05cm}+\hspace{-0.05cm} \nabla \hspace{-0.025cm} \fquad(\hspace{-0.025cm}\pvector\hspace{-0.025cm}) \hspace{-0.025cm}|\hspace{-0.025cm}_{\indexed[*]{\pvector}} \hspace{-0.05cm}\cdot\hspace{-0.05cm} (\pvector \hspace{-0.05cm}-\hspace{-0.075cm} \indexed[*]{\pvector}\hspace{-0.025cm}) \hspace{-0.05cm}-\hspace{-0.05cm} \indexed{\ascalar}[\iid] \hspace{-0.025cm}
			}^{2}_{2}
		\end{cases}
	\end{equation}
	$\scpenalty$ defines the desirability of selecting optimization values close to the underestimator, and thus enjoy practical feasibility for the nonconvex constraint. 
\end{mytheorem2*}
\begin{remark}
	As shown in Fig. \ref{fig3:quadractic_equality_constraint}, both methods iteratively approximate the problem as SOC programs, efficiently solvable with polynomial-time methods. In section sec. \ref{exp:approximation_limitations}, we will further discuss and compare the described methods.
\end{remark}
\subsection{Numerical optimization}
In this section, we describe numerical aspects such as convergence criteria and algorithmic implementation details for both optimization problems. 
\subsubsection{Convergence criteria}
The amount of constraint violation $\converr$ is used as the measure to decide upon convergence. It is defined as the supremum among the average errors of the state trajectory variables \eqref{eq_error}, which are computed by comparing the values of the optimization variables ($\indexed{\coms}[][\tid]$, $\indexed{\lmoms}[][\tid]$, $\indexed{\amoms}[][\tid]$) that solve the approximate problem and the values obtained by integrating endeffector wrenches ($\indexed[seq]{\coms}[][\tid]$, $\indexed[seq]{\lmoms}[][\tid]$, $\indexed[seq]{\amoms}[][\tid]$) that satisfy exactly all of the nonconvex constraints, as follows
%
% Convergence criteria
%
\begin{subequations}
\begin{align}
	%
	% Linear momentum in sequential form
	%
	& \indexed[seq]{\lmoms}[][\tid] \hspace{-0.05cm}=\hspace{-0.05cm} \indexed{\lmoms}[][0] \hspace{-0.025cm}+\hspace{-0.025cm} \sum_{\mathrm{\iid=1}}^{\tid}\hspace{-0.05cm} \left[ \robotmass \gravity + \sum\limits_{\effid \in \setacteff}\hspace{-0.075cm} \indexed{\forcevars}[][\effid,\iid] \right] \indexed{\timeopt}[][\iid] \\
	%
	% Center of mass in sequential form
	%
	& \indexed[seq]{\coms}[][\tid] \hspace{-0.05cm}=\hspace{-0.05cm} \indexed{\coms}[][0] \hspace{-0.025cm}+\hspace{-0.025cm} \frac{1}{\robotmass} \sum_{\mathrm{\jid=1}}^{\tid}\hspace{-0.05cm} \left[ \indexed{\lmoms}[][0] + \sum_{\mathrm{\iid=1}}^{\jid}\hspace{-0.05cm} \left( \robotmass \gravity + \sum\limits_{\effid \in \setacteff}\hspace{-0.075cm} \indexed{\forcevars}[][\effid,\iid] \right) \indexed{\timeopt}[][\iid] \right] \indexed{\timeopt}[][\jid] \\
	%
	% Angular momentum in sequential form
	%
	& \indexed[seq]{\amoms}[][\tid] \hspace{-0.05cm}=\hspace{-0.075cm} \indexed{\amoms}[][0] \hspace{-0.05cm}+\hspace{-0.075cm} \sum_{\mathrm{\iid=1}}^{\tid}\hspace{-0.1cm} \left[ \hspace{-0.05cm} \sum\limits_{\effid \in \setacteff}\hspace{-0.175cm} (\indexed{\effpos}[][\effid,\iid] \hspace{-0.05cm}+\hspace{-0.05cm} \indexed{\effrot}[\xid,\yid][\effid,\iid] \hspace{-0.025cm}\indexed{\copvars}[][\effid,\iid] \hspace{-0.075cm}-\hspace{-0.1cm} \indexed[seq]{\coms}[][\iid]) \hspace{-0.075cm}\times\hspace{-0.075cm} \indexed{\forcevars}[][\effid,\iid] \hspace{-0.05cm}+\hspace{-0.05cm} \indexed{\effrot}[\zid][\effid,\iid] \hspace{-0.025cm} \indexed{\coptrq}[][\effid,\iid] \hspace{-0.05cm}\right] \hspace{-0.1cm}\indexed{\timeopt}[][\iid] \\
	%
	% Maximum convergence error
	%
	& \converr \hspace{-0.05cm}=\hspace{-0.05cm} \sup \Bigg\{
	\underbrace{\sum_{\tid=1}^{\dishorizon} \frac{\norm{ \coms_{\tid}  - \indexed[seq]{\coms}[][\tid] }^{2}_{2}}{\dishorizon}}_{\indexed{\converr}[][\coms]},
	\underbrace{\sum_{\tid=1}^{\dishorizon} \frac{\norm{ \lmoms_{\tid} - \indexed[seq]{\lmoms}[][\tid] }^{2}_{2}}{\dishorizon}}_{\indexed{\converr}[][\lmoms]},
	\underbrace{\sum_{\tid=1}^{\dishorizon} \frac{\norm{ \amoms_{\tid} - \indexed[seq]{\amoms}[][\tid] }^{2}_{2}}{\dishorizon}}_{\indexed{\converr}[][\amoms]} \Bigg\} \label{eq_error}
\end{align}
\end{subequations}	

When the errors $\converr$ fall below a certain threshold for the constraint violation to be considered negligible for practical purposes, we consider that the algorithm has converged.
\subsubsection{Algorithmic implementation details}
To approximate the solution of problem \eqref{dynopt_problem}, we  iteratively solve an approximate problem (using an interior point solver for SOC programs based on \cite{Domahidi2013ecos}), where each nonconvex constraint \eqref{eq_dynopt_momentum}-\eqref{eq_dynopt_gamma} has been replaced by a convex approximation. At each iteration, we update the approximation (based on the optimal values of the previous iteration) and its parameters to reduce the constraint violation amount. The procedure is then repeated until convergence. For the trust-region method, the parameter $\trpenalty$ is decreased using iteratively increasing powers of a value less than one, i.e.  $\trpenalty \propto {{\nu}^k}$, where ${\nu}<1.0$ and ${k}$ denotes the iteration number. {In a similar fashion, for the soft-constraint method, a value for the penalty parameter $\scpenalty$ is selected according to the desired precision to be achieved (typically within the range $[1e4, 1e6]$) and higher relative to other objectives, so that it is prioritized.}

{We also highlight that the formulation of torques $\indexed{\efftrq}[][\effid,\tid]$ in Eq. \eqref{eq_dynopt_gamma} separately of $\indexed{\effcross}[][\effid,\tid]$ in Eq. \eqref{eq_dynopt_kappa} is required only when the torque limits constraint \eqref{eq_dynopt_joint_torques} is used, as it depends on the contact wrench $\indexed{\wrench}[][\effid,\tid] = \begin{bmatrix} {\forcevars}^{T}_{\effid,\tid} & \hspace{-0.2cm} {\efftrq}^{T}_{\effid,\tid} \end{bmatrix}^{T}$. Otherwise, the torques $\indexed{\efftrq}[][\effid,\tid]$ in Eq. \eqref{eq_dynopt_gamma} can be directly embedded within the torque $\indexed{\effcross}[][\effid,\tid]$ in Eq. \eqref{eq_dynopt_kappa}, thus generating a problem of smaller size.}

%%%%%%%%%%%%%%%%%%%%%%%%%%%%%%%%%
% Optimization of Contact Plans %
%%%%%%%%%%%%%%%%%%%%%%%%%%%%%%%%%
\section{Optimization of contact plans} \label{sec:contacts_planning}
In this section, we explain how contact locations can be optimized within problem \eqref{dynopt_problem} when they are considered optimization variables {that belong to a given contact surface}. We also describe an algorithm based on mixed-integer programming to efficiently select a sequence of terrain surfaces and contact locations consistent with the {centroidal} dynamics.
\subsection{Membership of contact locations to terrain surfaces} \label{sec:contact_membership}
%
%================ Contact Surface ================%
\begin{figure}
	\centering
	\includegraphics[width=0.22\textwidth]{\filename{figures/surfnotation/CntSurfNotation}}
	\caption[]{\small The description of a terrain surface ${\indexed{\surface}[][\rid]}$ comprises a set of coplanar corners ${\indexed{\surfpoint}[\iid][\rid] \in \setreals^{3\times1}}$, where {in this case} $\iid \in [1,4]$. {Out of them} the following quantities can be computed: surface normal ${\indexed{\surfnormal}[][\rid] \in \setreals^{3\times1}}$, surface rotation {$\frotation(\indexed{\surface}[][\rid]) \in \setreals^{3\times3}$ (whose third column points in the direction of the surface normal)}, any surface point ${\indexed[surf]{\surfpoint}[][\rid] = \indexed{\surfpoint}[\iid][\rid]}$ and a membership constraint ${\bar{\surfpoint} \in \membership(\indexed{\surface}[][\rid]), \forall \bar{\surfpoint} \in \indexed{\surface}[][\rid]}$, that simply defines the set of points ${\bar{\surfpoint} \in \setreals^{3\times1}}$ that lie on the terrain surface.}
	\label{fig4:contact_surface}
\end{figure}
%=================================================%
%
Given a description of the terrain surface {$\indexed{\surface}[][\rid]$} (over which it is safe to make contact), a contact location can be optimized by including its membership constraint to surface {$\indexed{\surface}[][\rid]$} to the optimization problem. A terrain surface {$\indexed{\surface}[][\rid]$} (as defined in Fig. \ref{fig4:contact_surface}) is such that any contact point $\indexed{\effpos}[][\effid]$, selected from its interior, guarantees that the entire endeffector is in contact. The expression $\indexed{\effpos}[][\effid] \in {\membership(\indexed{\surface}[][\rid])}$ that constrains an endeffector position $\indexed{\effpos}[][\effid]$ to belong to surface {$\indexed{\surface}[][\rid]$} {is defined as follows}
%
% Endeffector membership to contact surface
%
\begin{equation}
	\indexed{\effpos}[][\effid] \in {\membership(\indexed{\surface}[][\rid])} {\hspace{0.2cm}\defequal\hspace{0.2cm}}
	%
	% Left hand side membership constraint
	%
	\begin{bmatrix}
		\begin{array}{r}
			{\indexed{\surfmat}[][\rid]} \\
			{\indexed{\surfnormal}[][\rid]} \\
			-{\indexed{\surfnormal}[][\rid]}
		\end{array}
	\end{bmatrix} \indexed{\effpos}[][\effid] \leq
	%
	% Right hand side membership constraint
	%
	\begin{bmatrix}
		\begin{array}{r}
			{\indexed{\surfvec}[][\rid]} \\
			{\indexed{\surfnormal}[][\rid] \cdot \indexed[surf]{\surfpoint}[][\rid]} \\
			-{\indexed{\surfnormal}[][\rid] \cdot \indexed[surf]{\surfpoint}[][\rid]}
		\end{array}
	\end{bmatrix}
	\label{eq_belonging_to_surface}
\end{equation}
\noindent Equation \eqref{eq_belonging_to_surface} defines a set of halfspaces, whose intersection constrains a contact point $\indexed{\effpos}[][\effid]$ to lie on a safe contact surface. For instance, ${\indexed{\surfmat}[][\rid]} \indexed{\effpos}[][\effid] \leq {\indexed{\surfvec}[][\rid]}$ denote the halfspaces that define lateral limits of the terrain surface, while ${\indexed{\surfnormal}[][\rid]} \cdot \indexed{\effpos}[][\effid] = {\indexed{\surfnormal}[][\rid] \cdot \indexed[surf]{\surfpoint}[][\rid]}$ implies that the normal distance from the plane should be zero, i.e. the contact point has to lie on the terrain surface. {Note that the row-size of the matrix $\indexed{\surfmat}[][\rid]$ and vector $\indexed{\surfvec}[][\rid]$ depends on the number of halfspaces required to define the terrain region $\indexed{\surface}[][\rid]$, while the column size of the matrix $\indexed{\surfmat}[][\rid]$ is as $\indexed{\effpos}[][\effid]$, namely 3.}
\subsection{Dynamics-based contacts planning}
Thus far, we have assumed that to solve problem \eqref{dynopt_problem} a {set of terrain surfaces, from where contacts are selected}, was given. {Alternatively,} a contact sequence could also be given by for example a contact planner such as \cite{Tonneau:2018dm,lin_efficient_2019}. In the following, we propose a mixed-integer formulation that enables {the selection of terrain surfaces} and contact sequences {based on a measure of dynamical robustness}.

\subsubsection{Terrain description and contact model}
We now describe how a terrain is modeled and how contacts are selected within this description {of the terrain} using the notation of \cite{DBLP:conf/humanoids/DeitsT14}.

The terrain consists of a set of $\numsurf$ convex, obstacle free regions {$\indexed{\surface}[][\rid]$ where $\rid \in \left\{ 1,\cdots,\numsurf \right\}$} and we consider {the selection of} a sequence of {$\numcntsopt$ contact locations $\indexed{\effpos}[][\cid]$ where $\cid \in \left\{ 1,\cdots,\numcntsopt \right\}$}. {We note that} the mapping between index ${\cid}$ of {the selected} contact location ${\indexed{\effpos}[][\cid]}$ and, {endeffector $\effid$} and the range of timesteps $\tid$, in which endeffector location $\indexed{\effpos}[][\effid,\tid]$ is active, is predefined. {For instance, we could optimize $\numcntsopt=4$ contacts with $\numcntsopt/2$ contacts for each foot in a locomotion task, or we could optimize a larger number of contacts $\numcntsopt=6$, where the 2 additional contacts are free slots to select hand contacts. Note that stance and flight timings can later be changed within the dynamics problem. Also $\rid = \cntsmap(\effid,\tid)$ maps $\effid,\tid$ to surface $\rid$ chosen for contact $\cid$.}

The matrix of binary variables ${\surfselmat \in \{0,1\}^{(\numcntsopt-\indexed{\numcntsopt}[][0])\times\numsurf}}$ {(indexed by contact $\cid \in \{1,\cdots,\numcntsopt-\indexed{\numcntsopt}[][0]\}$ and terrain surface $\rid \in \{1,\cdots,\numsurf\}$)} defines the terrain surface ${\indexed{\surface}[][\rid]}$, whose domain contains the contact location {$\indexed{\effpos}[][\cid]$} ({$\indexed{\numcntsopt}[][0]$} are contacts initially active and thus {with} a predefined pose). {The model is defined as follows}
%
% Description of contact constraints
%
\begin{subequations}
	\begin{align}
		%
		% surface selection matrix
		%
		&{\indexed{\surfselmat}[][\cid,\rid] \implies \indexed{\effpos}[][\cid] \in \membership(\indexed{\surface}[][\rid])} \label{eq_cntopt_cntassign} \\
		%
		% Fixed or not according to endeffector type
		%
		&{\sum_{\rid} \indexed{\surfselmat}[][\cid,\rid]}
		\begin{cases}
		= 1, \quad \textrm{for feet contacts} \\
		\leq 1, \quad \textrm{for hands contacts}
		\end{cases} \label{eq_cntopt_integrality} \\
		%
		% Force zeroing for non-active hand contacts
		%
		&1 - {\sum_{\rid} \indexed{\surfselmat}[][\cid,\rid]} \implies (\indexed{\forcevars}[][\effid,\tid] = 0), \quad \textrm{for hands} \label{eq_cntopt_nonactivefrcs}  \\[0.2em]
		%
		% Friction cones for active forces
		%
		&{\indexed{\surfselmat}[][\cid,\rid]} \implies {\indexed[cone]{\friccone}[][\friccoeff]} {\frotation(\indexed{\surface}[][\rid])} \indexed{\forcevars}[][\effid,\tid] \leq 0, \quad \textrm{friction cone} \label{eq:frccone_cntopt}
	\end{align}
\end{subequations}
\noindent{An element $\indexed{\surfselmat}[][\cid,\rid]$ being one implies the membership constraint $\indexed{\effpos}[][\cid] \in \membership(\indexed{\surface}[][\rid])$ as shown in Eq. \eqref{eq_cntopt_cntassign}.} Thus, $\indexed{\surfselmat}[][\cid,\rid]$ decides upon the terrain region from where a contact location can be selected. Integrality constraints \eqref{eq_cntopt_integrality} enforce membership of a contact location to at most one terrain surface. When no contact region is selected {(e.g. no hand contact)}, control variables such as contact forces should be inactive (Eq. \eqref{eq_cntopt_nonactivefrcs}). {When a contact region is selected,} local endeffector forces ${\frotation(\surface)^{T}} \indexed{\forcevars}[][\effid,\tid]$ must satisfy friction cone constraints, as in \eqref{eq:frccone_cntopt}. ${\indexed[cone]{\friccone}[][\friccoeff]}$ is a matrix function of $\friccoeff$ such that its product with the local force, {returns a vector of negative values}.

\subsubsection{Reachability constraints}
Reachability constraints between footstep locations are selected based on kinematic reachability using linear inequalities such as in \cite{khadiv:humanoids2016} for forward {or lateral} motions or, based on the intersection of SOC constraints \cite{DBLP:conf/humanoids/DeitsT14} for more general settings. They can be described in a convex form using linear inequalities based on kinematic reachability such as in
\begin{align}
	&\indexed[min]{\Delta\effpos} \leq {|} {\indexed{\effpos}[][\cid]} - {\indexed{\effpos}[][\cid-1]} {|} \leq \indexed[max]{\Delta\effpos}
\end{align}
\noindent {where two subsequent contacts are restricted to be within the bounds $\indexed[min]{\Delta\effpos}$ and $\indexed[max]{\Delta\effpos}$}. Reachability constraints can also be described as in \cite{DBLP:conf/humanoids/DeitsT14} using an intersection of SOC constraints
\begin{subequations}
	\begin{align}
		%
		% Only one active approximation
		%
		&{\sum\limits_{\hid \in \numpwappr} \indexed[sec]{\sinmat}[][\hid,\cid] = \sum\limits_{\hid \in \numpwappr} \indexed[sec]{\cosmat}[][\hid,\cid]} = 1 \label{eq_surf_integrality} \\	
		%
		% Sine matrix binary variable
		%
		&{\indexed[sec]{\sinmat}[][\hid,\cid] \implies
			\begin{cases}
				\indexed[sin]{\yawlimit}[][\hid] \leq \indexed{\yawangle}[][\cid] \leq \indexed[sin]{\yawlimit}[][\hid+1] \\
				\indexed[sin]{\sinapprox}[][\cid] = \indexed[sin]{\slopeyaw}[][\hid] \indexed{\yawangle}[][\cid] + \indexed[sin]{\interpyaw}[][\hid]
			\end{cases}} \label{eq_surf_sinmat} \\
		%
		% Cosine matrix binary variable
		%
		&{\indexed[sec]{\cosmat}[][\hid,\cid] \implies
			\begin{cases}
				\indexed[cos]{\yawlimit}[][\hid] \leq \indexed{\yawangle}[][\cid] \leq \indexed[cos]{\yawlimit}[][\hid+1] \\
				\indexed[cos]{\cosapprox}[][\cid] = \indexed[cos]{\slopeyaw}[][\hid] \indexed{\yawangle}[][\cid] + \indexed[cos]{\interpyaw}[][\hid]
			\end{cases}} \label{eq_surf_cosmat} \\
		%
		% Heuristic separation among endeffector locations
		%
		&\norm{ {
			\begin{bmatrix}
				\indexed{\effpos}[\xid][\cid] \\
				\indexed{\effpos}[\yid][\cid]
			\end{bmatrix}} \hspace{-0.05cm}-\hspace{-0.05cm}
			\left( {
				\begin{bmatrix}
					\indexed{\effpos}[\xid][\cid-1] \\
					\indexed{\effpos}[\yid][\cid-1]
				\end{bmatrix}} \hspace{-0.05cm}+\hspace{-0.05cm}
			{
				\begin{bmatrix}
					\indexed[cos]{\cosapprox}[][\cid] & \hspace{-0.15cm} -\indexed[sin]{\sinapprox}[][\cid] \\
					\indexed[sin]{\sinapprox}[][\cid] & \hspace{-0.15cm} \phantom{-} \indexed[cos]{\cosapprox}[][\cid]
				\end{bmatrix} \indexed{\focuspnts}[][1,2]}
			\right) } \leq {\indexed{\focusdist}[][1,2]} \label{eq_surf_separation}
	\end{align}
\end{subequations}

In the latter case e.g., a piecewise affine approximation of sine and cosine functions is used to model footsteps rotation ${\yawangle_{\cid} \in \setreals}$ in a convex form. {The matrices of} binary variables ${\indexed[sec]{\sinmat}, \indexed[sec]{\cosmat} \in \{0,1\}^{\numpwappr \times \indexed{\numcntsopt}[][f]}}$ {(indexed by affine approximation $\hid \in [1,\numpwappr]$ and contact $\cid \in [1, \indexed{\numcntsopt}[][f]]$ )} are used to select the active affine approximation of sine or cosine ${\hid}$ for each footstep ${\cid}$. {$\numpwappr$ denotes the number of affine functions used to approximate sine and cosine, and $\indexed{\numcntsopt}[][f]$ the number of footstep contacts to be selected out of the total number of contacts $\numcntsopt$}. As shown before, integrality constraints (Eq. \eqref{eq_surf_integrality}) guarantee that only one approximation is active at each footstep ${\cid}$.

{An element $\indexed[sec]{\sinmat}[][\hid,\cid], \indexed[sec]{\cosmat}[][\hid,\cid]$ being one implies the activation of a single affine approximation for sine and cosine functions, as shown in \eqref{eq_surf_sinmat}-\eqref{eq_surf_cosmat}.} Each affine approximation is defined by a region of validity of the footstep rotation angle ${\indexed{\yawangle}[][\cid] \in [\indexed[sin]{\yawlimit}[][\hid], \indexed[sin]{\yawlimit}[][\hid+1]]}$ (for sine) or ${\indexed{\yawangle}[][\cid] \in [\indexed[cos]{\yawlimit}[][\hid], \indexed[cos]{\yawlimit}[][\hid+1]]}$ (for cosine) and, the corresponding affine approximation ${\indexed[sin]{\sinapprox}[][\cid] = \indexed[sin]{\slopeyaw}[][\hid] \indexed{\yawangle}[][\cid] + \indexed[sin]{\interpyaw}[][\hid]}$ (for sine) or ${\indexed[cos]{\cosapprox}[][\cid] = \indexed[cos]{\slopeyaw}[][\hid] \indexed{\yawangle}[][\cid] + \indexed[cos]{\interpyaw}[][\hid]}$ (for cosine), where $\indexed[sin]{\slopeyaw}[][\hid], \indexed[sin]{\interpyaw}[][\hid], \indexed[cos]{\slopeyaw}[][\hid], \indexed[cos]{\interpyaw}[][\hid] \in \setreals$ are parameters that define slope and intercept values of each affine approximation. The footstep rotation angle $\indexed{\yawangle}[][\cid]$, sine $\indexed[sin]{\sinapprox}[][\cid]$ and cosine $\indexed[cos]{\sinapprox}[][\cid]$ of this angle constitute optimization variables. 

Finally, these variables are used to model the range of available positions for the next footstep (Eq. \eqref{eq_surf_separation}) based on the current footstep position and yaw angle as the intersection of two SOC constraints, parameterized by a pair of points $\indexed{\focuspnts}[][1,2] \in \setreals^{2\times1}$ (located sideways of the footstep position $\cid-1$ and rotated by the yaw angle), and a pair of distances $\indexed{\focusdist}[][1,2] \in \setreals$.
\subsubsection{Dynamics model and objective function}
To keep computational complexity low, in the mixed-integer approach to select contact sequences, we use a light version of problem \eqref{dynopt_problem}, where we do not consider the endeffector torques $\indexed{\efftrq}[][\effid,\tid]$ {(in other words, a point contact model is assumed)}, we use a linear approximation of the friction cones and, either a centroidal momentum dynamics model with fixed or non-fixed timings. The objective function $\indexed{\fcost}[cnt][\tid]$ similarly to \eqref{eq_dynopt_cost} regularizes states and controls and also incorporates user-defined objectives.
\subsubsection{Numerical optimization}
To evaluate the performance of our method at synthesizing contact plans and selecting contact surfaces, we implement a custom mixed-integer solver able to solve a sequence of SOC programs. It relies on two functions to bound the optimal value of a given search space. The lower bound comes from a relaxation of the search space binary variables and the upper bound by any solution where the binary variables are actually binary. The rest of the constraints are treated using the iterative models previously described. The feasible search space is partitioned into convex sets and each partition bounded. The algorithm converges once global lower and upper bounds are close enough, otherwise the partitions are refined and the search process is repeated. The implementation of the custom mixed-integer solver is based on a branch and bound method for global nonconvex optimization, as detailed in \cite{mixed_integer_solver}. In simple scenarios, we use linear reachability constraints, and SOC constraints in more complex ones, as will be shown in Section \ref{sec:experiments}.

%%%%%%%%%%%%%%%%%%%%%%%%
% Experimental Results %
%%%%%%%%%%%%%%%%%%%%%%%%
\section{Experimental Results} \label{sec:experiments}
\begin{figure}
	\begin{subfigure}[b]{0.49\linewidth}
		\centering
		\includegraphics[height=3cm, trim={19.0cm 16.0cm 21.0cm 10.0cm}, clip]{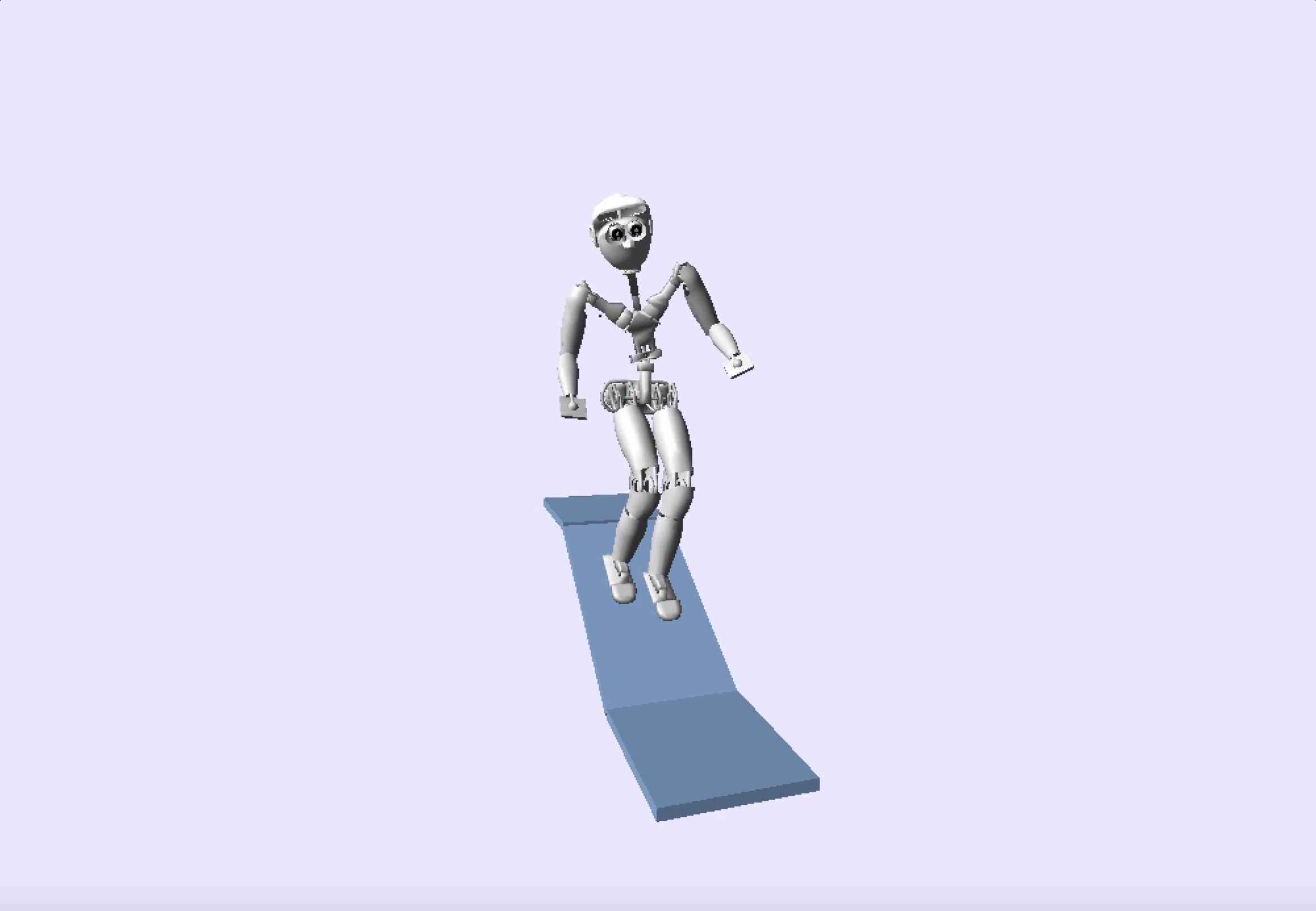}
		\caption[]{\small Simulated humanoid robot}
		\label{fig:humanoid_robot}
	\end{subfigure}
	\hfill
	\begin{subfigure}[b]{0.49\linewidth}
		\centering
		\includegraphics[height=3cm, trim={0.2cm 0.0cm 0.2cm 0cm},clip]{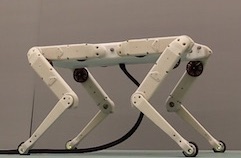}
		\caption[]{\small Real quadruped robot}
		\label{fig:quadruped_robot}
	\end{subfigure}
	\caption[]{\small {Robotic platforms used throughout the experimental section. Left, a simulated humanoid robot with 32 torque-controlled degrees of freedom is used in the SL simulation environment \cite{SLSimLab}. 
	It has 7 degrees of freedom in each limb and 3 in the torso. Right, we show the quadruped robot 'Solo' with 8 torque-controlled joints \cite{grimminger2019open}.}}
	\label{fig:robotic_platforms}
\end{figure}

In this section, we show experimental results about the optimization of contact and movement plans {using the algorithms previously described}. We have tested them in several challenging multi-contact scenarios
using simulated humanoid and quadruped robots and a real quadruped
robot (Fig. \ref{fig:humanoid_robot}).
%
% including walking on uneven terrain, climbing stairs using hands, {among others using a simulated humanoid robot (shown in Fig. \ref{fig:humanoid_robot}). We also present results on our real quadruped robot Solo (shown in Fig. \ref{fig:quadruped_robot})}. 
The resulting motions are visible in the accompanying video.

%============ Function Approximations ============%
%=================================================%
%
\subsection{On the optimization of movement plans} \label{exp:dynamics_planning} 
In this section, we analyze {solutions} of problem \eqref{dynopt_problem} in terms of convergence to feasibility (measured by the amount of constraint violation ${\converr}$ of the solution) and time complexity to converge to the desired feasibility threshold. We also present results regarding the qualitative improvement of motions that include time and/or contact locations in the optimization. Finally, we will show how full-body motions can be optimized using a kino-dynamic approach, how actuation limits can be included in the dynamics optimization, and tracking performance of time-optimized movement plans.
\subsubsection{Convergence to feasibility  and time complexity} \label{exp:feasibility_convergence}
\begin{figure}
	\centering
	\begin{subfigure}[b]{0.24\linewidth}   
		\centering
		\includegraphics[width=1.00\linewidth, trim={10cm 2.0cm 10cm 3.0cm}, clip]{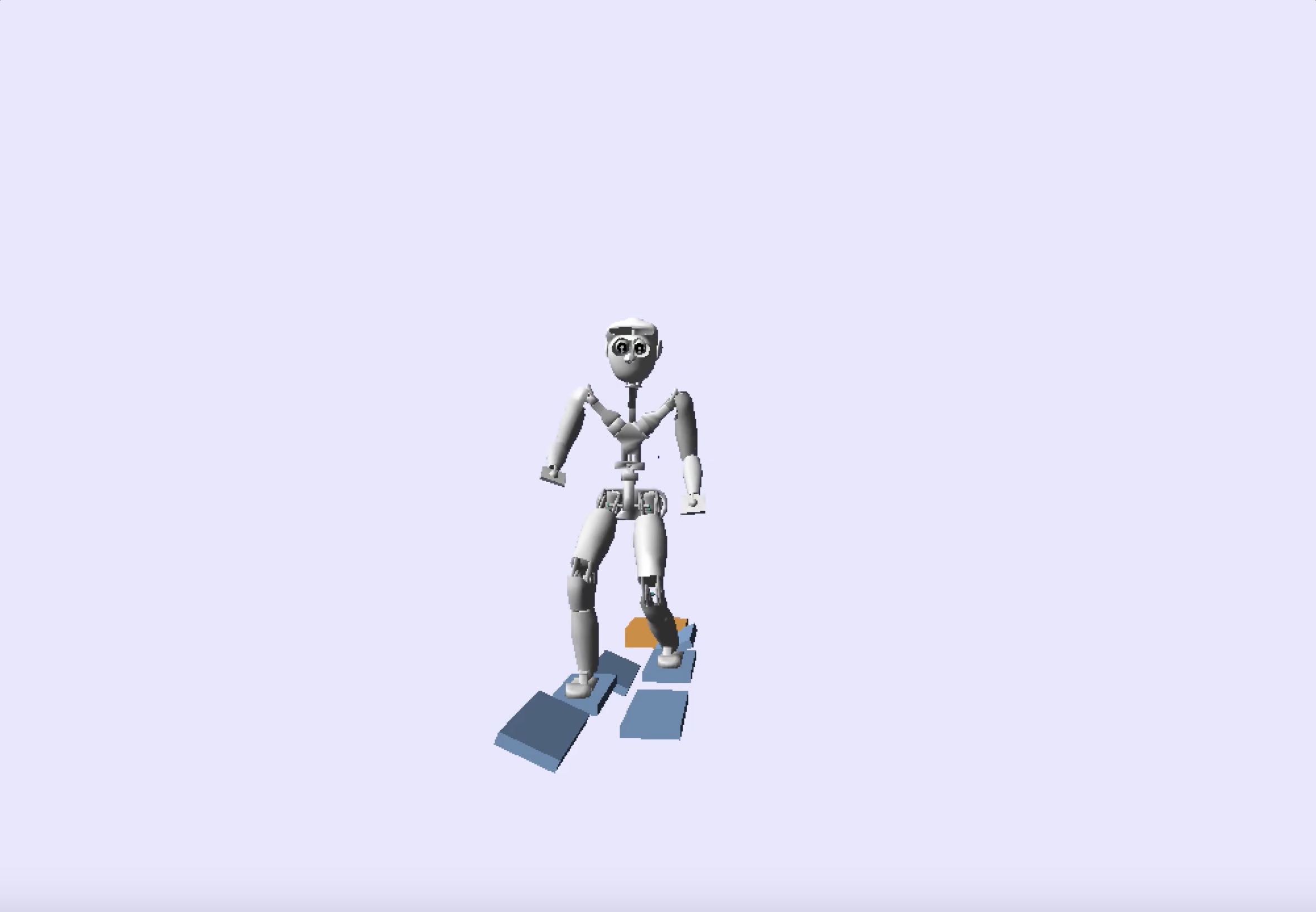}
		\caption[]{\footnotesize Rough terrain}
		\label{fig:motion01}
	\end{subfigure}
	\begin{subfigure}[b]{0.24\linewidth}   
		\centering
		\includegraphics[width=1.00\linewidth, trim={10cm 1.5cm 10cm 3.5cm}, clip]{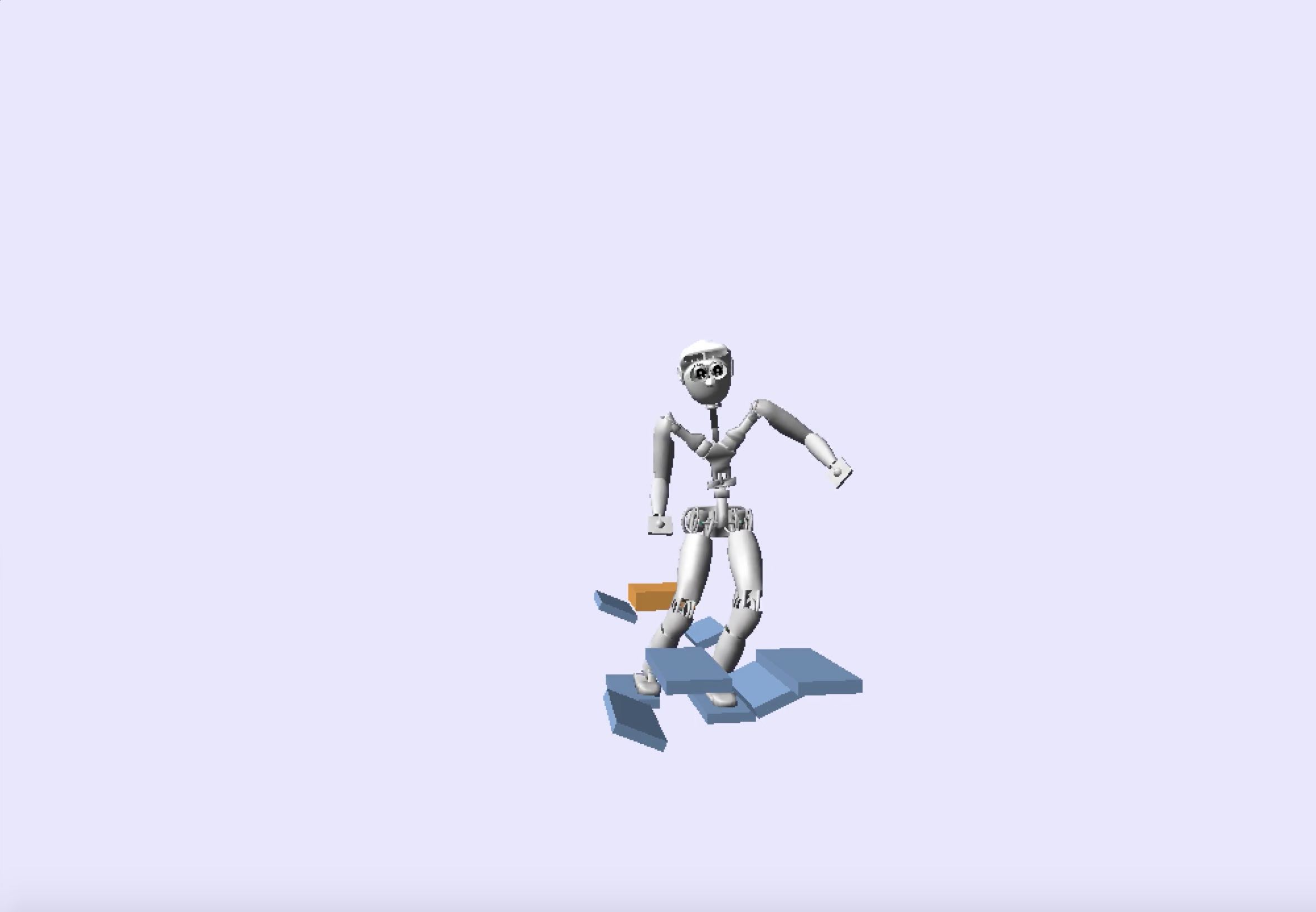}
		\caption[]{\footnotesize Down-Up}
		\label{fig:motion02}
	\end{subfigure}
	\begin{subfigure}[b]{0.24\linewidth}   
		\centering
		\includegraphics[width=1.00\linewidth, trim={10cm 2.0cm 10cm 3.0cm}, clip]{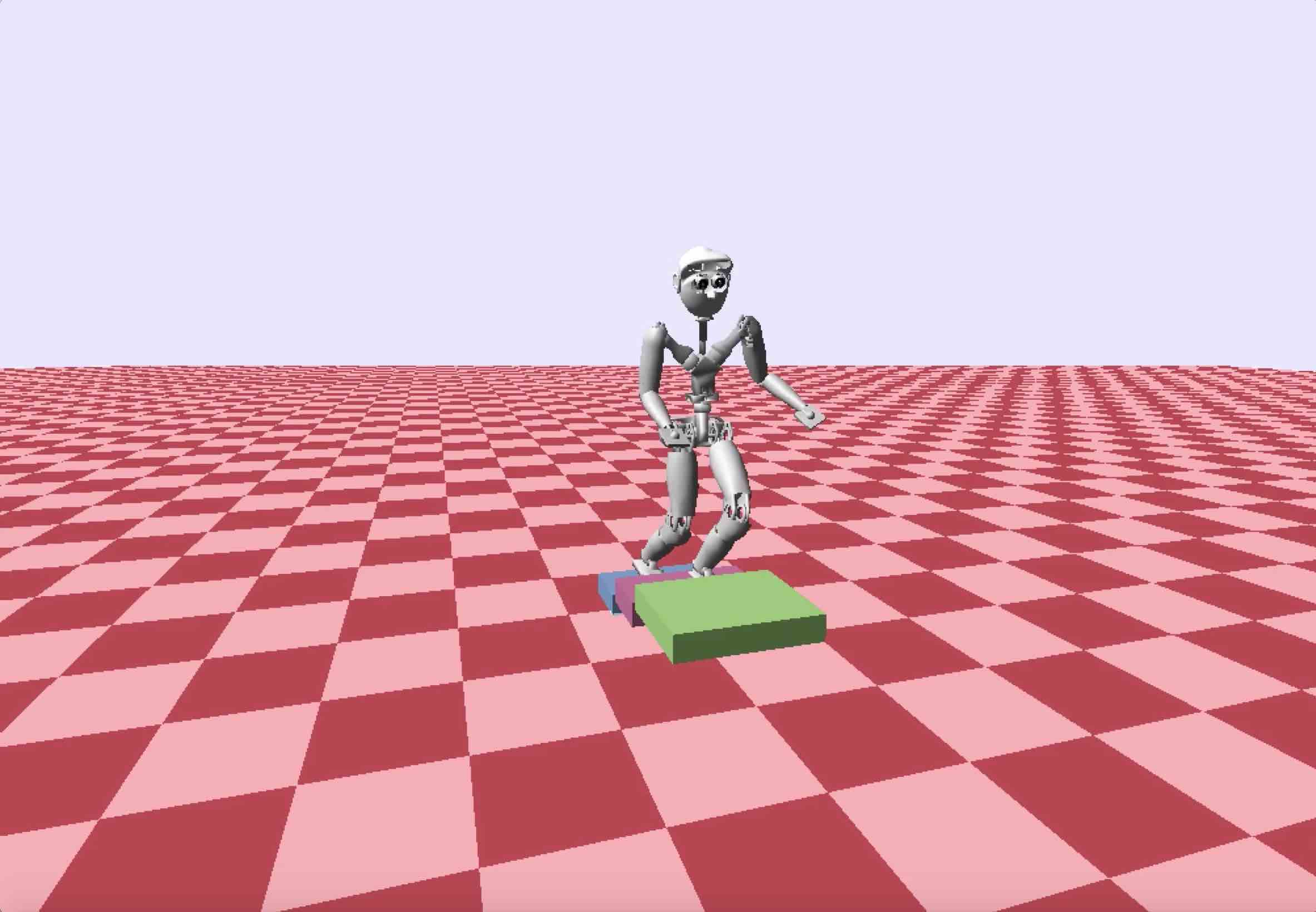}
		\caption[]{\footnotesize Walking stairs}
		\label{fig:motion03}
	\end{subfigure}
	\begin{subfigure}[b]{0.24\linewidth}   
		\centering
		\includegraphics[width=1.00\linewidth, trim={10cm 2.0cm 10cm 3.0cm}, clip]{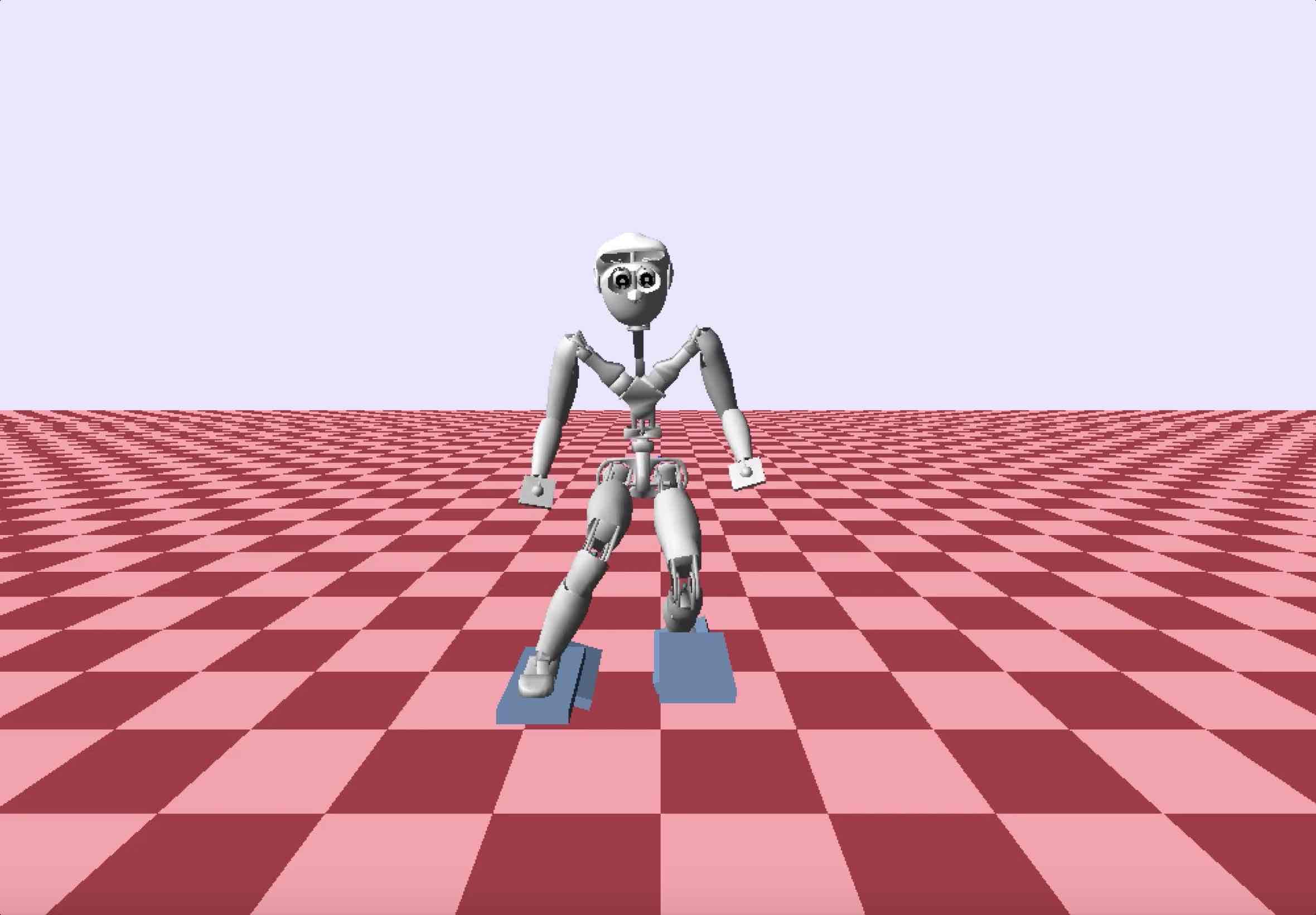}
		\caption[]{\footnotesize Up stairs}
		\label{fig:motion04}
	\end{subfigure}
	\begin{subfigure}[b]{0.24\linewidth}   
		\centering
		\includegraphics[width=1.00\linewidth, trim={10cm 2.0cm 10cm 3.0cm}, clip]{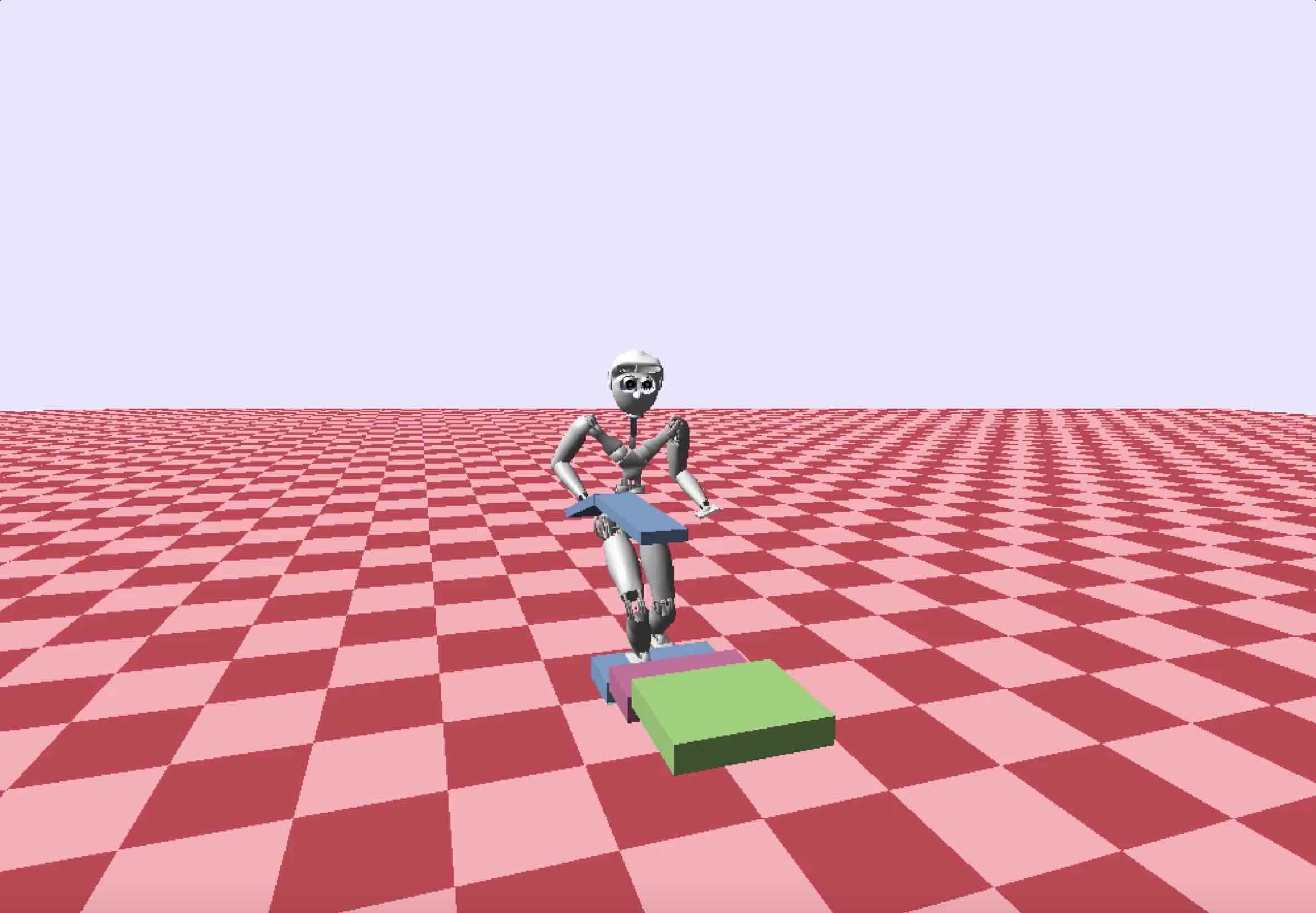}
		\caption[]{\footnotesize Using hands}
		\label{fig:motion05}
	\end{subfigure}
	\begin{subfigure}[b]{0.24\linewidth}   
		\centering
		\includegraphics[width=1.00\linewidth, trim={10cm 1.5cm 10cm 5.5cm}, clip]{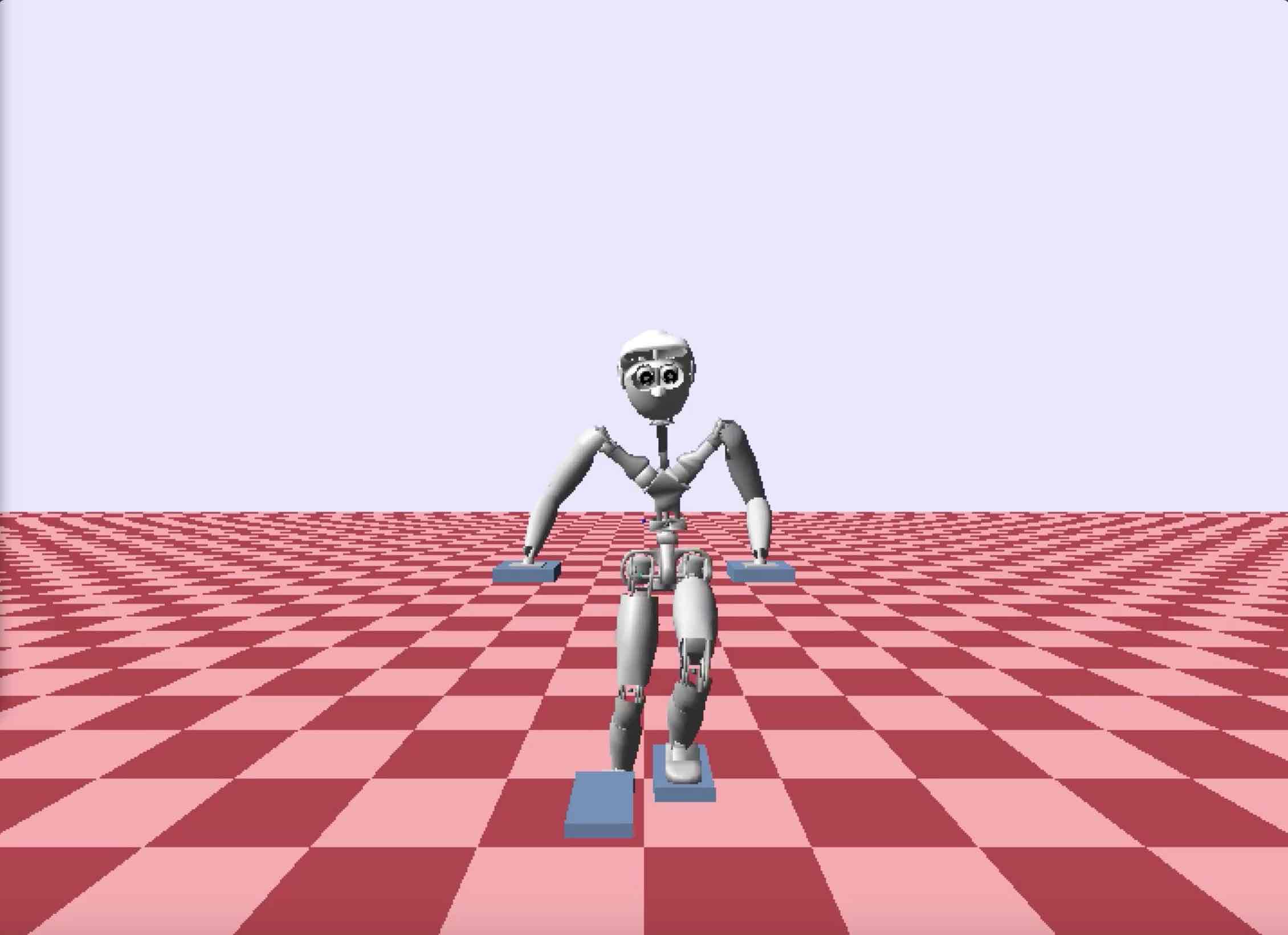}
		\caption[]{\footnotesize Up with hands}
		\label{fig:motion06}
	\end{subfigure}
	\begin{subfigure}[b]{0.24\linewidth}   
		\centering
		\includegraphics[width=1.00\linewidth, trim={10cm 2.0cm 10cm 3.0cm}, clip]{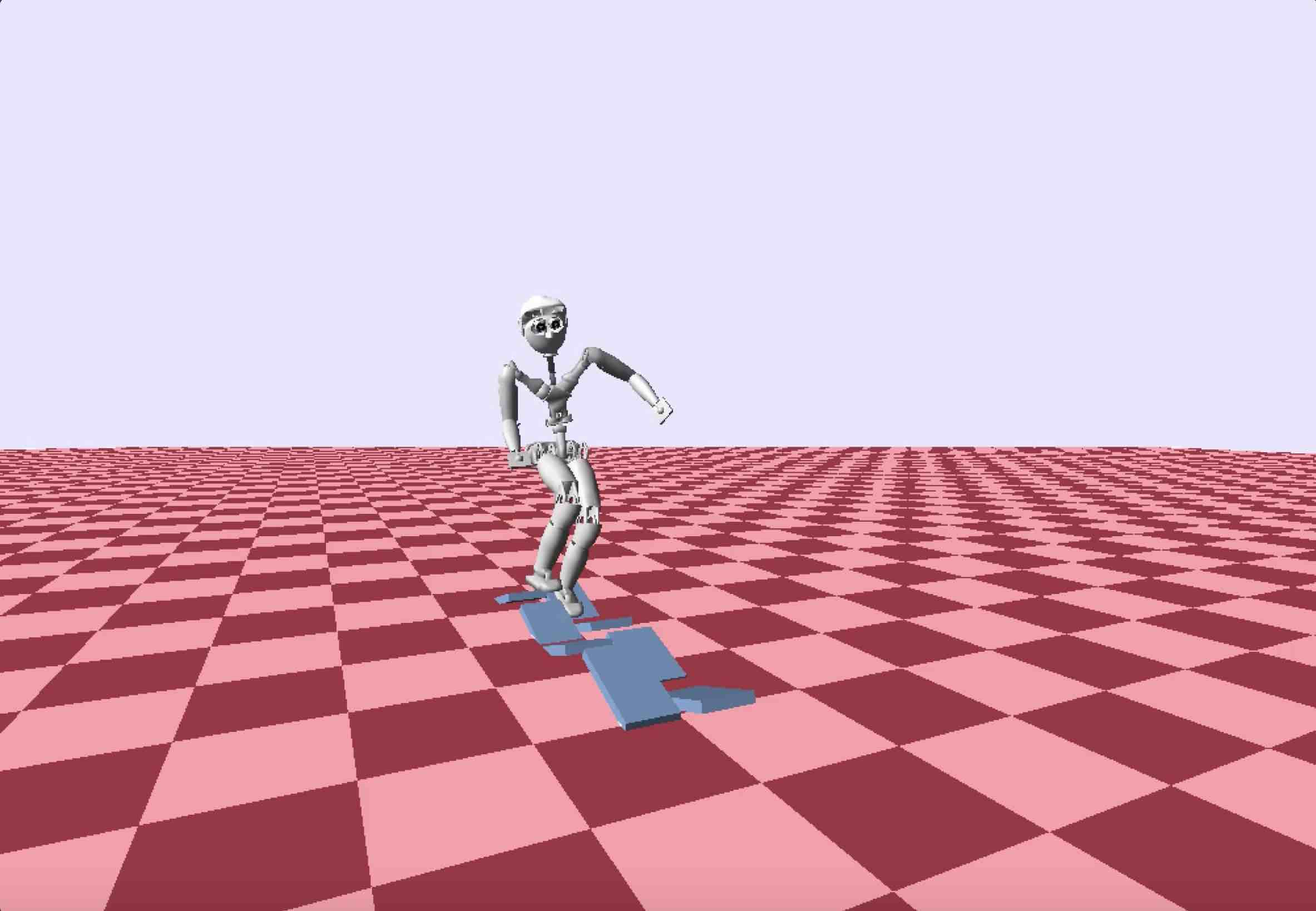}
		\caption[]{\footnotesize Tilted terrain}
		\label{fig:motion07}
	\end{subfigure}
	\begin{subfigure}[b]{0.24\linewidth}   
		\centering
		\includegraphics[width=1.00\linewidth, trim={10cm 2.0cm 10cm 3.0cm}, clip]{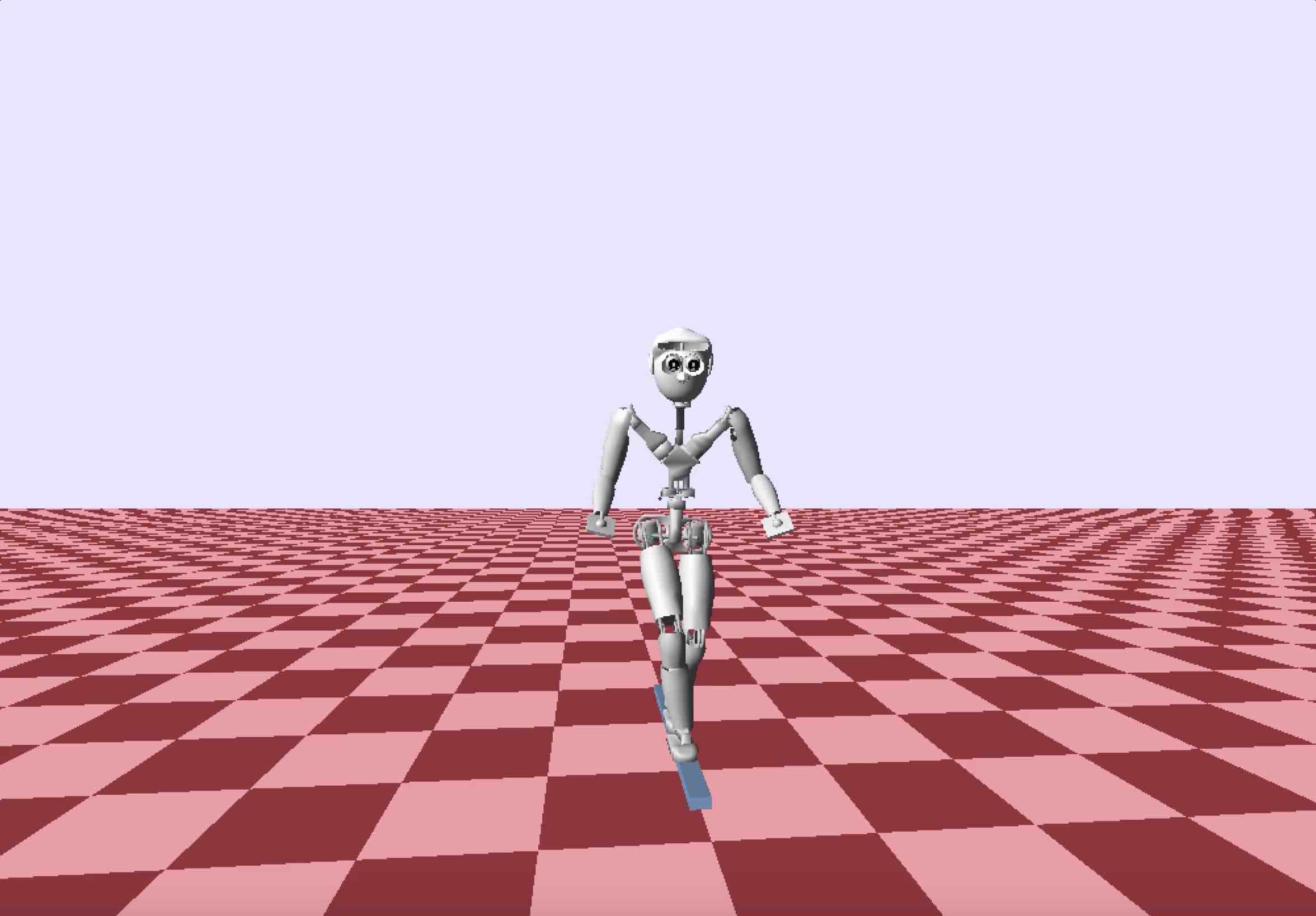}
		\caption[]{\footnotesize Narrow path}
		\label{fig:motion08}
	\end{subfigure}
	\caption[]{\small Examples of time-optimized dynamic movement plans.} 
	\label{fig5:movement_planning}
\end{figure}

\begin{figure}
	\centering
	\includegraphics[width=1.00\linewidth, trim={0.0cm 0.0cm 0.0cm 0.0cm}, clip]{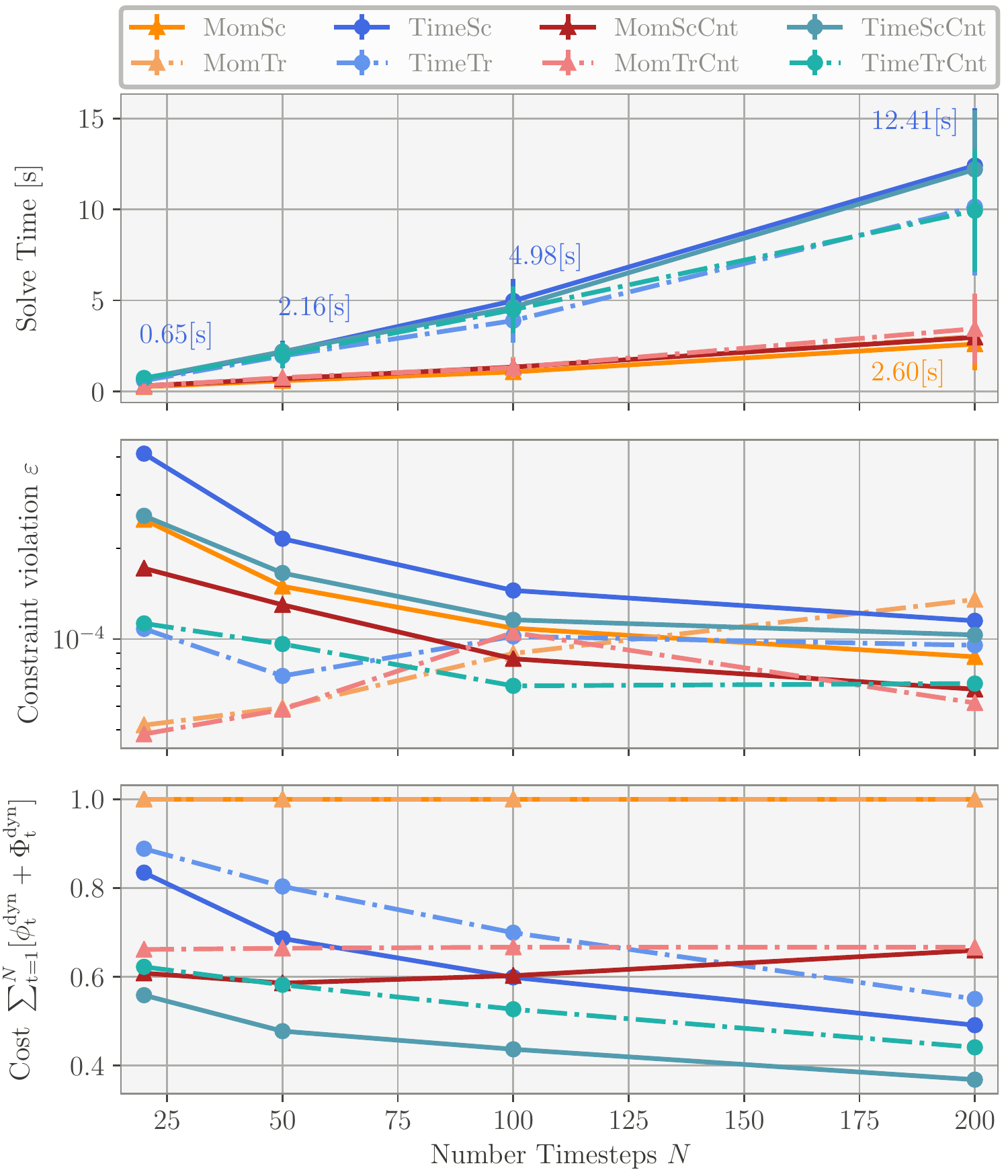}
	\caption[]{\small \emph{Top:} \ctxt{Roughly linear-time} complexity of movement plans \ctxt{within the shown range of timesteps {$\dishorizon$}}: with or without time optimization $Time-Mom$, using soft-constraint or trust-region heuristics $Sc-Tr$, and with or without optimization of contact locations $Cnt$. \emph{Center:} Corresponding normalized convergence errors or amount of constraint violation ${\converr}$ as given by \eqref{eq_error} and \emph{Bottom:} numerical relative cost reduction of motions optimized including time and/or contact locations with respect to motions using fixed contacts and timings. Each datapoint averages information from 8 experiments (shown in Fig. \ref{fig5:movement_planning}) optimized using the same objective function but different number of timesteps {$\dishorizon$} and heuristics.} 
	\label{fig:conv_per_num_timesteps}
\end{figure}

To analyze convergence properties and computational complexity of the algorithm, we use a set of 8 optimized motions (shown in Fig. \ref{fig5:movement_planning}) to gather statistics about its performance. Table \ref{tab:costfunction} shows a typical cost function and the relative importance of the weighted costs used to optimize a motion.
%
%================ Eight Scenarios ================%
%=================================================%
%
\begin{table}[]
	\begin{tabular}{|@{ }L{3.6cm}@{ } | @{ }L{2.5cm}@{ } | @{ }C{2.0cm}@{ }|}
		\hline
		\centering\textbf{{Cost}} & $\;\;$\textbf{{Functional Form}} & \textbf{{Scaling Order}} \\
		\hline
		$\;\quad$ {CoM terminal cost}		& ${\quad\fquad(\coms_{\dishorizon} - (\coms_{0}+\Delta\coms))}$ 										& ${1e+4}$	\\
		$\;\quad$ {Time regularization}	& ${\quad\sum_{\tid}\fquad(\timeopt_{\tid}-\indexed[0]{\timeopt}[][\tid])}$	& ${1e+3}$	\\
		$\;\quad$ {Momenta terminal cost} 	& ${\quad\fquad(\moms_{\dishorizon})}$										& ${1e+2}$	\\
		$\;\quad$ {Endeffector consensus cost} & ${\quad\sum_{\tid}\fquad(\indexed{\effpos}[][\effid,\tid] - \indexed[kin]{\effpos}[][\effid,\tid])}$	& ${1e+0}$ \\
		$\;\quad$ {Momenta consensus cost} & ${\quad\sum_{\tid}\fquad(\indexed{\moms}[][\tid] - \indexed[kin]{\moms}[][\tid])}$	& ${1e+0}$ \\
		$\;\quad$ {Momenta rate  cost}		& ${\quad\sum_{\tid}\fquad(\dot{\moms}_{\tid})}$								& ${1e-1}$	\\
		$\;\quad$ {Momenta running cost}  	& ${\quad\sum_{\tid}\fquad(\moms_{\tid})}$									& ${1e-2}$ 	\\
		$\;\quad$ {Force running cost}		& ${\quad\sum_{\tid}\fquad(\forcevars_{\effid,\tid})}$						& ${1e-3}$	\\
		$\;\quad$ {Torque running cost}	& ${\quad\sum_{\tid}\fquad(\coptrq_{\effid,\tid})}$						& ${1e-3}$	\\
		\hline
	\end{tabular}
	\caption[]{\small {Example composition of the main components of the cost $\sum_{\tid}(\indexed{\fcost}[dyn][\tid] + \indexed{\fconsensus}[dyn][\tid])$ used to synthesize the walking motion of Figure \ref{fig:motion04}}.}
	\label{tab:costfunction}
\end{table}
In Fig. \ref{fig:conv_per_num_timesteps}, we present statistics about time complexity, convergence to feasibility and relative cost reduction when using the same objective function but different number of discretization timesteps and algorithmic settings. In particular, we look at what happens when the optimization includes or not time as an optimization variable $Time \textrm{ vs. } Mom$, includes or not optimization of contact locations $Cnt$, uses the soft-constraint or trust-region relaxation approaches $Sc \textrm{ vs. } Tr$. As an example, $MomSc$ refers to a motion optimized with fixed timings, fixed contacts and using the soft-constraint heuristic, while $TimeTrCnt$ refers to a motion optimized including timings, contact locations and using the trust-region heuristic.

%============= Momentum optimization =============%
%=================================================%

First of all, in the center plot we show the amount of constraint violation ${\converr}$ of the optimized solutions, as measured by \eqref{eq_error}. We note that the algorithm converges when ${\converr}$ or its reduction from one iteration to another ${\indexed{\converr}[][\numiters]-\indexed{\converr}[][\numiters-1]}$ fall below a desired threshold (typically in the order of $1e-4$) and, as visible on the plot, our method converges in all experiments to the desired feasibility thresholds in all settings. 

On the top, we show statistics about the time-complexity of the algorithm for convergence to the desired feasibility thresholds; in particular, this shows evidence of linear complexity in the number of timesteps for momentum and time optimization problems. We notice that for fixed-time optimization problems, neither heuristics nor the optimization of contact locations affect the solving time performance. A similar behavior can be seen for time optimization problems with the difference that the trust-regions are slightly faster than the soft-constraints. 

Finally, on the bottom plot we show numerically the relative reduction of the cost when optimizing time and contact locations. In orange tones, we see the reference normalized costs of momentum optimization problems using fixed contact locations and timings for trust-region and soft-constraint heuristics, namely $MomSc$ and $TimeSc$. As expected both achieve a similar minimum and have thus the same normalized cost of one. As shown above, considering contact locations as optimization variables (in the form of linear constraints and over a given terrain surface) has minimum impact on solving time performance, yet it significantly reduces the objective value (between 35 and 40 percent) because this degree of freedom allows the optimizer to select motions with lower momentum values, e.g. motions with less lateral sway of the CoM (see $MomScCnt$ and $MomTrCnt$ in red tones).

%=============== Average iterations ==============%
\begin{figure}
	\centering
	\includegraphics[width=0.48\textwidth]{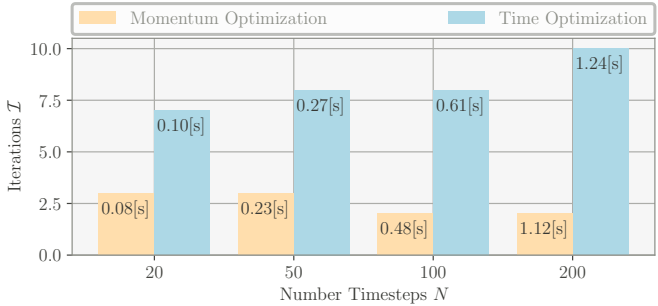}
	\caption[]{\small Average number of iterations required to solve an optimization problem with or without time optimization for different number of discretization timesteps and, average time to solve each iteration. Each datapoint is based on 32 experiments, with different heuristic $Sc-Tr$ and with our without optimization of contacts $Cnt$.}
	\label{fig:average_timings}
\end{figure}
%=================================================%
%
The effect of time optimization on the objective value is dependent on the problem time horizon (or number of timesteps $\dishorizon$). For simplicity, we can assume that the value of one timestep is 0.1 seconds (which is the discretization time we use for fixed time optimization) and thus the horizontal axis spans between 2 and 20 seconds. For instance, in problems with short-time horizons such as those at the leftmost side, the cost difference between motions that consider or not time as an optimization variable is modest, but as the look-ahead horizon increases (right side) time optimization becomes a powerful way of shaping the motion to achieve lower costs. We notice that in this case the soft-constraint heuristic ($timeSc$ and $timeScCnt$) finds in average slightly lower local minima than the trust-region heuristic ($timeTr$ and $timeTrCnt$).

In Fig. \ref{fig:average_timings}, we show the average number of iterations required to solve a momentum or time optimization problem for varying number of timesteps, as well as the average time required to solve each of these iterations. For instance, momentum optimization problems require 2-3 iterations, while time optimization problems 7-10. However, the difference in solving times of one iteration is small, e.g. for a time horizon of 2 seconds ($\dishorizon = 20$) the solving times are 80 and 100 [ms] for momentum and time optimization problems respectively. This suggests that the approach could be used in a receding horizon setting. In such setting, the optimizer could be warm-started from the previous solution to significantly increase resolution time (typically one would only need to solve one iteration of the problem for a short look-ahead horizon).
\subsubsection{Qualitative improvement of solutions} \label{exp:qualitative_results}
\begin{figure}
	\centering
	\hspace{\stretch{1}}
	\drawvideo{5}{80}{%
		\includegraphics[width=0.20\linewidth, trim={18cm 4cm 20cm 0cm}, clip]{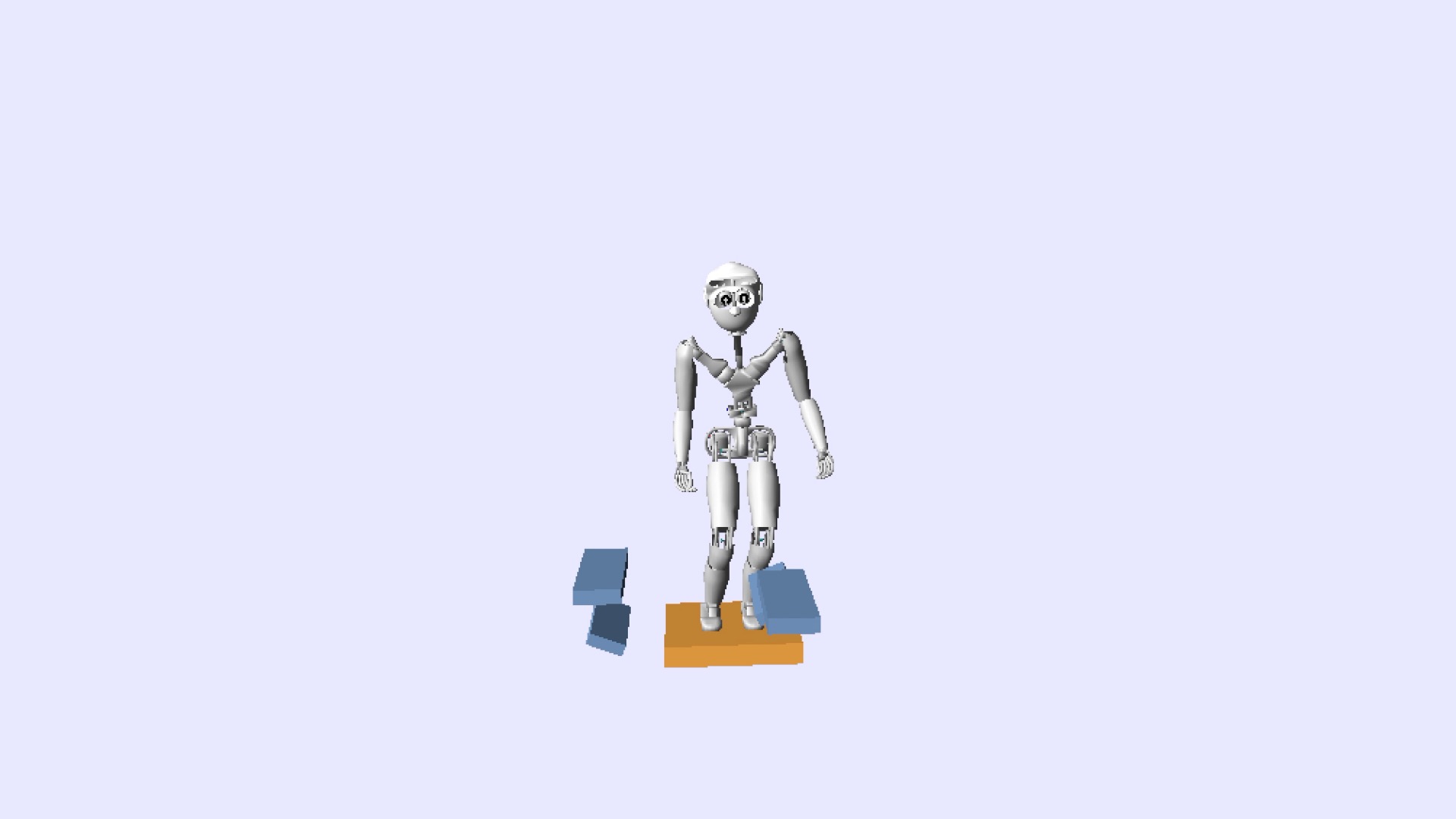}%
		\includegraphics[width=0.20\linewidth, trim={18cm 4cm 20cm 0cm}, clip]{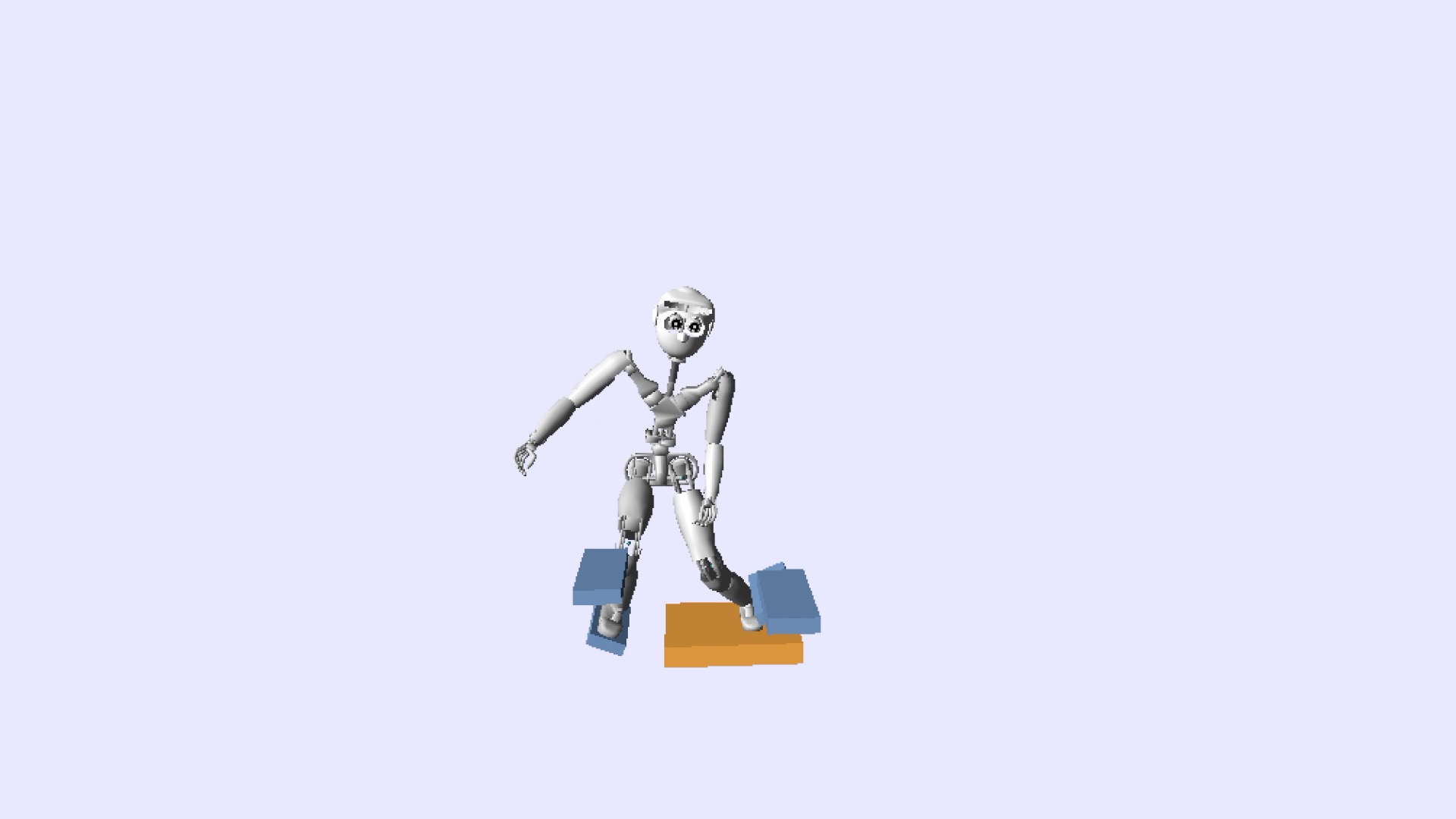}%
		\includegraphics[width=0.20\linewidth, trim={18cm 4cm 20cm 0cm}, clip]{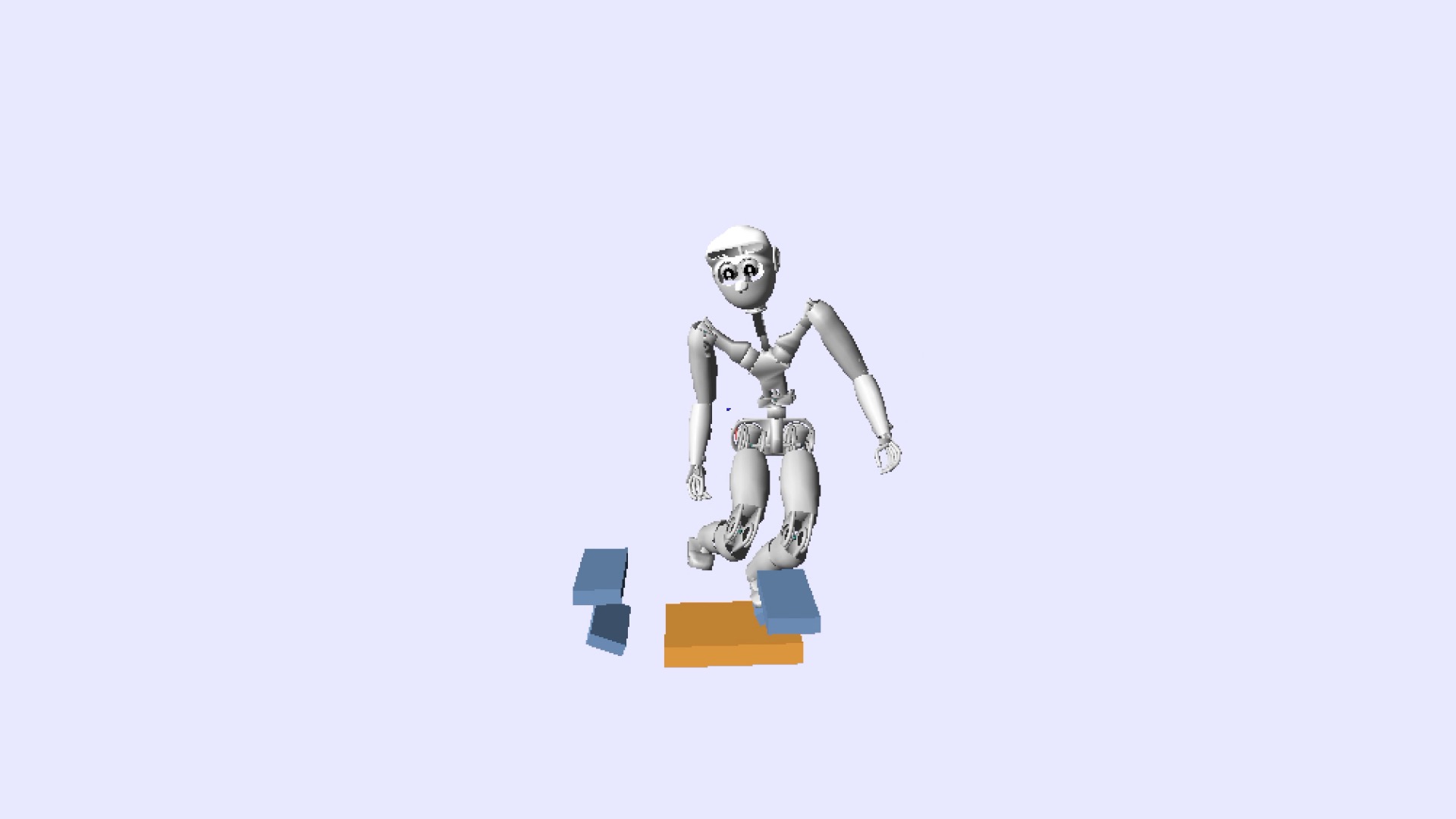}%
		\includegraphics[width=0.20\linewidth, trim={18cm 4cm 20cm 0cm}, clip]{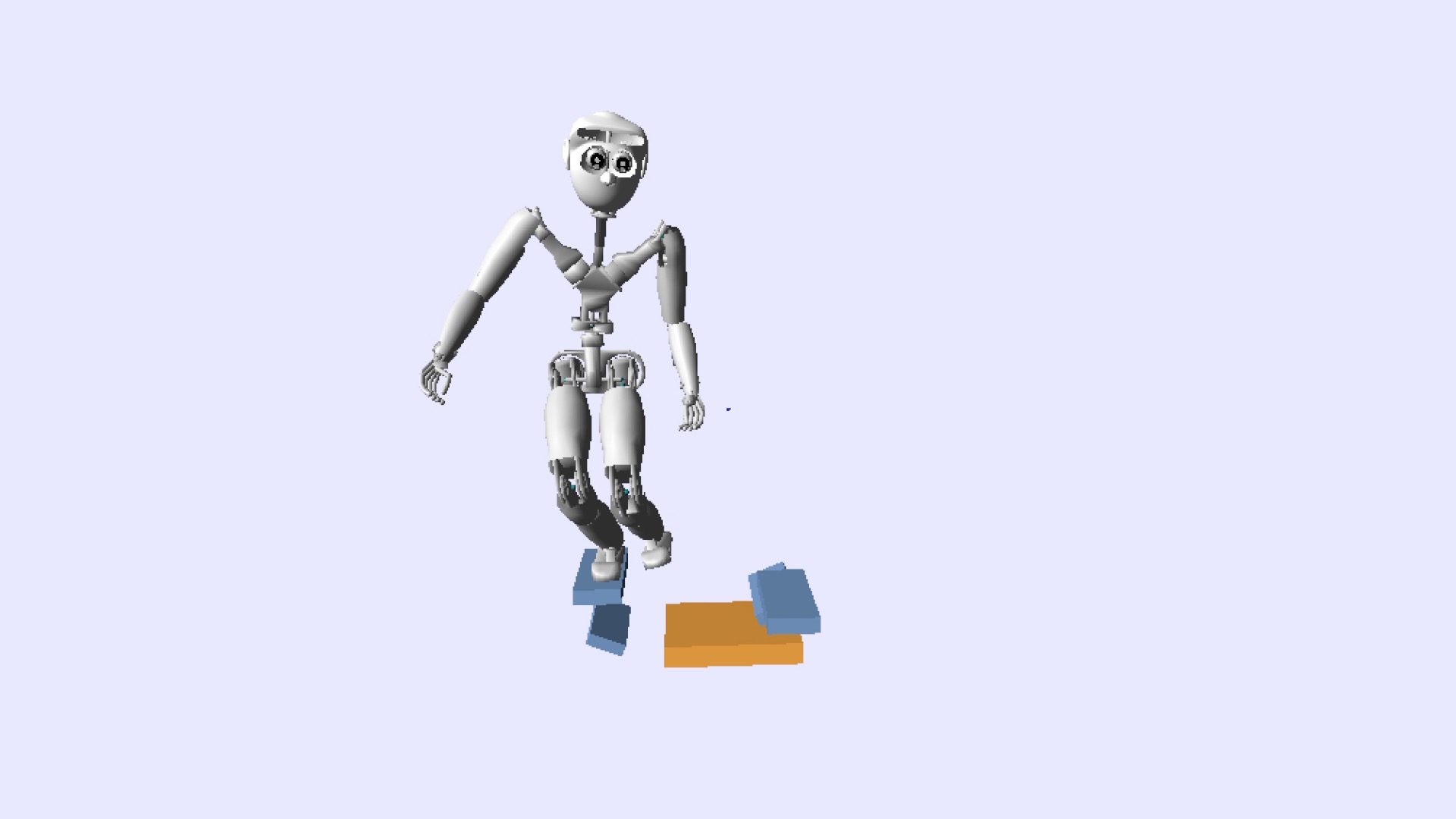}%
		\includegraphics[width=0.20\linewidth, trim={18cm 4cm 20cm 0cm}, clip]{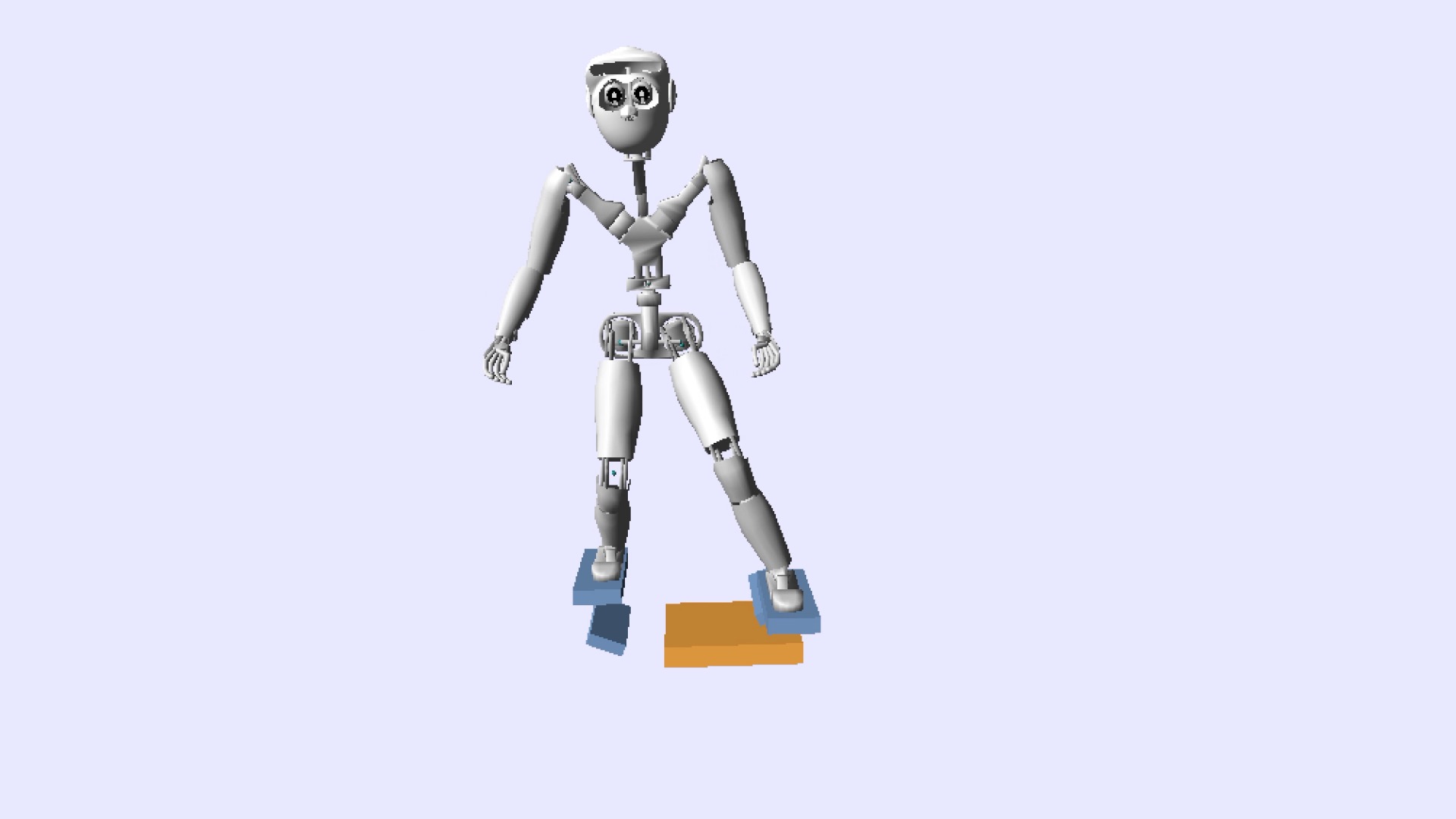}%
	}\hspace{\stretch{1}}\\
	\vspace{0.2cm}
	\includegraphics[width=0.49\textwidth]{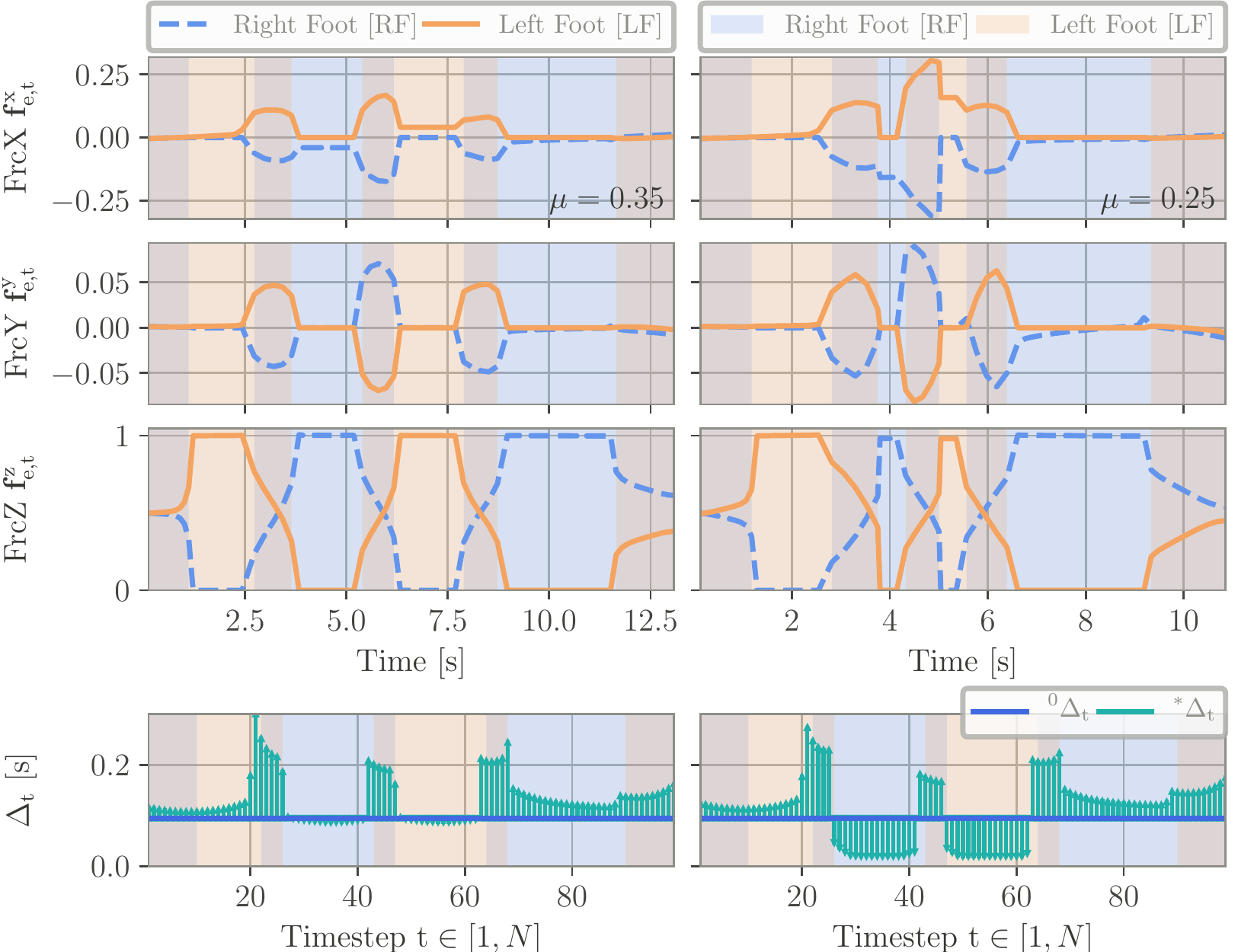}
	\caption[]{\small Comparison of optimal normalized endeffector forces and timing results for two different values of friction coefficient $\friccoeff$. Timings ${\indexed[0]{\timeopt}[][\tid]}$ are the initial ones and ${\indexed[*]{\timeopt}[][\tid]}$ the final optimized ones.}
	\label{fig:lowfriction_comparison}
\end{figure}

Here, we discuss qualitative results that cannot be described from the statistical analysis above. We therefore restrict our analysis to specific instances of the problem. In Fig. \ref{fig:lowfriction_comparison} we show time optimal results for a walking up tilted stairs motion traversed with two different values of the friction coefficient $\friccoeff$. In the first case ($\friccoeff = 0.35$), the tendency is to increase the value of timestep variables $\timeopt_{\tid}$ during double supports to have enough time to slowly accelerate the CoM while respecting physical constraints, resembling statically stable motions. In an environment with flat surfaces, the same approach would be valid even if the friction coefficient is further reduced (e.g. $\friccoeff = 0.25$). However, in a terrain with tilted surfaces such a strategy is not viable. In such a setting, even the fixed-time version of our algorithm cannot find a dynamically feasible solution. Yet, our time optimization approach is able to find a solution, whose main strategy is to quickly traverse the tilted surfaces to get to the uppermost flat contact surfaces. During this phase, lateral contact forces are exploited to the limit, and then a similar strategy to the previous case is found.

%============== Low friction motion ==============%
%=================================================%

\begin{figure}
	\centering
	\hspace{\stretch{1}}
	\drawvideo{5}{80}{%
		\includegraphics[width=0.20\linewidth, trim={18cm 0cm 15cm 0cm}, clip]{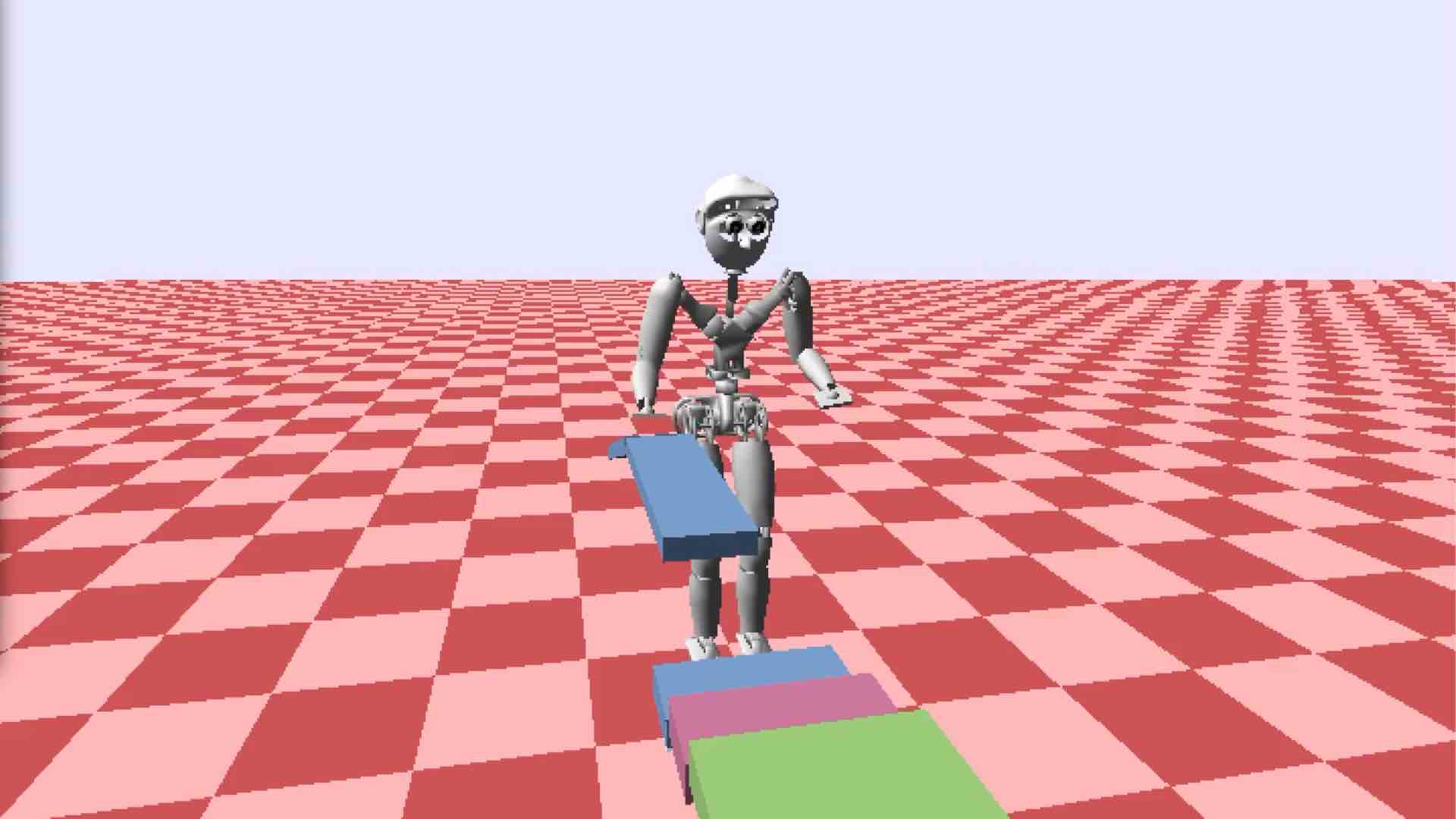}%
		\includegraphics[width=0.20\linewidth, trim={18cm 0cm 15cm 0cm}, clip]{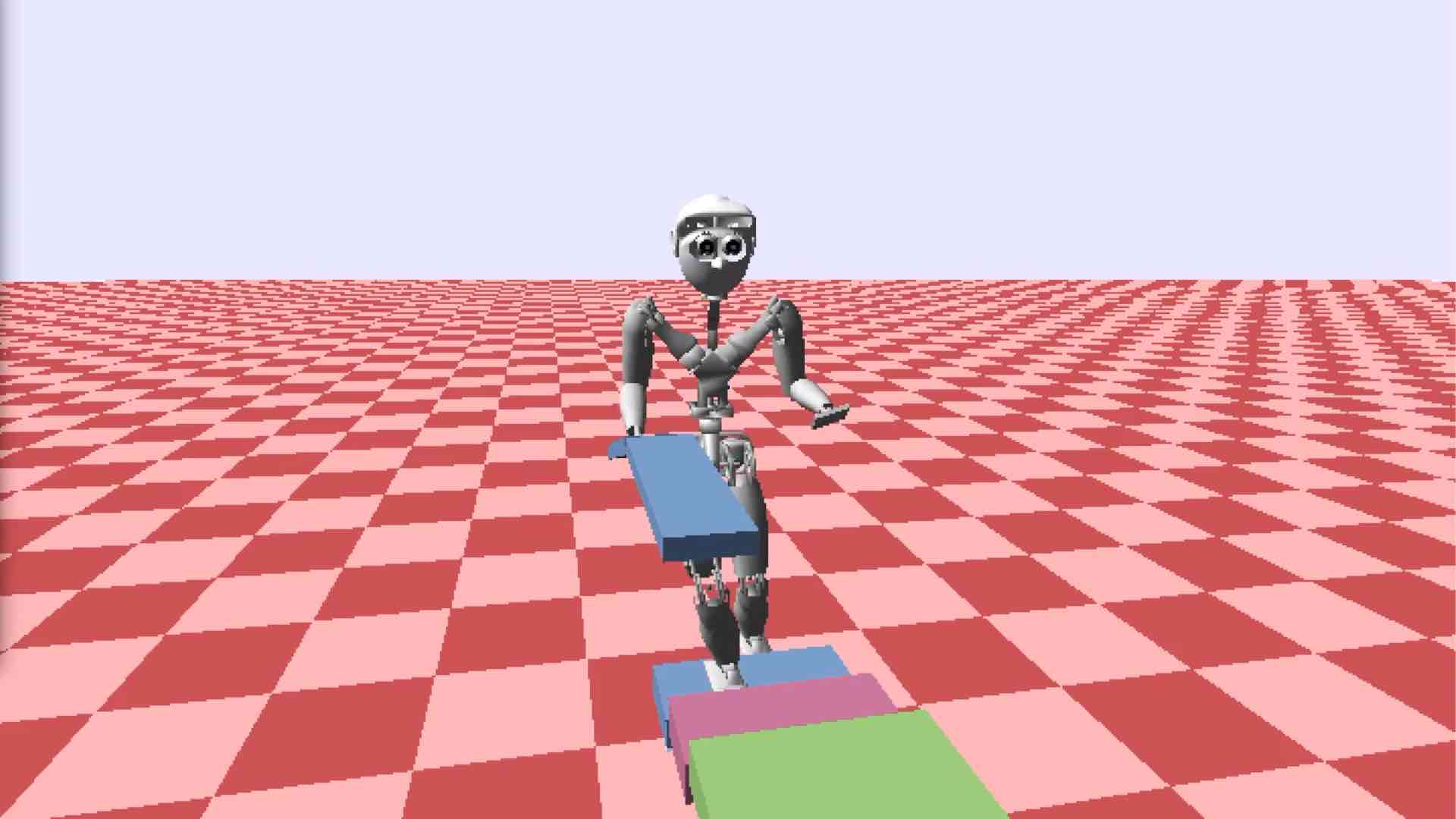}%
		\includegraphics[width=0.20\linewidth, trim={18cm 0cm 15cm 0cm}, clip]{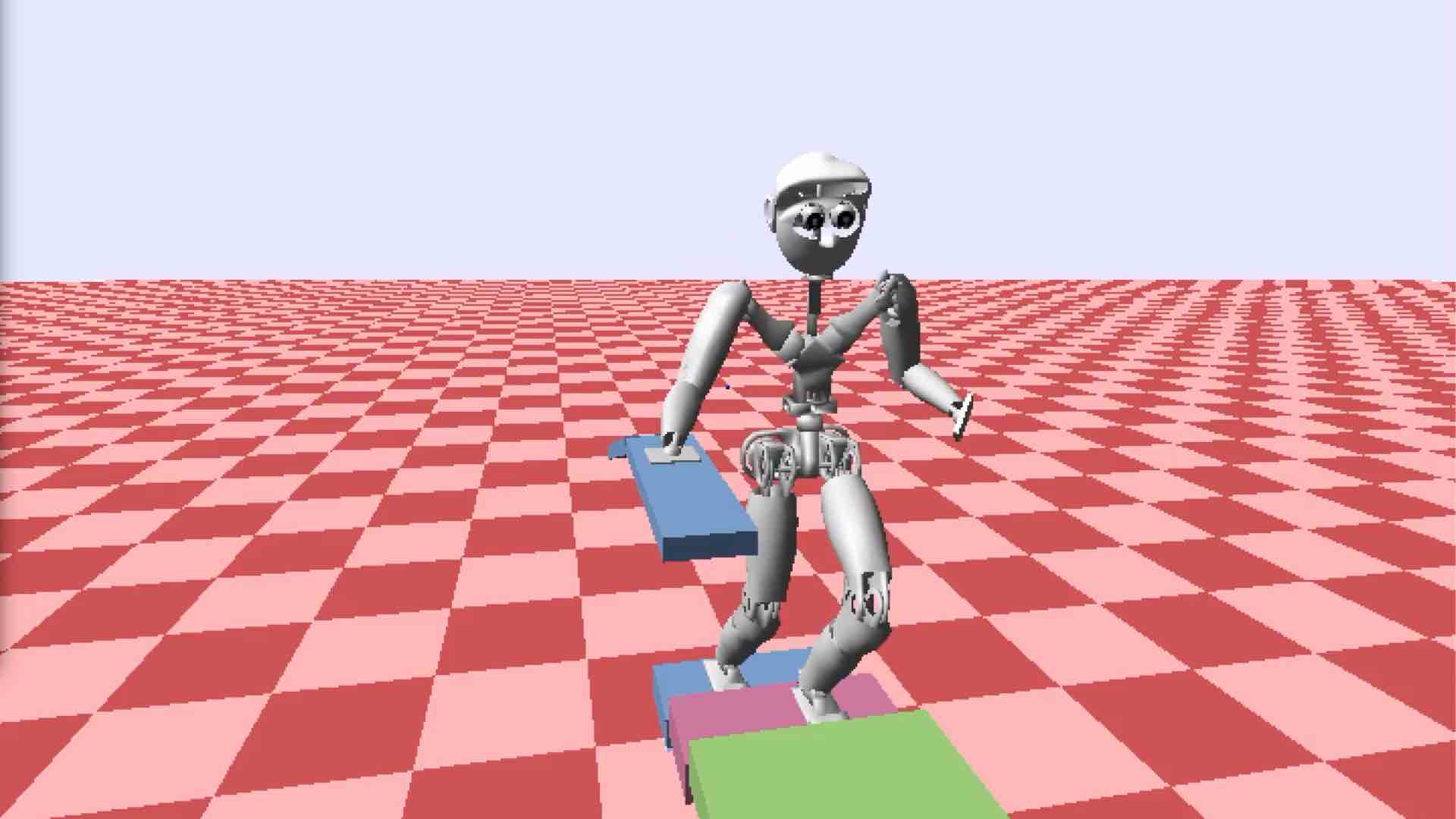}%
		\includegraphics[width=0.20\linewidth, trim={18cm 0cm 15cm 0cm}, clip]{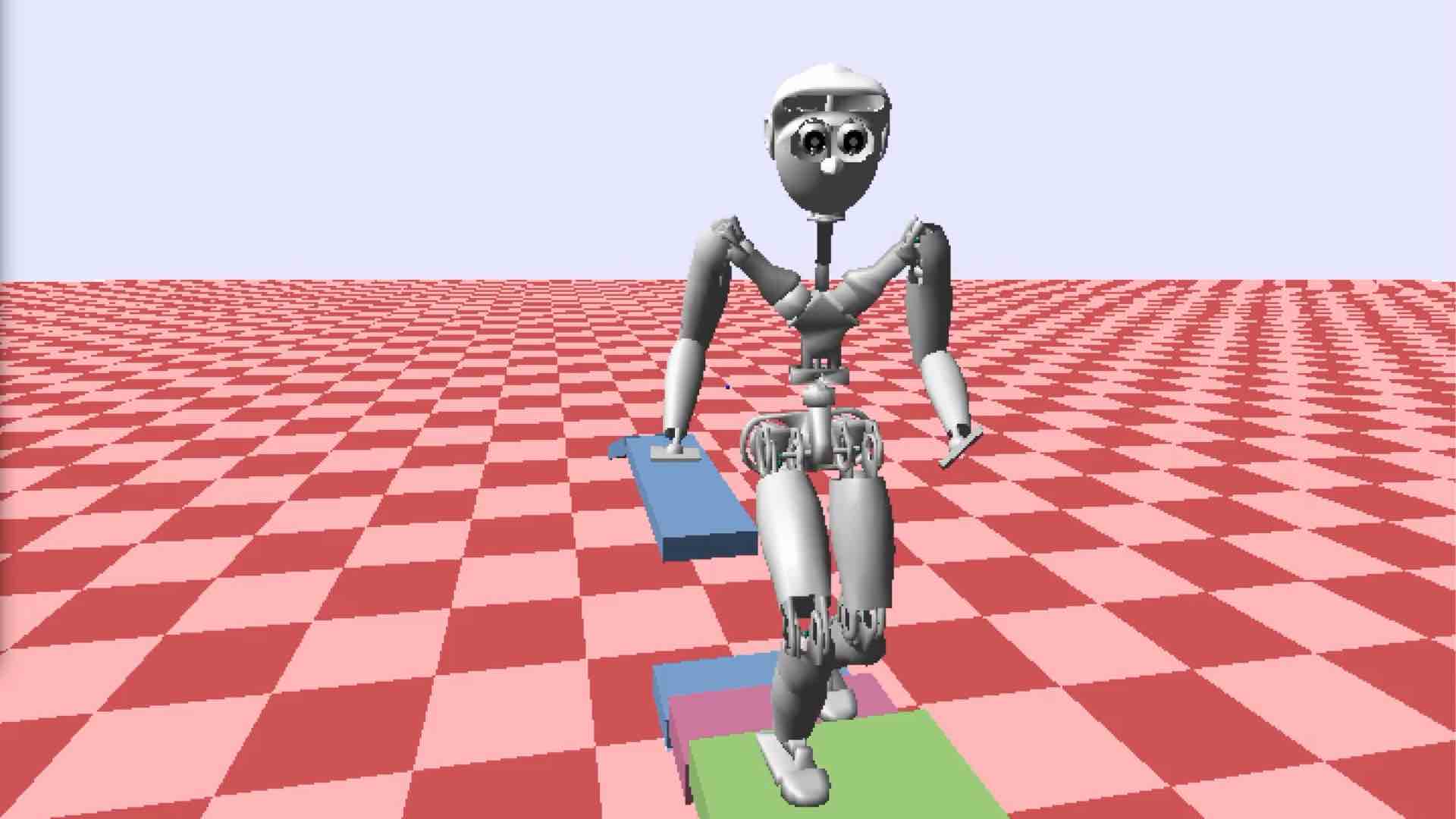}%
		\includegraphics[width=0.20\linewidth, trim={18cm 0cm 15cm 0cm}, clip]{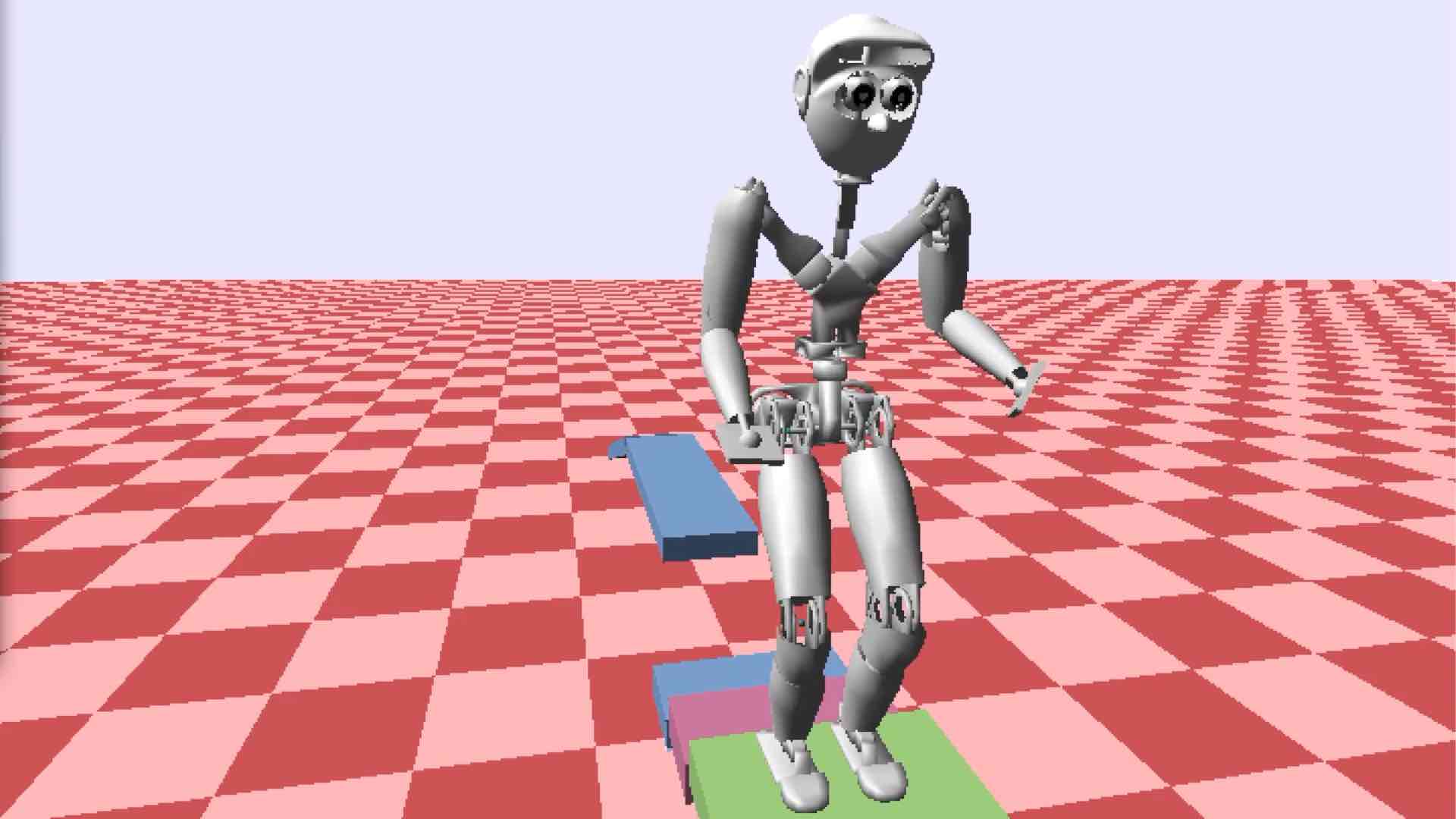}%
	}\hspace{\stretch{1}}\\
	\vspace{0.2cm}
	\includegraphics[width=0.49\textwidth]{\filename{figures/handstairs/HandingMotion}}
	\caption[]{\small Comparison between CoM and normalized linear momentum in the lateral direction for a walking up stairs motion using hands. {The squares, circles, diamonds and stars show the endeffector locations $\indexed{\effpos}[][\effid,\tid]$ optimized under different settings, as shown in the legend.} Bottom plot shows the contact activation of endeffectors over the time horizon for momentum (Mom) and time (Time) optimization problems (low value is {inactive} and high value is active).}
	\label{fig:walking_up_with_hands}
\end{figure}

In Fig. \ref{fig:walking_up_with_hands}, we show a walking up stairs motion using hand contacts. In our experiments, in such multi-contact scenarios time optimization does not significantly change motion timings, as can be seen in the bottom plot (that graphically illustrates endeffector activations $\setacteff$) by comparing the timings of a fixed-time optimization problem (Mom) and those of a time optimization problem (Time). However, optimizing contact locations allows us to find motions with less CoM sway. This is visible, for example, in the CoM trajectories for a momentum optimization without optimization of contact locations $MomSc$ or even a time optimization without optimization of contacts $TimeSc$ and a momentum or time optimization that includes optimization of contact locations such as $MomScCnt$ and $TimeScCnt$ respectively. These motions are more energetically efficient and arguably easier to control with only a small additional computational cost in the optimization. Note that the top plot in Fig. \ref{fig:walking_up_with_hands} is a top view of the walking up stairs movement using hands, not to be confused with a planar motion.

%============== Motion using hands ===============%
%=================================================%
%
\subsubsection{Kino-dynamic full-body optimization}
In this section, we show how our algorithm can be used in the kino-dynamic approach described in Section \ref{sec:problem_formulation}, and illustrated in Figure \ref{fig:KinDynStructure}, to generate whole-body time-optimal motions.

\begin{figure}
	\centering
	\includegraphics[width=0.49\textwidth]{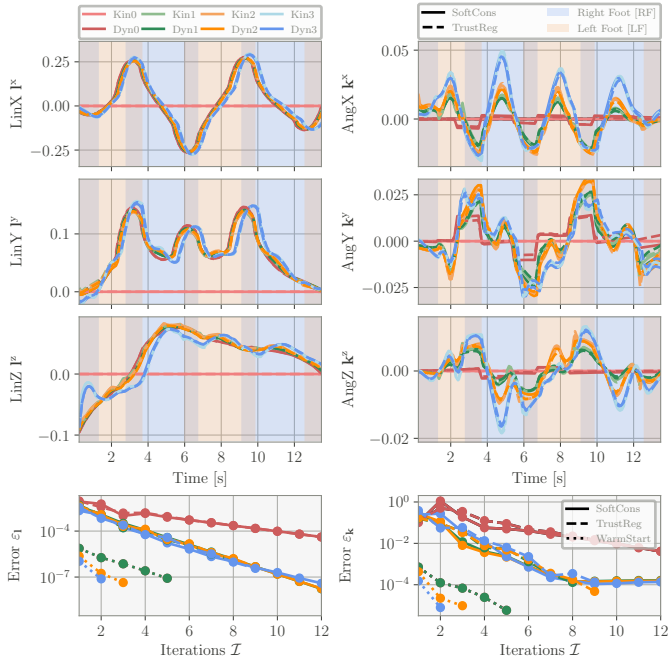}
	\caption[]{\small This figure shows convergence to feasibility of each dynamics optimization along three kino-dynamic iterations. We compare desired kinematic momentum trajectories $Kin$ and dynamic momentum trajectories $Dyn$ (computed out of optimal controls) at the end of each dynamic optimization. Bottom plots (left for linear momentum and right for angular momentum) show how errors {$\indexed{\converr}[][\lmoms]$ and $\indexed{\converr}[][\amoms]$} decrease until convergence along each kino-dynamic iteration. Momentum values are normalized by robot mass. Vertical colored bars show the activation of each endeffector over time.} 
	\label{fig:kinodynamic_approach}
\end{figure}

First, we use the climbing uneven stairs motion depicted in Fig. \ref{fig:motion04} to illustrate algorithmic convergence of our method to kino-dynamic consistency. In Fig. \ref{fig:kinodynamic_approach}, we graphically compare (on the top 3 plots) kinematic $\indexed[kin]{\moms}$ and dynamic momentum trajectories $\indexed[dyn]{\moms}$ at the end of each dynamics optimization. We use dark colors to show dynamic trajectories $\indexed[dyn]{\moms}$, and the same, but light color, for kinematic ones $\indexed[kin]{\moms}$. Solid lines correspond to motions optimized using soft-constraints, and dashed lines to motions optimized using trust-regions. {It can be seen from the plots that they} qualitatively converge to similar solutions, as it is difficult to distinguish them from each other.

%============ Feasibility convergence ============%
%=================================================%

On the bottom plot, we show how quantitatively the norms $\indexed{\converr}[][\lmoms]$ and $\indexed{\converr}[][\amoms]$, that compare momentum trajectories obtained from optimal controls and the momentum trajectory variables that track desired kinematic momentum trajectories, decrease until convergence at each kino-dynamic iteration. Note that the first dynamics optimization (shown in red) takes the longest to converge and that trajectories optimized in subsequent iterations without using any information from previous ones converge faster (see e.g. how solid and dashed lines from the first iteration compare to those at subsequent iterations). In practice however, by warm-starting the heuristics of dynamic optimizations with the results and information of previous iterations, the optimization problems can be solved much faster and with fewer iterations, as shown in dotted trajectories. Despite that at each iteration kinematic and dynamic momentum trajectories match, in practice we use at least two iterations to converge to a motion easily executable on a physical simulator.

Note as well how linear momentum converges fast and to high levels of precision, while angular momentum does it only to modest levels. See for example, how solid and dashed lines achieve in 4 iterations the required precision for linear momentum errors $\indexed{\converr}[][\lmoms]$, while it takes around 8 for angular momentum errors $\indexed{\converr}[][\amoms]$. This is due to the fact that on the one hand angular momentum depends on the CoM and can only achieve a higher precision once this variable has converged, and on the other hand due to the fact that given a CoM trajectory, angular momentum can be further optimized along it by exploiting the control degrees of freedom left.

\begin{figure}
	\centering
	\drawvideo{5}{80}{%
		\includegraphics[width=0.20\linewidth, trim={5cm 3cm 12cm 5cm}, clip]{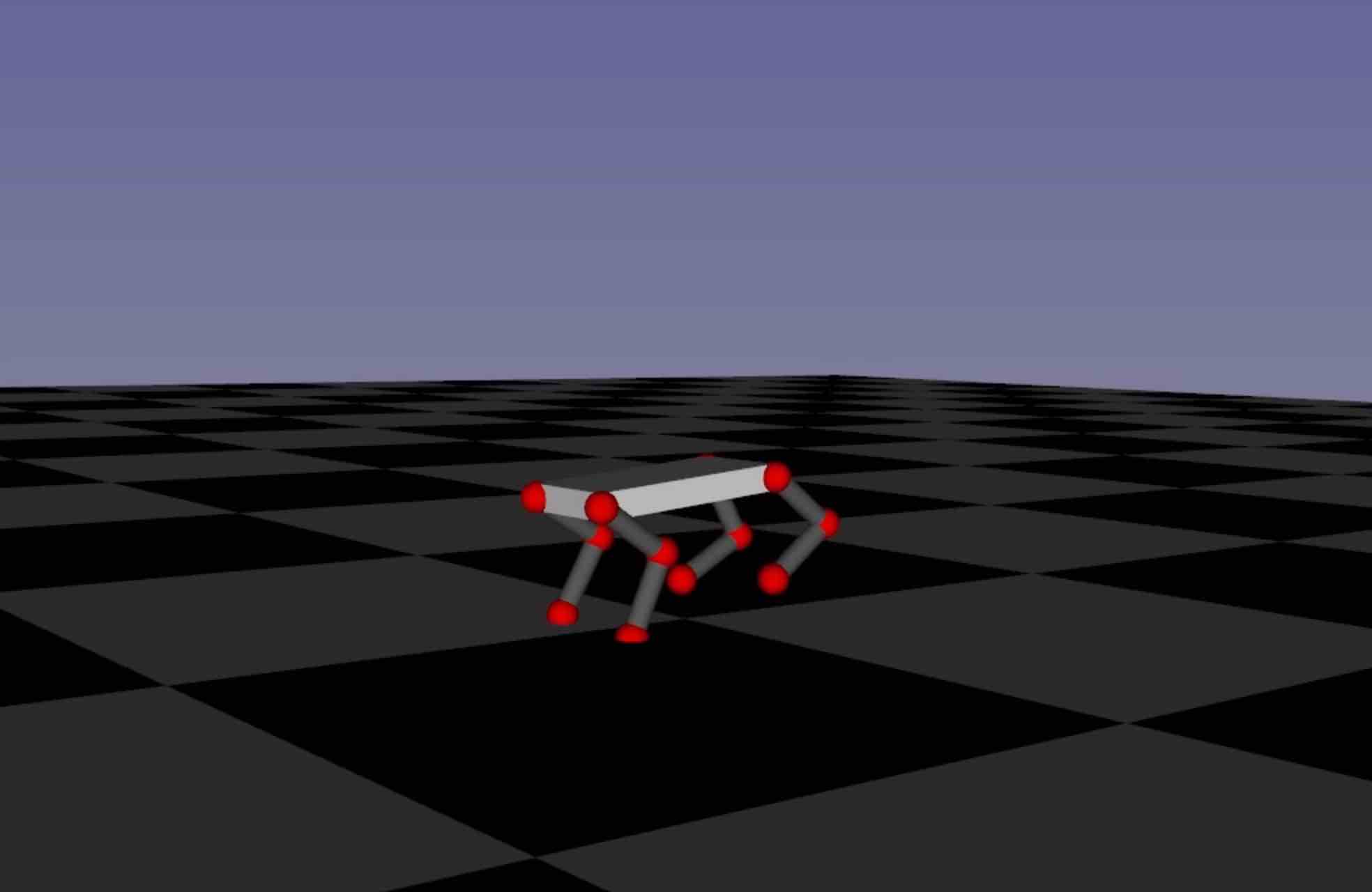}%
		\includegraphics[width=0.20\linewidth, trim={5cm 3cm 12cm 5cm}, clip]{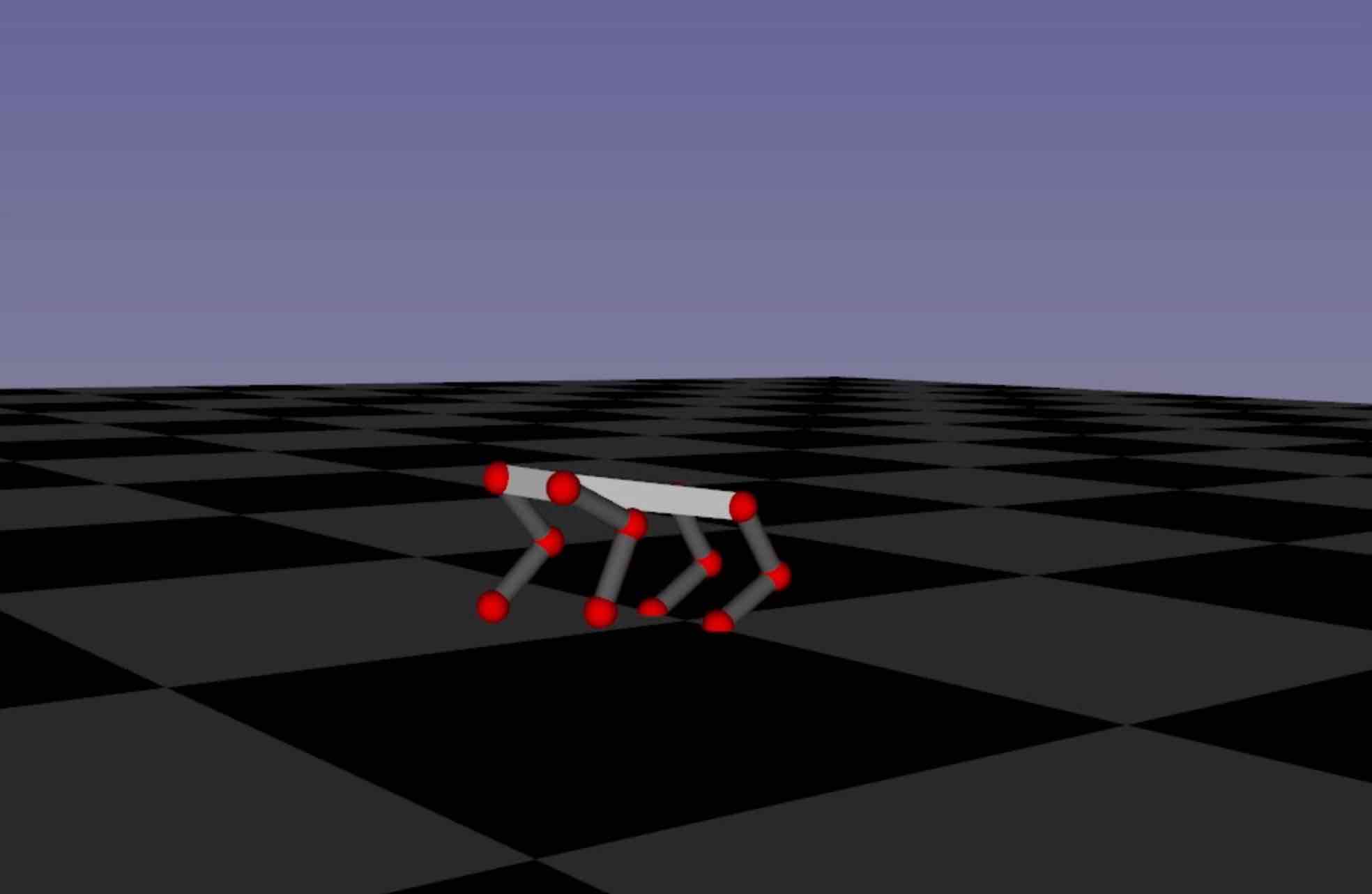}%
		\includegraphics[width=0.20\linewidth, trim={5cm 3cm 12cm 5cm}, clip]{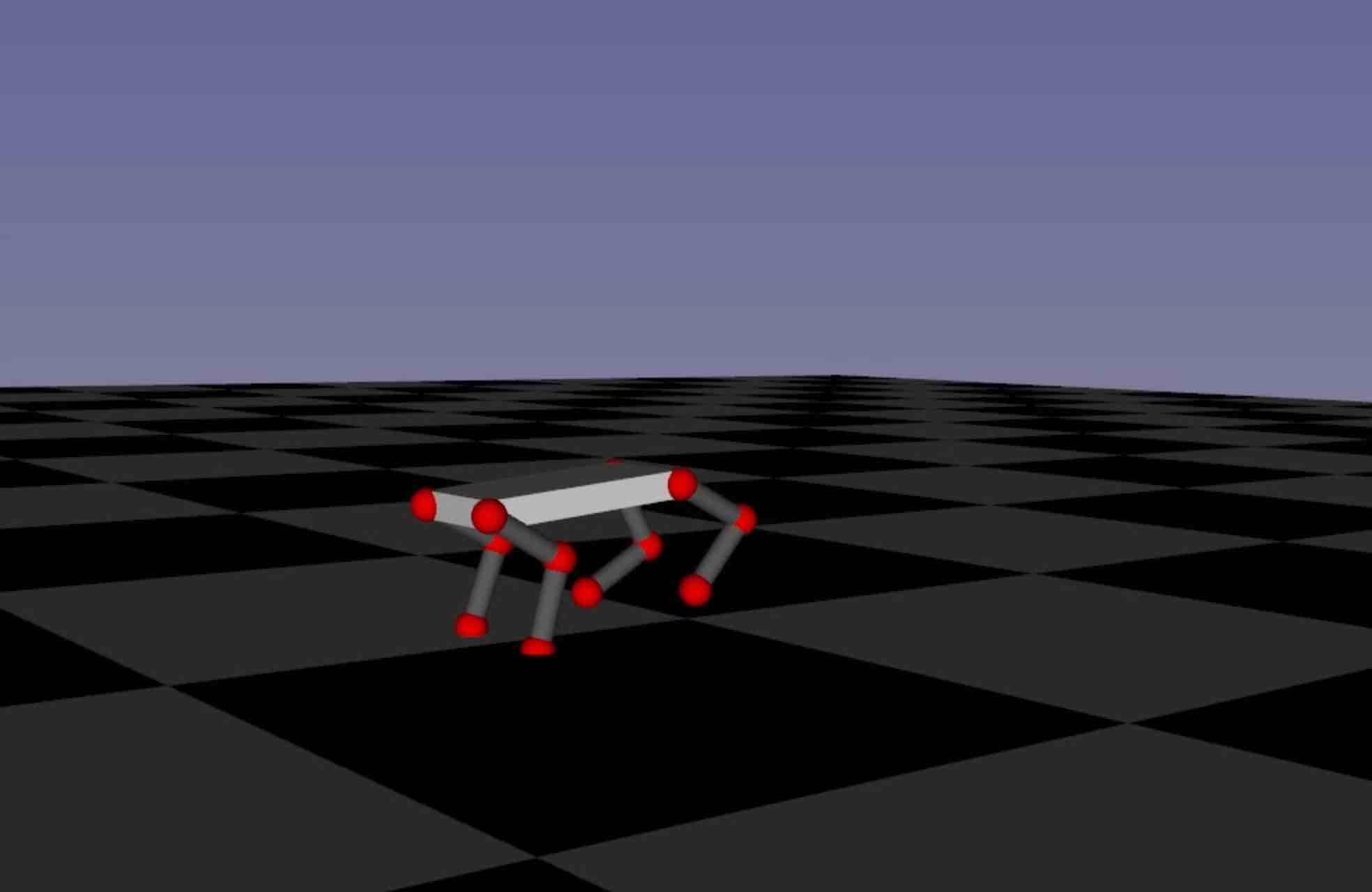}%
		\includegraphics[width=0.20\linewidth, trim={5cm 3cm 12cm 5cm}, clip]{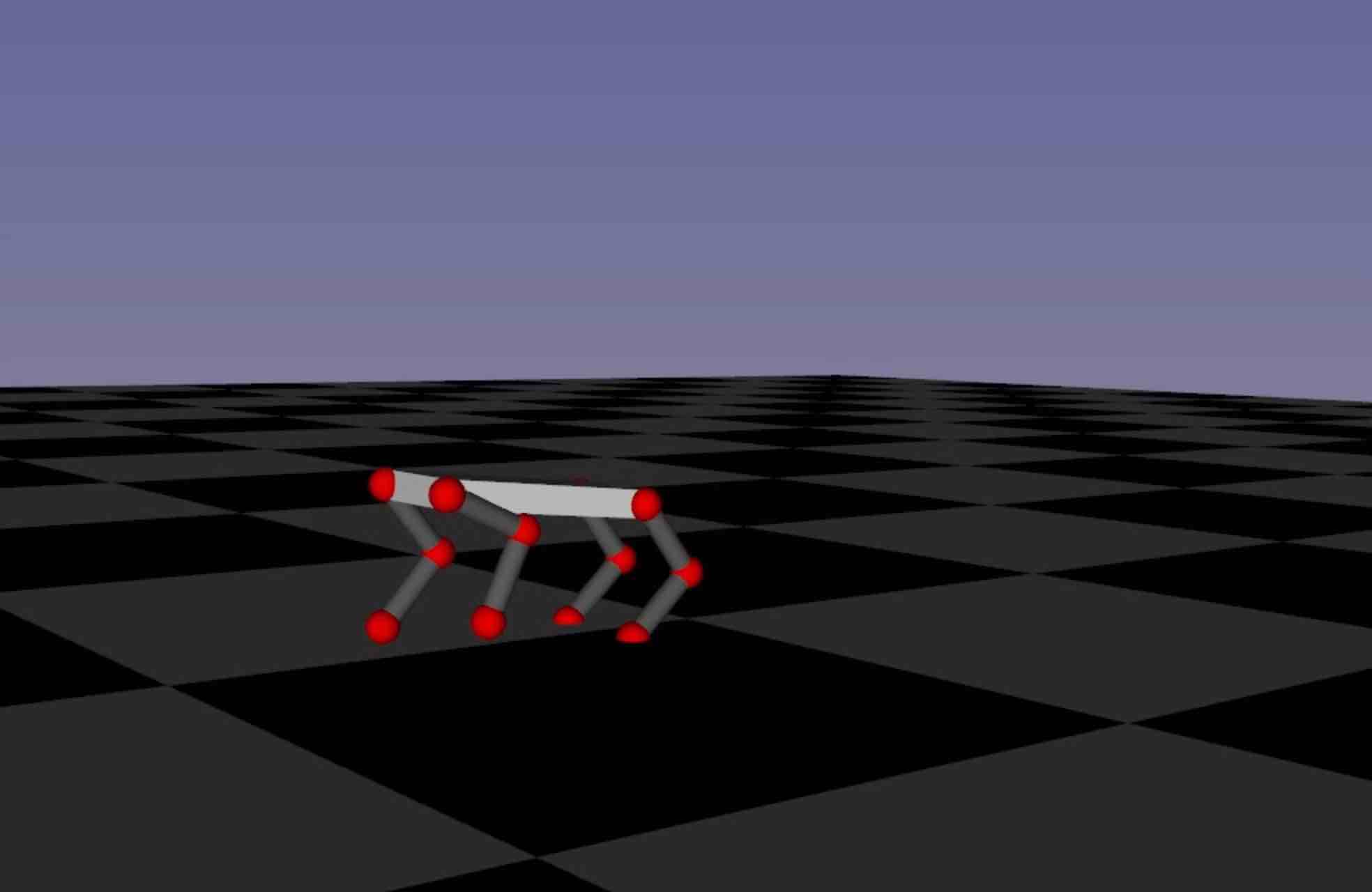}%
		\includegraphics[width=0.20\linewidth, trim={5cm 3cm 12cm 5cm}, clip]{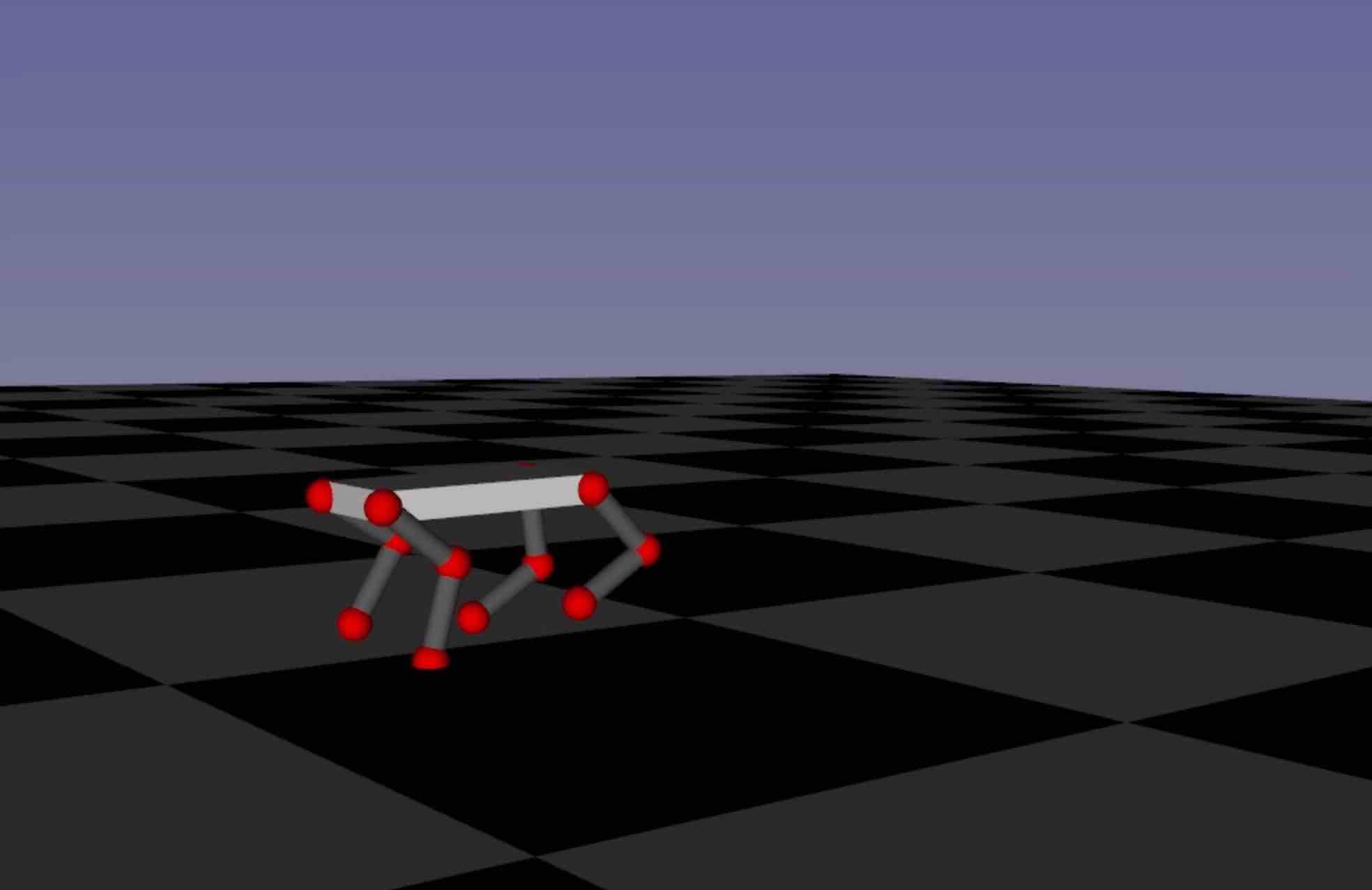}%
	}\\
	\vspace{0.2cm}
	\includegraphics[width=0.49\textwidth]{\filename{figures/gallop/GallopTracking}}
	\caption[]{\small Kino-dynamic results for the optimization of a galloping motion that includes simultaneous flight phases for all endeffectors.}
	\label{fig:motion_with_flight_phases}
\end{figure}

Finally, we present results on a simulated quadruped robot, where we show in Fig. \ref{fig:motion_with_flight_phases} the kino-dynamic trajectories of a galloping motion, very difficult to optimize due to the presence of simultaneous flight phases for all endeffectors, where only gravity is acting on the system. Despite this challenge, kino-dynamic trajectories converge qualitatively well thanks to the exploitation of optimal timing for all available endeffector forces.

%================ Galloping motion ===============%
%=================================================%
%
\subsubsection{Execution of movement plans} \label{exp:motion_execution}
{In this section, we show that optimal motion plans optimized in the previous section using a kino-dynamic approach can be executed in a physical simulator using the architecture described in Fig. \ref{fig:ExecutionArchitecture}}.

\begin{figure}
	\centering
	\drawvideo{5}{80}{%
		\includegraphics[width=0.20\linewidth, trim={18cm 1cm 18cm 4cm}, clip]{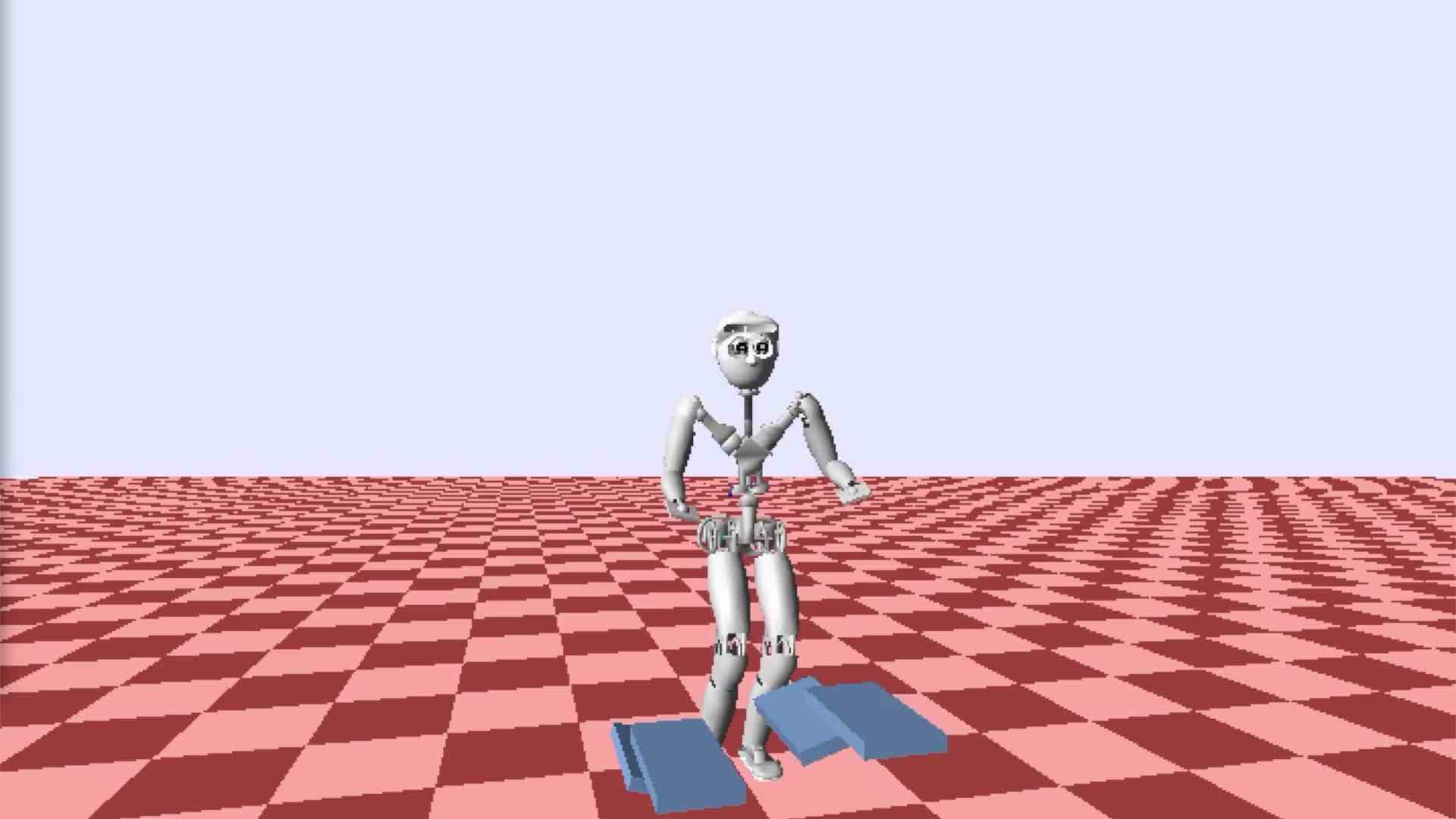}%
		\includegraphics[width=0.20\linewidth, trim={18cm 1cm 18cm 4cm}, clip]{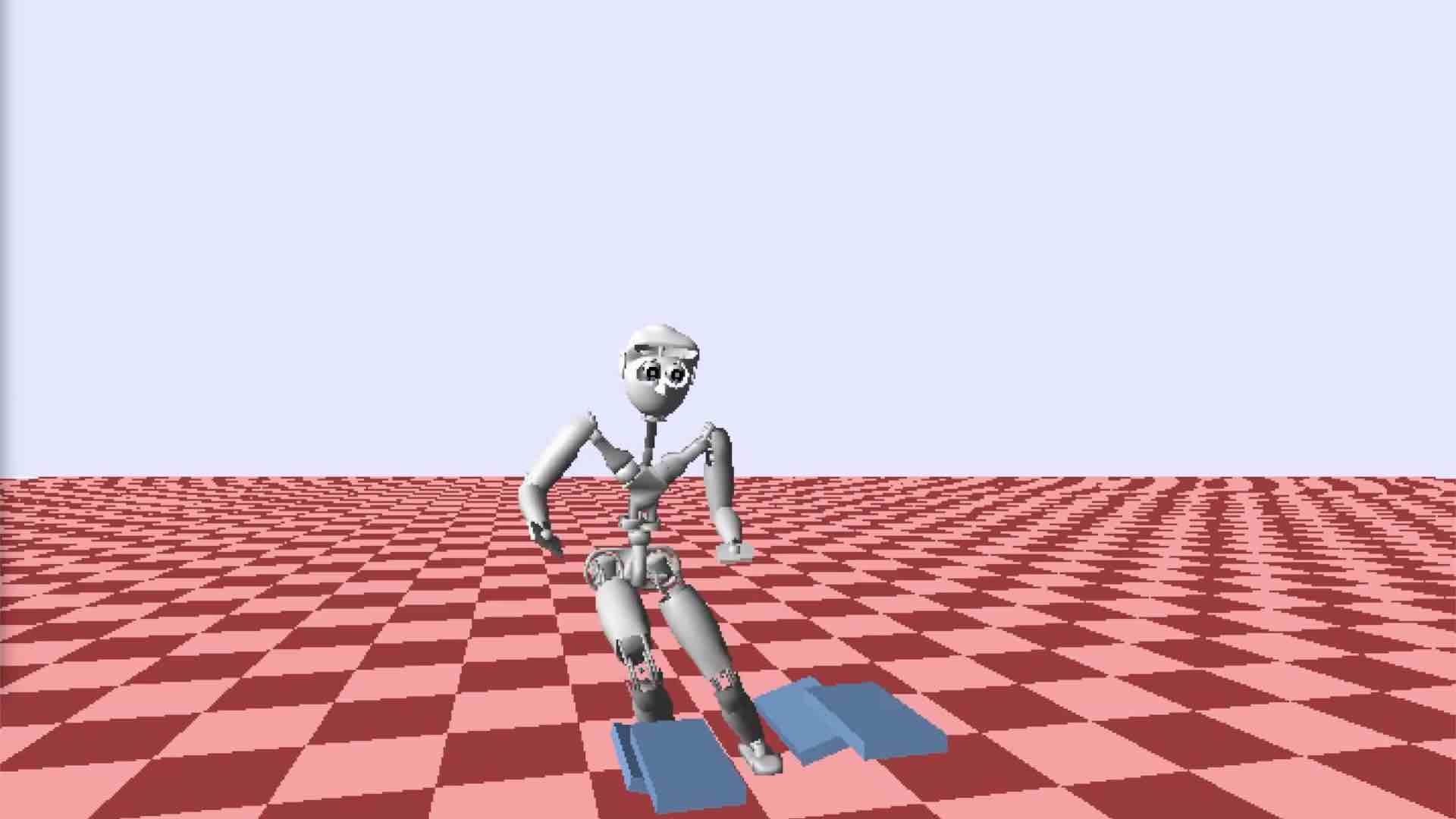}%
		\includegraphics[width=0.20\linewidth, trim={18cm 1cm 18cm 4cm}, clip]{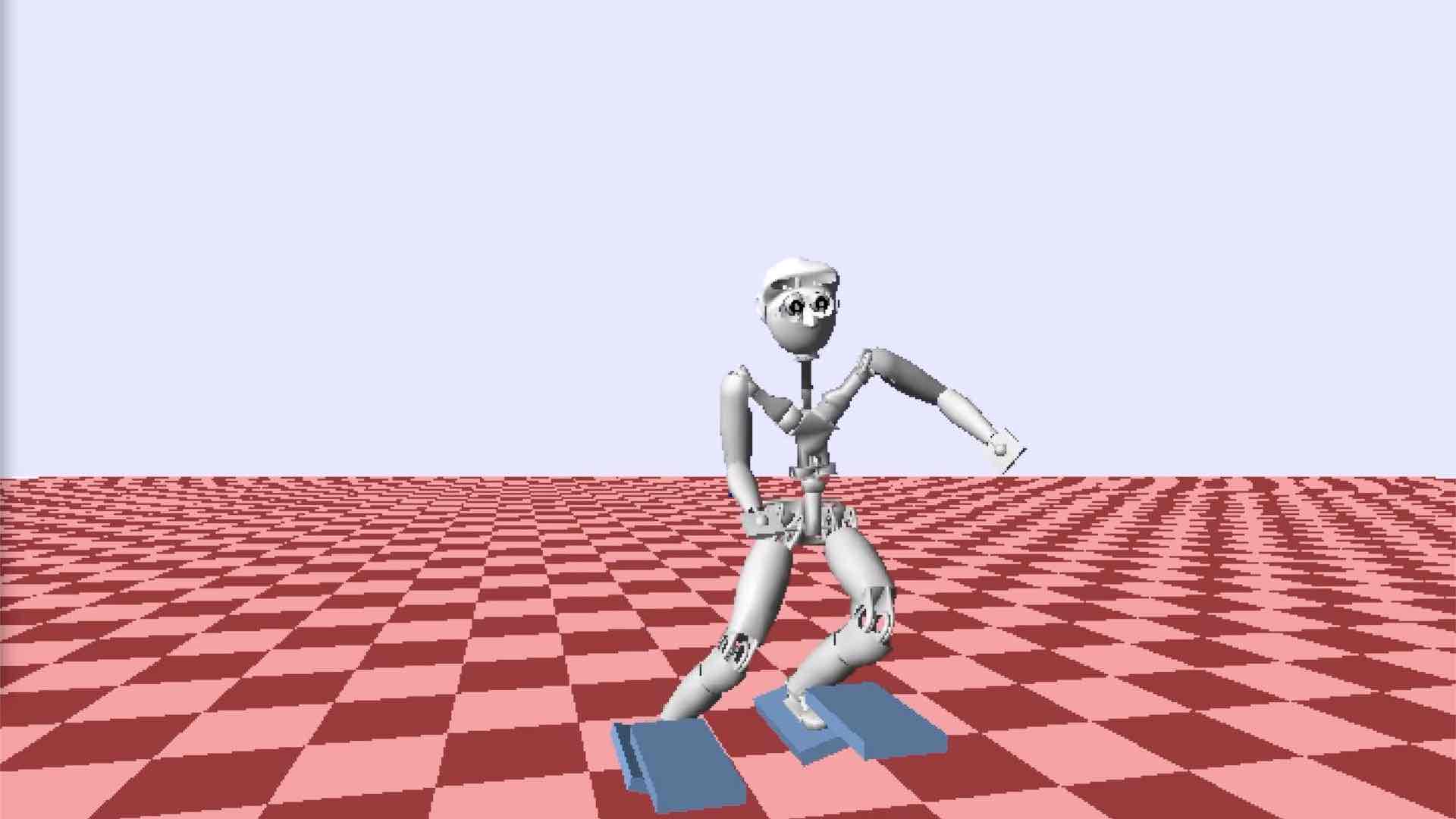}%
		\includegraphics[width=0.20\linewidth, trim={18cm 1cm 18cm 4cm}, clip]{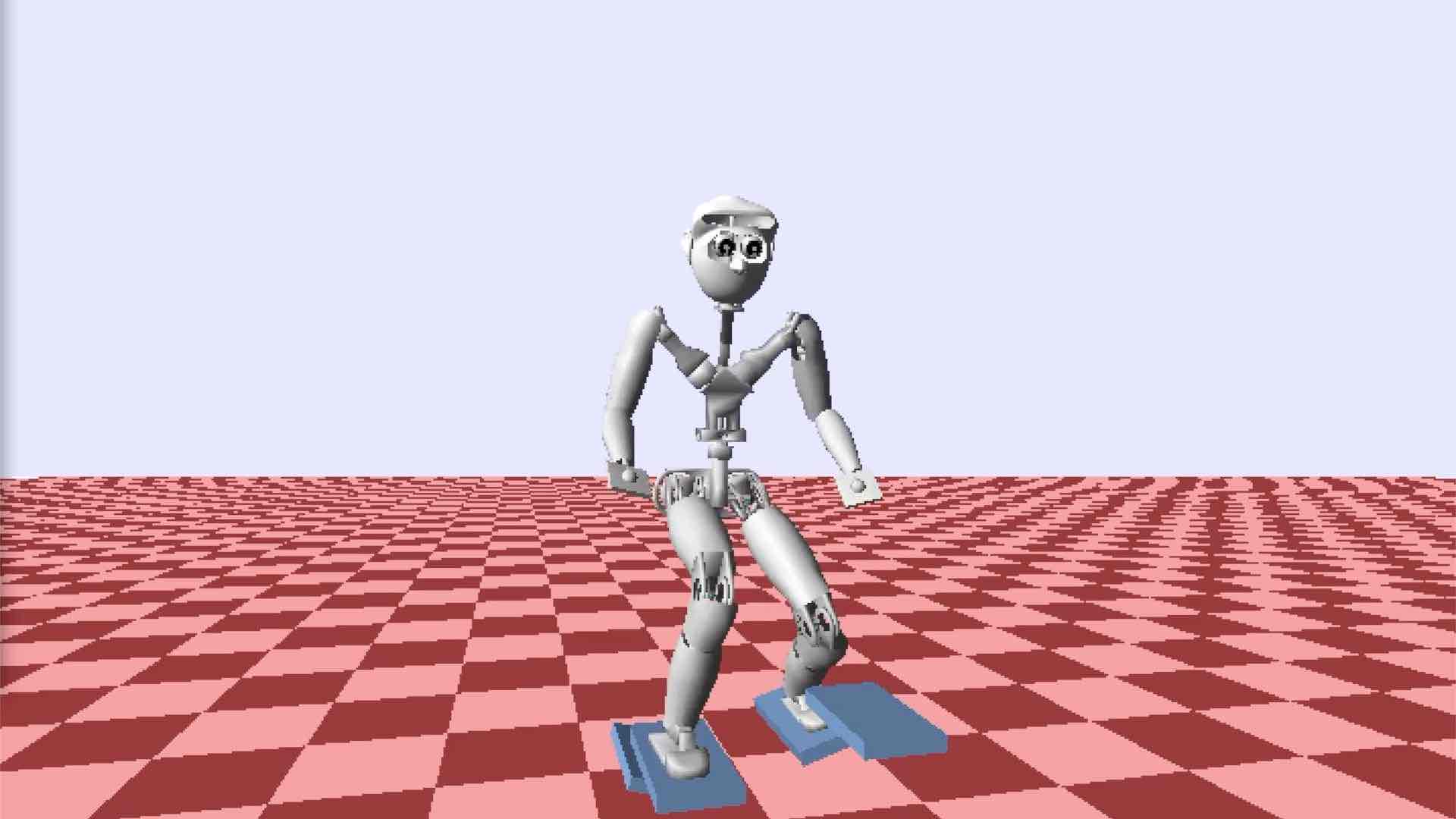}%
		\includegraphics[width=0.20\linewidth, trim={18cm 1cm 18cm 4cm}, clip]{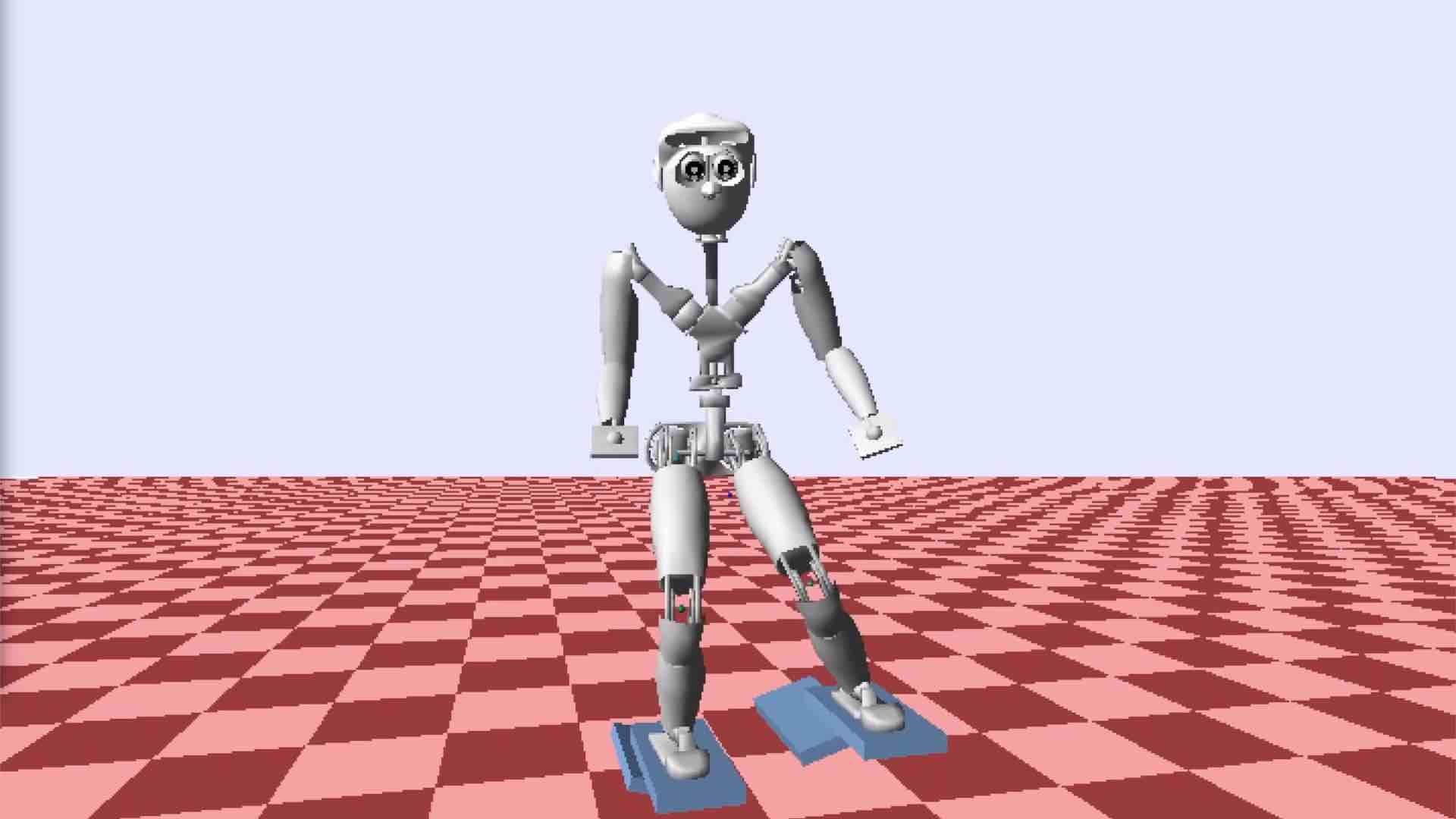}%
	}\\
	\vspace{0.2cm}
	\includegraphics[width=0.49\textwidth]{\filename{figures/walking/WalkingExecution}}
	\caption[]{\small Tracking of desired momentum trajectories for the climbing up stairs motion (shown in Fig. \ref{fig:motion04}) using time optimization.}
	\label{fig:tracking_stepping_motion}
\end{figure}

In Fig. \ref{fig:tracking_stepping_motion}, we first show tracking of an optimized movement plan for a robot climbing uneven stairs using inverse dynamics controllers \cite{AlexAuroPaper} that realize closed-loop behaviors based on risk-sensitive feedback design \cite{FarbodRiskSensitive} that explicitly considers process and measurement noise \cite{MeasurementUncertainty} to compute time-varying feedback gains. In our experience, such a controller leads to overall lower impedance gains in comparison to typical LQR design, which is beneficial to increase compliance at contact with an environment that differs from the ideal scenario used for planning. Note that such a feedback controller is important in this case, as the kino-dynamic optimizer is not used in a receding horizon fashion. The top three plots show the optimized momentum trajectories ($\indexed[dyn]{\lmoms}, \indexed[dyn]{\amoms}$ in blue) as well as their tracking ($\indexed[exe]{\lmoms}, \indexed[exe]{\amoms}$ in red). At the bottom left corner, endeffectors activation over time $\setacteff$ are shown, as given by the optimal timings $\indexed[*]{\timeopt}[][\tid]$ at the bottom right corner.

%========== Stepping tracking execution ==========%
%=================================================%

{In Fig. \ref{fig:torques_limits}, we show  that} actuation limits are not always satisfied, if they are not explicitly considered.
{For instance, on the left column, we analyze torques} in the climbing up stairs motion (Fig. \ref{fig:motion04}). Here, the knee flexion-extension (KFE) joint torque exceeds its limit by 30 Nm (bottom-left in blue). {To enforce torque limits, the solution of the kinematics problem ($\indexed[*]{\robotpos}, \indexed[*]{\robotvel}, \indexed[*]{\robotacc}$) is used to build a linear approximation of Eq. \eqref{eq_actuated_part} along the motion trajectory \ctxt{(used to build the constraint of Eq. \eqref{eq_dynopt_joint_torques}).}
This constraint relates contact wrenches $\indexed{\wrench}[][\effid](\tid) = \begin{bmatrix} \indexed{\efftrq}[][\effid,\tid] & \indexed{\forcevars}[][\effid,\tid] \end{bmatrix} \forall \effid \in \setacteff$ and torques $\jnttrq(\tid)$, making it possible to adapt contact wrenches to satisfy torque limits. The top three left plots show how the right foot's wrench can be adapted from a motion that does not satisfy torque limits (NoTrqLimPlan in blue) to one that does (TrqLimPlan in orange). Further, in green the torque limit being satisfied during execution is shown.} Another way to satisfy torque limits is by redistribution of contact forces among the available endefectors (Fig. \ref{fig:torques_limits} right). In this case, timesteps were kept constant, and the optimizer distributed contact forces in such a way that the left leg is supported by the left hand in order to synthesize a motion within the leg actuation limits. {Joint torques plotted correspond to those degrees of freedom of left limbs that control the endeffector position. \ctxt{Furthermore,} Fig. \ref{fig:movement_generation_statistics} shows the effect of torque limits on solve time performance.}
\ctxt{
\begin{remark}
While we only demonstrated the ability of our approach to include joint actuation limits in the dynamic optimization problem, it would also be straightforward to add such limits in the kinematic optimization problem. Indeed, it would be possible to add linear joint acceleration constraints using Eq. \eqref{eq_actuated_part} and the solution of the dynamic optimization problem to approximate the contact forces.
\end{remark}
}

%========== Torque limits optimization ===========%
\begin{figure}
	\centering
	\includegraphics[width=0.49\textwidth]{\filename{figures/trqlimit/TorqueLimits}}
	\caption[]{\small Satisfaction of actuation limits: \textit{To the left} (walking motion Fig. \ref{fig:motion04}), it can be achieved by adapting endeffector wrenches or timings; \textit{To the right} (multi-contact fixed-time motion Fig. \ref{fig:motion06}), it can happen by redistribution of contact forces among active endeffectors.}
	\label{fig:torques_limits}
	\vspace{0.2cm}
	\includegraphics[width=0.49\textwidth]{\filename{figures/timetrqlimit/TrqOptTiming}}
	\caption[]{\small Effect of actuation limit constraints on solving time of fixed-time optimization problems for different number of discretization timesteps. Results shown correspond to a walking down and up motion (Fig. \ref{fig:motion02}) using soft-constraints for torque limits.}
	\label{fig:movement_generation_statistics}
\end{figure}
%=================================================%

\begin{figure}
	\centering
	\includegraphics[width=0.49\textwidth]{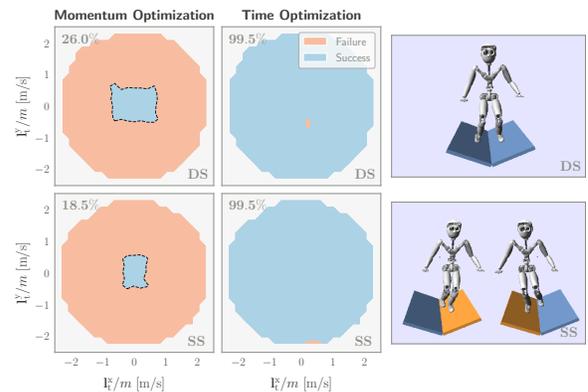}
	\caption[]{\small Comparison of the regions where a dynamically feasible solution is attainable for single and double support experiments using fixed and optimal timings.}
	\label{fig:capture_regions}
\end{figure}

Finally, Fig. \ref{fig:capture_regions} compares the ability of the algorithm at synthesizing a dynamically feasible solution under different initial and final conditions. Initial conditions include varying {CoM} velocities in the horizontal plane and distinct contact supports (one or two feet), while the final condition is a contact configuration as the initial one (single or double support). A solution is colored in orange if after one step, a motion trajectory with momentum values under a small threshold has not been found. The experiment suggests that optimal timings can significantly extend the regions where a feasible dynamical solution is attainable, under given physical conditions and objective function, as well as that timing adaptation is important beyond known results for flat ground walking \cite{majidJournal}.

%=========== Single and Double Support ===========%
%=================================================%

\subsection{On the optimization of contact plans} \label{exp:contacts_planning}
This subsection discusses results on the {surface selection and} contacts planning algorithm using a mixed-integer program that makes use of a dynamics model to measure the quality of the motion induced by the selected contacts plan.

\begin{figure}
	\centering
	\includegraphics[width=0.49\textwidth]{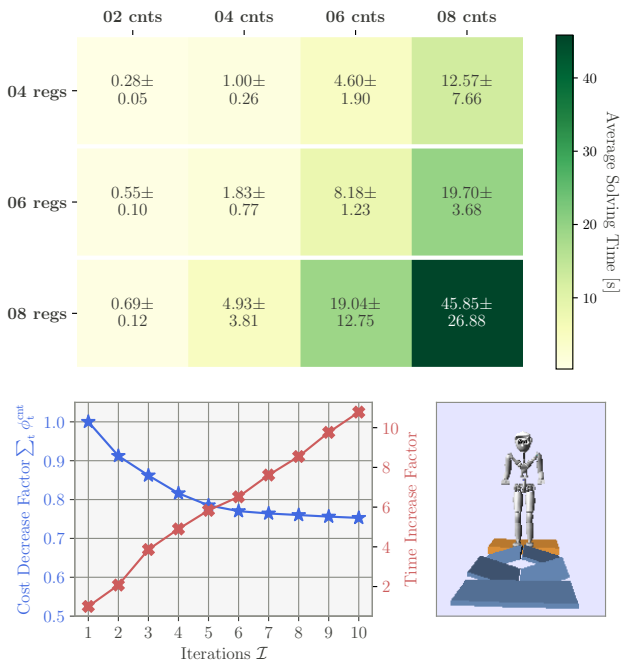}
	\caption[]{\small Statistics about solving time for a contacts planning problem under different number of stepping regions and horizon of the number of contacts. It further compares the cost improvement and increment in solving time for different number of iterations $\numiters$.}
	\label{fig:contacts_planner}
\end{figure}

Figure \ref{fig:contacts_planner} shows a schematic of the experiment setup, average timing results, and a comparison between cost decrease and solving time increase for each {iteration} of the problem internally solved. {In} the experiment a robot {traverses} an uneven terrain from the initial stepping stone (in orange) to a desired position forward using a desired number of contacts {$\numcntsopt$}. Further, the number of the terrain stepping stones is adapted as shown in the statistics table on the number of regions axis. On the figure's top, mean and one standard deviation of solving times are shown for several configurations of surfaces and number of contacts to optimize. Note how short contact plans can be quickly {solved}, while longer ones require more computational effort. {In} those cases, more efficient techniques for contact planning can be used \cite{Tonneau:2018dm}. For example, a predictive neural network could be used to speed up the evaluation of dynamically feasible contact sequences as in \cite{yuchi}.

%============= ContactOpt Statistics =============%
%=================================================%

%============= Quadruped Experiments 1 =============%
\begin{figure*}
	\centering
	\includegraphics[height=0.09\linewidth, trim={0cm 0cm 0cm 0cm}, clip]{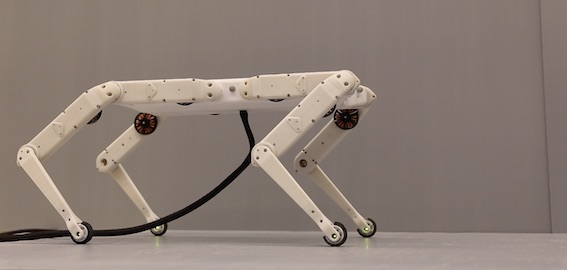}%
	\includegraphics[height=0.09\linewidth, trim={0cm 0cm 0cm 0cm}, clip]{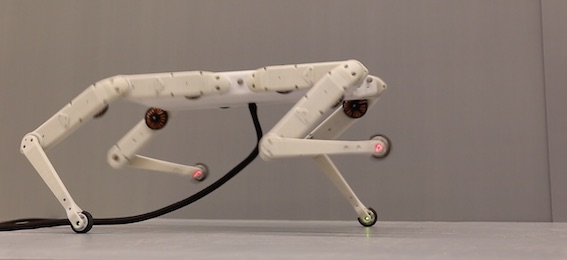}%
	\includegraphics[height=0.09\linewidth, trim={0cm 0cm 0cm 0cm}, clip]{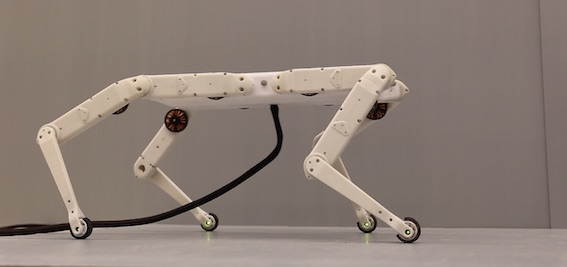}%
	\includegraphics[height=0.09\linewidth, trim={0cm 0cm 0cm 0cm}, clip]{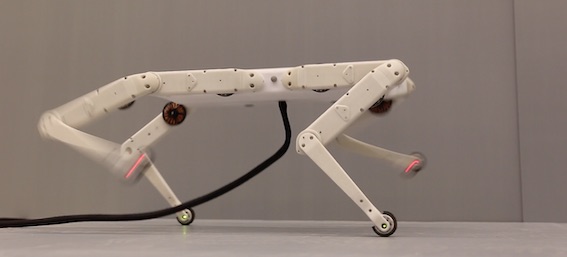}%
	\includegraphics[height=0.09\linewidth, trim={0cm 0cm 0cm 0cm}, clip]{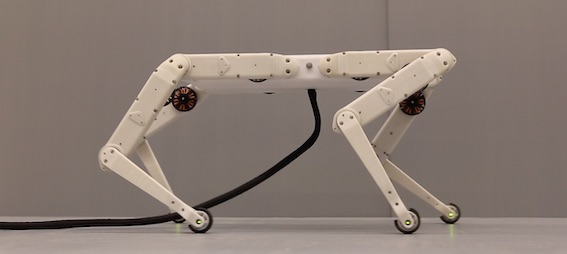}%
	\vspace{0.1cm}
	
	\includegraphics[height=0.09\linewidth, trim={0cm 0cm  0cm 0cm}, clip]{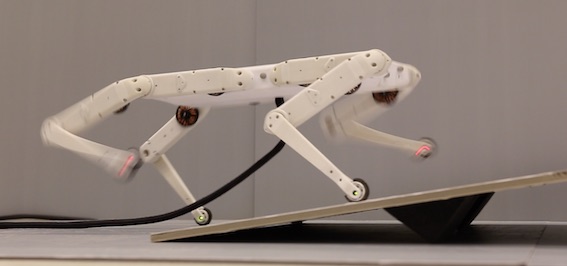}%
	\includegraphics[height=0.09\linewidth, trim={0cm 0cm  0cm 0cm}, clip]{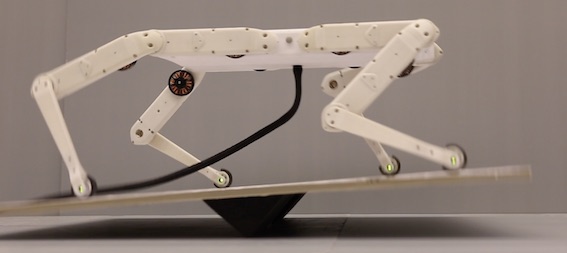}%
	\includegraphics[height=0.09\linewidth, trim={0cm 0cm .5cm 0cm}, clip]{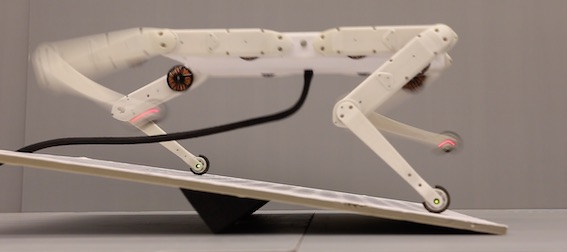}%
	\includegraphics[height=0.09\linewidth, trim={0cm 0cm  0cm 0cm}, clip]{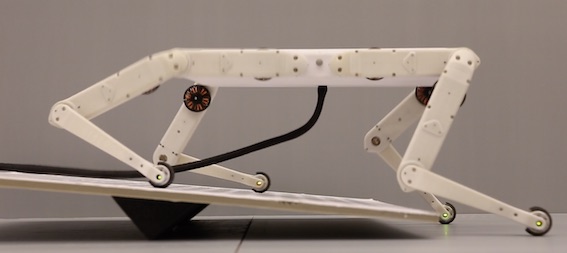}%
	\includegraphics[height=0.09\linewidth, trim={0cm 0cm .2cm 0cm}, clip]{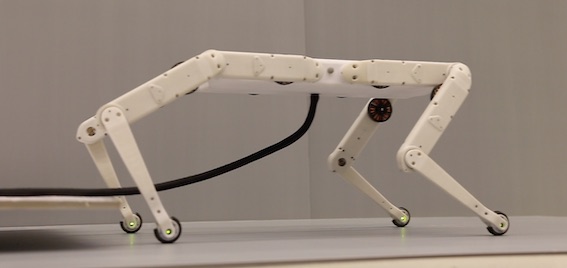}%
	\vspace{0.1cm}
	\caption[]{\small Snapshots of the experiments in scenario 1; top) trot on flat surface, bottom) trot on seesaw }
	\label{fig:experiments_1}
\end{figure*}
%=================================================%

%============= Quadruped Experiments 2 =============%
\begin{figure*}
	\centering
	
	\includegraphics[height=0.092\linewidth, trim={.4cm 0cm 0cm 0cm}, clip]{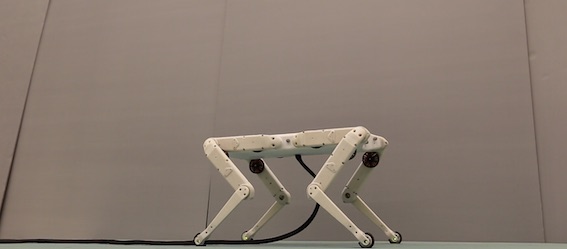}%
	\includegraphics[height=0.092\linewidth, trim={ 0cm 0cm 0cm 0cm}, clip]{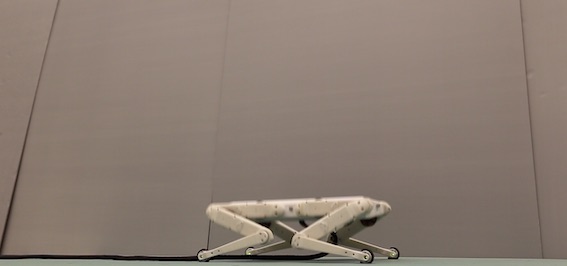}%
	\includegraphics[height=0.092\linewidth, trim={ 0cm 0cm 0cm 0cm}, clip]{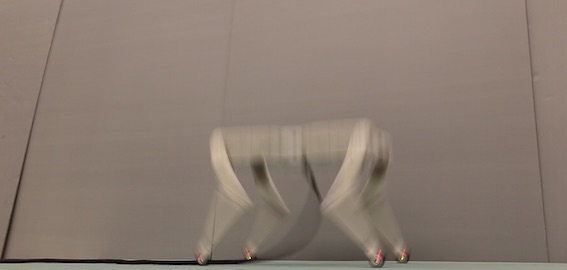}%
	\includegraphics[height=0.092\linewidth, trim={ 0cm 0cm 0cm 0cm}, clip]{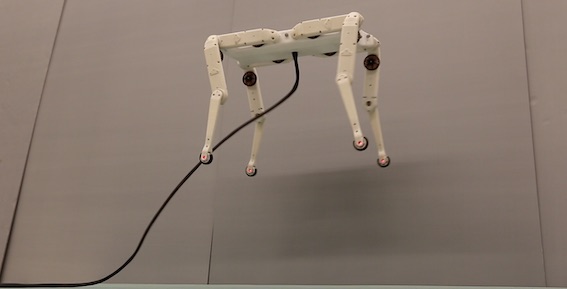}%
	\includegraphics[height=0.092\linewidth, trim={ 0cm 0cm 0cm 0cm}, clip]{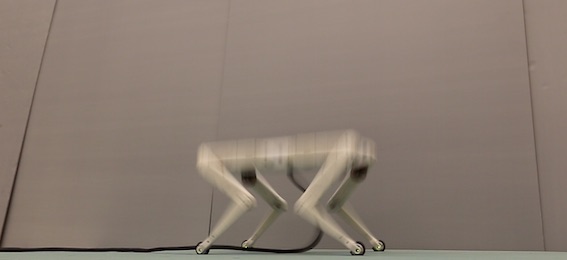}%
	\vspace{0.1cm}
	
	\includegraphics[height=0.092\linewidth, trim={.3cm 0cm 0cm 0cm}, clip]{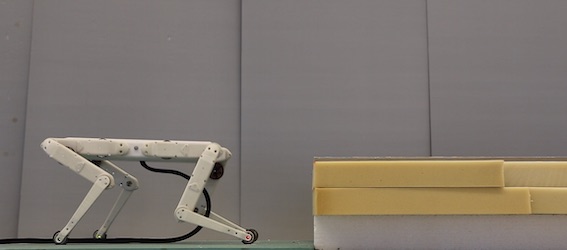}%
	\includegraphics[height=0.092\linewidth, trim={ 0cm 0cm 0cm 0cm}, clip]{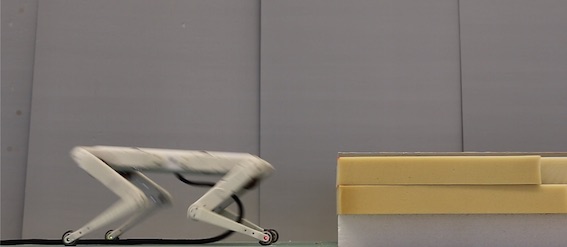}%
	\includegraphics[height=0.092\linewidth, trim={ 0cm 0cm 0cm 0cm}, clip]{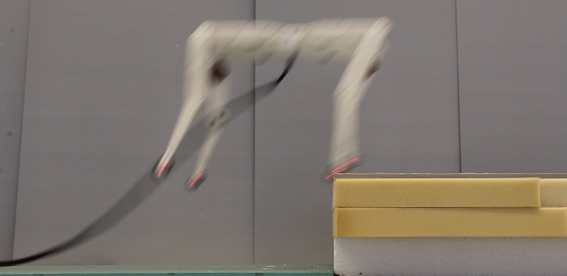}%
	\includegraphics[height=0.092\linewidth, trim={ 0cm 0cm 0cm 0cm}, clip]{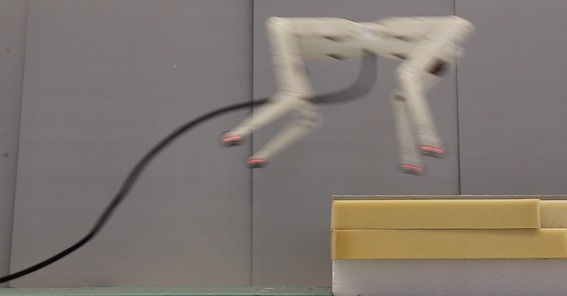}%
	\includegraphics[height=0.092\linewidth, trim={ 0cm 0cm 0cm 0cm}, clip]{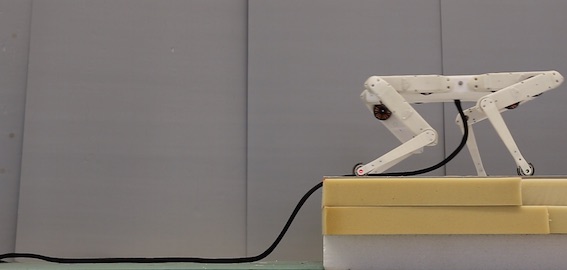}%
	\vspace{0.1cm}
	\caption[]{\small Snapshots of the experiments in scenario 2; top) jump in place, bottom) jump on a box }
	\label{fig:experiments_2}
\end{figure*}
%=================================================%

%============= Quadruped Experiments 3 =============%
\begin{figure*}
	\centering
	
	\includegraphics[height=0.096\linewidth, trim={0cm 0cm 0cm 0cm}, clip]{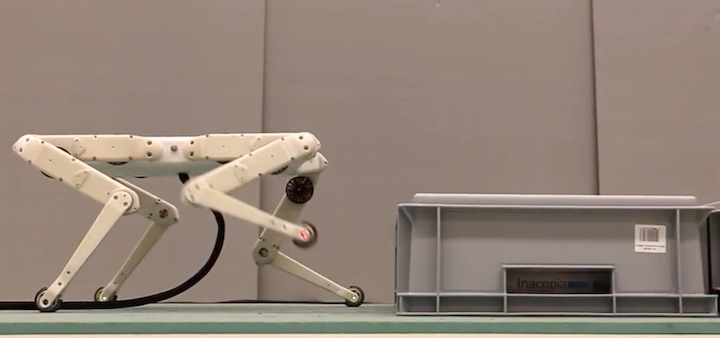}%
	\includegraphics[height=0.096\linewidth, trim={0cm 0cm 0cm 0cm}, clip]{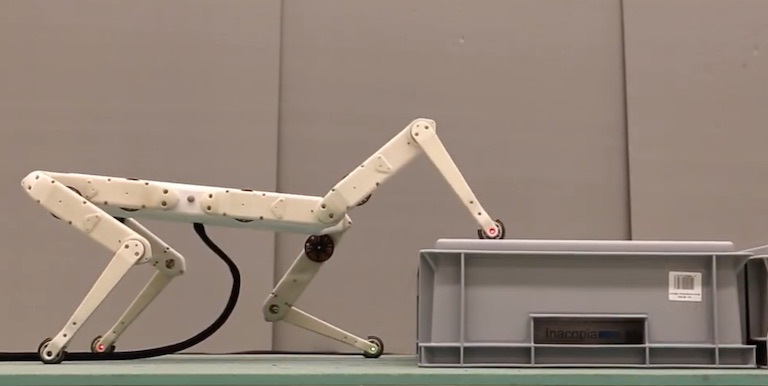}%	
	\includegraphics[height=0.096\linewidth, trim={0cm 0cm 0cm 0cm}, clip]{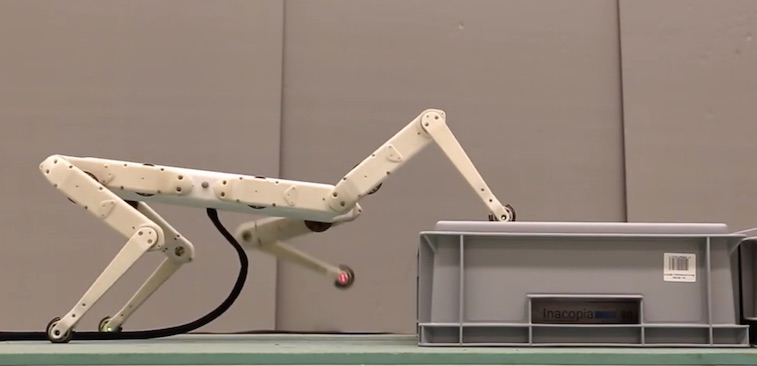}%
	\includegraphics[height=0.096\linewidth, trim={0cm 0cm 0cm 0cm}, clip]{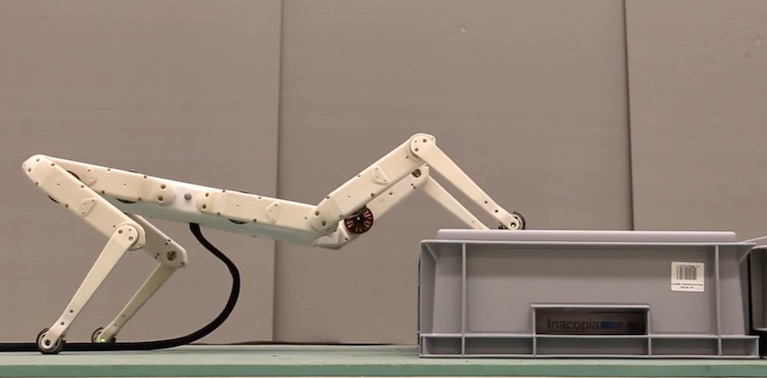}%
	\includegraphics[height=0.096\linewidth, trim={0cm 0cm 0cm 0cm}, clip]{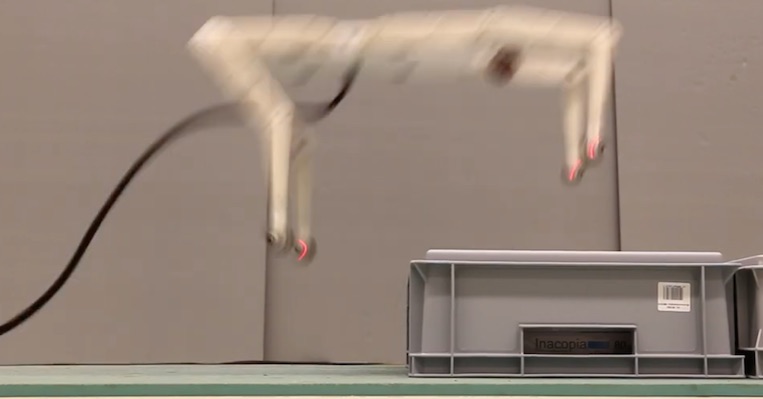}%
	\vspace{0.1cm}
	
	\includegraphics[height=0.095\linewidth, trim={0cm 0cm 0cm 0cm}, clip]{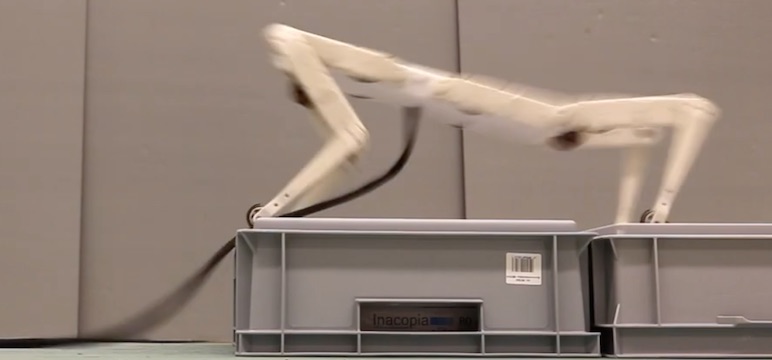}%
	\includegraphics[height=0.095\linewidth, trim={0cm 0cm 0cm 0cm}, clip]{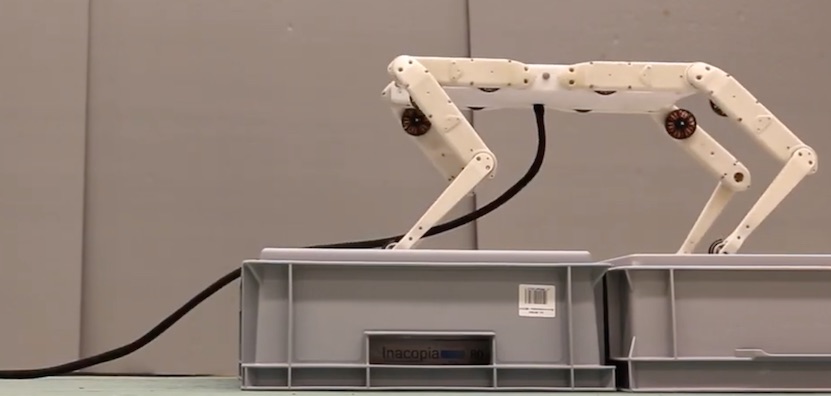}%	
	\includegraphics[height=0.095\linewidth, trim={0cm 0cm 0cm 0cm}, clip]{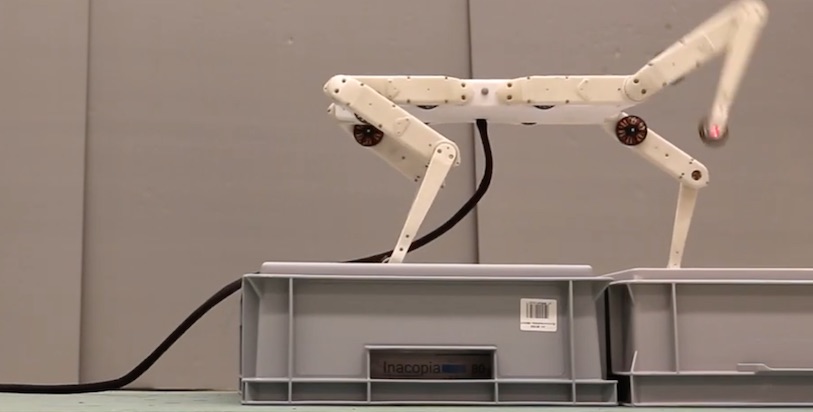}%
	\includegraphics[height=0.096\linewidth, trim={0cm 0cm 0cm 0cm}, clip]{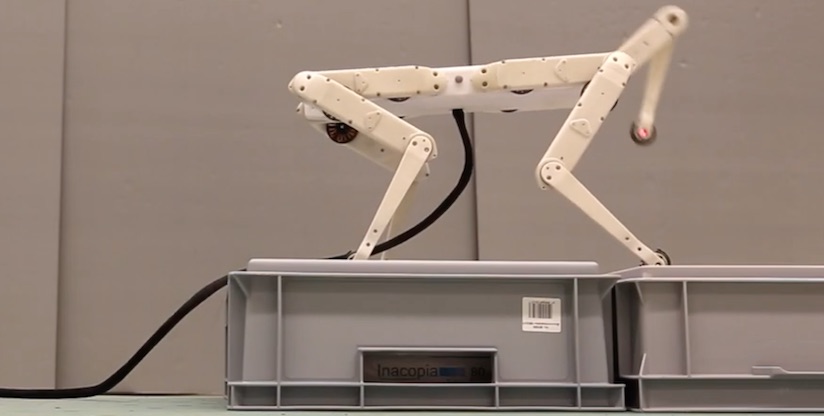}%
	\includegraphics[height=0.096\linewidth, trim={0cm 0cm 0cm 0cm}, clip]{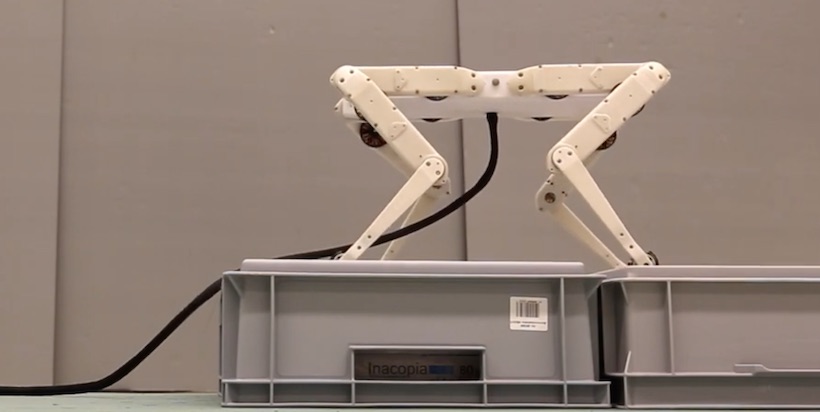}%
	\vspace{0.1cm}
	\caption[]{\small Snapshots of the experiments in scenario 3; step and jump on an obstacle}
	\label{fig:experiments_3}
\end{figure*}
%=================================================%

{Finally, Fig. \ref{fig:contacts_planner} (bottom-left) shows the cost evolution of a contacts optimization problem ($\sum_{\tid}{\indexed{\fcost}[cnt][\tid]}$ in blue) as well as the time required to solved it (in red) as a function of the number of iterations $\numiters$ used to approximate the dynamic constraints. These values have been normalized by the values corresponding to $\numiters=1$, such that both curves depict the cost decrease and time increase factors relative to those that use only one iteration. Notice how initially two additional iterations ($\numiters=3$) reduce the cost by $\sim$15\% while increasing the solving time by a factor 4. Towards the end, however, an additional iteration increases the solving time linearly, but reduces the cost only minimally. This suggests that solving the problem to high-precision optimality (e.g. $\numiters=10$) is impractical because of the large required solving time; however, a sub-optimal solution (e.g. $\numiters=1$) is reasonable and can provide a good initialization contact plan for the motion optimization. The functional form of the cost function $\indexed{\fcost}[cnt][\tid]$ and importance weights are defined similarly to Table \ref{tab:costfunction}.}

\subsection{Real robot experiments} \label{exp:robot_experiment}

This section presents the execution of kino-dynamic motion plans on our quadruped Solo \cite{grimminger2019open}. Our main goal is to demonstrate that these plans are of sufficient quality to be executed on a real robot using only an instantaneous feedback controller and no re-planning.
% we can control the contact forces at the end-effectors by only controlling the motor current \cite{grimminger2019open}. \\
%
We use a passivity based controller to track the optimized motions. The controller tracks desired CoM, angular momentum, base orientation, feet trajectories and also uses the desired feedforward centroidal wrench from the planner. This controller is described in detail in \cite{grimminger2019open}.

We consider {three} different scenarios to show the capability of the planner to generate feasible motions. In the first scenario, we provide the kino-dynamic planner with a periodic sequence of contact points to generate a trotting motion. In the second scenario, we consider a jumping motion with a flight phase. {Finally, in the third scenario, we present a motion that combines a non trivial sequence of contacts and a jumping motion.} In {all} scenarios, we use the approach presented in Section \ref{sec:problem_formulation} and \ref{sec:opt_movement} to generate kino-dynamically feasible motions. Note that for all the experiments we iterated only once between kinematic and dynamic optimizers. Note also that {some of} the motions presented here are the same motions used in \cite{grimminger2019open} to evaluate the control law. We reproduce them here for completeness and focus our analysis on the motion plans not discussed in \cite{grimminger2019open}.

\subsubsection{Scenario 1, trot} \label{exp:trot}

In the first scenario, we give a periodic contact sequence to the planner, where diagonal feet move forward as much as a step length in a specified time (Fig. \ref{fig:experiments_1}, top row). Since the robot does not have the abduction/adduction hip joint, it is very important that the planner generates stable motions taking into account the robot full dynamics and that can be tracked by the controller without step adjustment. In our experiments, we noticed the importance of having fully consistent motion plans (and not solely centroidal dynamic motions), especially during contact transitions. Furthermore, it was also important to have a feedback controller explicitly tracking desired centroidal wrench and feet trajectories. We were able to successfully execute trotting motions at various speeds.
Moreover, in order to test the sensitivity of the motion plans to moderate environmental uncertainty, we planned a flat ground trot and successfully executed it on a seesaw. This result suggests that the optimized motions are neither sensitive to model mismatch nor to small environmental changes. It is particularly interesting to note that we were able to execute rather long motions of around 10 [s] without re-planning.

%Furthermore, since we have a sequence of contact transitions, the role of feedback control to track the centroidal and feet trajectories is crucial. We tested the trotting motion with different velocities successfully on the robot and showed only two of them in the accompanying video. Furthermore, to show the robustness of the motions, we put a seesaw on the path of the robot (Fig. \ref{fig:experiments_1}, bottom row), while no information about it is given to the planner or controller. Interestingly and thanks to the feedback control, the robot is able to tolerate this disturbance and successfully traverse the seesaw.

\subsubsection{Scenario 2, jump} \label{exp:jump}

To show the capability of the planner to generate highly dynamic motions, in this scenario we provide the planner with contact sequences with a flight phase. First, we implemented a jump in place (Fig. \ref{fig:experiments_2}, top row), where the robot only needed to generate vertical thrust. In this scenario the robot was able to jump 65 cm, while the robot height in its natural standing phase is 24 cm. The generated plan is good enough such that the feedback controller is able to track desired linear momentum in the vertical direction and realize the desired jump in place. 
We then implemented a forward jump on an 18 cm box (Fig. \ref{fig:experiments_2}, bottom row). In this case the planner needs to generate linear momentum in both vertical and horizontal directions to jump 60 cm forward and around 30 cm upward at the apex of the flight phase while ensuring that the generated angular momentum at take-off enables landing with the proper orientation.

\subsubsection{{Scenario 3, step and jump on obstacle}} \label{exp:step_jump}

{In this scenario, we present a motion that is a combination of transition between different multi-contact sequences, and a flight phase for jumping on an obstacle (Fig. \ref{fig:experiments_3}). Here, our main goal is to showcase the capability of the planner in generating highly constrained multi-contact motion together with a highly dynamic motion. To step on the obstacle, the planner exploits the high range of motion of the robot hip joint and step on the obstacle without the need to change the base orientation to avoid collision of the front legs with the obstacle. Then, through generating enough thrust on a non-coplanar set of contact points and in a non-trivial end-effectors configuration, the robot jumps on top of the obstacle. Finally, through another multi-contact set of change in contact configuration, it brings back the joint configuration to the default one. This experiment scenario further illustrates the versatility of our optimizer to generate motions in complex environments.}
%%%%%%%%%%%%%%
% Discussion %
%%%%%%%%%%%%%%
\section{Discussion} \label{sec:discussion}
\subsection{Time and computational complexity} \label{exp:time_complexity}
In general, finding a solution to the dense version of any of the convex approximations we solve, requires a polynomial time algorithm (of order ${\pazocal{O}(\numsoc^{\frac{1}{2}}[\numsoc+\sizesoc]\sizesoc^{2}) \approx \pazocal{O}(\numsoc^{\frac{3}{2}})}$, $\numsoc$ being the number of quadratic constraints and $\sizesoc$ its size) \cite{InteriorPointPTM}. However, within the problem size ranges of interest to us and thanks to the exploited problem sparsity patterns (e.g. due to time indexing), we observe (Fig. \ref{fig:conv_per_num_timesteps}) that the problem has approximately linear time complexity. It is possible to note this linear tendency for both momentum and time optimization problems, despite their different rates of \ctxt{growth} due to distinct problem sizes and even problems that consider actuation limits show this linear tendency (Fig. \ref{fig:movement_generation_statistics}).

When considering torque limits the doubled computational effort due to the addition of ${2 \numjnts \dishorizon}$ inequality constraints for a problem with ${\dishorizon}$ timesteps and robot with ${\numjnts}$ joints ($\approx 32$ in our case) can be reduced by considering only the weakest joints or only those involved in the motion. All in all, computation times are still lower than the planned horizon, making it possible to run the algorithm online (for example the next plan can be computed, while the current one is being executed).
\subsection{On limitations and comparison of the approximations} \label{exp:approximation_limitations}
Problem \eqref{dynopt_problem} is nonconvex and thus hard to solve. The proposed heuristics lighten to some extent the effort required to find a solution by searching for an approximate one within the convex space of the problem. This however comes with certain limitations. For instance, when using trust-regions, they might be inappropriately built leading to non-optimal solutions, or even unsuitably initialized which could render the interior of the convex cone empty leading to primal infeasibility. For the soft-constraint method, the difficulty lies in finding an appropriate trade-off between two competing objectives: amount of constraint violation and problem conditioning. An adaptive solution that iteratively reduces the value of the allowed amount of constraint violation $\trpenalty$  works well for the trust-region heuristic, {though care is required to slowly converge from the relaxed to the approximate problem without rendering the problem infeasible due to excessive reduction of $\trpenalty$}. {For the soft-constraint method, a value high enough to prioritize the soft-constraint over the rest of the cost terms works well.}

{We have used both methods to synthesize a relatively high number of motions, so as to be able to successfully train a neural network \cite{yuchi}. From this experience, we highlight that both methods work equally well. However, we would like to remark two cases where on would be more appropriate than the other. First case would be when a certificate of optimality or infeasibility matters, e.g. to compute a viable set to be used as a terminal set constraint. In this case, the trust-region method is more appropriate as the slack or degree of constraint violation is controlled using a primal constraint and the certificate is valid for the given precision. The second case would be when the solver is to be warm-started not from information from previous iterations, but using a predictive model (e.g. a neural network). In this case, the soft-constraint method would not run into the risk of infeasibility due to an invalid initialization, making it a more appropriate approach to handle this case.}

{Notice that a single timeline was used to parameterize and optimize motions in eq. \eqref{dynopt_problem}. However, this might be a limitation for more general and complex motions that require an independent timeline for each endeffector. Finally, notice that while the method is very general in nature and works well to solve problem \eqref{dynopt_problem}, it is the case, as with any other nonlinear optimization method such as sequential quadratic programming \cite{SnoptPaper}, that it might not be appropriate or fail with other problem instances. }
\subsection{\texorpdfstring{{Stability of the computed motions}}{Stability analysis}} \label{exp:stability}
{
Our method generates dynamically feasible motions that satisfy general contact stability criteria such as \cite{AdiosZMP}.
If the final position of the robot has zero velocity, then we are guaranteed that the motion (if perfectly executed) will lead to a stable behavior, i.e.
a behavior that will lead to the robot to stop and remain stabilized.
Additionally, the construction of the feedback controllers ensures that
the controlled motion will be locally stable, i.e. it will reject small perturbations. While
we do not have any guarantees on the size of the region of stability, our
experimental evaluations demonstrate that the motions are good enough to
be executed in a simulator or on a real robot with substantially different dynamics. 
We noticed in our real-robot experiments that the synergy between the feedback controller and the motion plan is important and that none of them is solely responsible for a success execution of the motions, especially when executing a 10s long multi-contact motion.

% cannot be attributed solely to none of them, but to the synergy between planner and controller, which are capable of generating accurate kino-dynamic motions and of robustly following them, despite the length duration of around 10[s].

Ideally, it would be desirable to use the optimizer in a receding horizon
manner, raising the issue of closed-loop stability of the optimizer.
Several methods have been proposed to ensure stability of model predictive control problems such as the use of a terminal equality constraint \cite{ocp_terminal_constraint}, terminal cost \cite{ocp_terminal_cost}, terminal constraint set \cite{ocp_terminal_constraint_set} or terminal cost and constraint set \cite{ocp_terminal_cost_and_constraint_set}. 
% Further, \cite{ocp_stability} showed that all methods ensure stability, as they satisfy four essential principles. 
In this work, we use a terminal cost that keeps the terminal state within a viable set to generate balanced motions (see table \ref{tab:costfunction}). This should thus lead to closed-loop stable behaviors.

Moreover, our approach exploits sequential convex approximations (cf. section \ref{sec:opt_movement}) to achieve polynomial-time convergence and provide a certificate of optimality or infeasibility for the motion to the desired precision. We highlight that these features do not come for free in any off-the-shelf solver. 
For instance, an off-the-shelf interior point method 
for general nonlinear problems will not take advantage of the
structure of the problem as we do. This will result in a poor approximation
of the non-convex constraints unable to capture the global convex part of the problem, thus leading to slower convergence.
Lastly, the certificate of optimality certifies that problem constraints are satisfied to the desired precision.}
\subsection{Cost definition and importance weights} \label{exp:cost_definition}
As pointed out throughout this work, efficiency is a key concern. Consequently, the cost function (used to synthesize motions) is composed using convex quadratic expressions, as shown in Table \ref{tab:costfunction}. The set of importance weights for these costs is, however, expected as an input (see Fig. \ref{fig:ExecutionArchitecture}), as it gives the user the flexibility to shape solutions using the knowledge about the particular robot and application. For instance, it allows to express different preferences of endeffector force distributions in humanoid and quadruped robots. Similarly, a preference for highly dynamic and aggressive motions such as jumping (Fig. \ref{fig:motion_with_flight_phases}) over more conservative and slow motions (Fig. \ref{fig5:movement_planning}) can be expressed by lower penalties over control variables. However, automatically computing appropriate cost weights to
generate desired behaviors remains an open research problem.

\subsection{Comparison to other approaches} \label{exp:approaches_comparison}
%

%In this space, the computational efficiency of the presented approach is regarded in comparison to other methods applied on similar real-robotic experiments, so as to highlight its competitiveness. 
In \cite{JustinMomentumOptimization}, the motion and timings for a walking on stairs using a handrail scenario, given a sequence of contacts, are optimized in less than 5.5s. 
However,
the multiple shooting solver used in this approach is closed-source to the best of our knowledge.
In our approach, such a motion can be optimized in around 4.8s. 
In \cite{DBLP:conf/iros/KoenemannPTTSBM15}, one iteration of a multi-contact motion of 0.5s duration can be optimized within 0.05s. Thus, extrapolating, one iteration for a 7s motion could be optimized within 0.7s. This approach, however, does not take into account hard constraints. In our approach, the cost of such an iteration is around 0.61s. 
In \cite{winkler18}, a bipedal motion of 4.4s is optimized within 4.1s together with the contact sequence but uses a simplified dynamics model, assuming for example a constant locked inertia tensor at the CoM. Our method would achieve a comparable time by optimizing 4 contacts within a time horizon of 5s.
Our contacts planning approach based on mixed-integer programming is competitive only for small problems that optimize a few contacts, due to the combinatorial complexity of mixed-integer programs. For longer contact sequences, other state of the art approaches are more competitive, but typically use simplified dynamics to test for contact transition feasibility \cite{Tonneau:2018dm,fernbach2018croc,lin2016using}.
Note however, that the kino-dynamic optimizer can be used to generate data and
learn how to predict dynamic contact feasibility and significantly speed up
contact search \cite{yuchi}.

These few examples highlight the competitiveness of the presented method while
enabling the resolution of the problem without simplifications.
However, we are not yet capable to compute solutions for model predictive control (e.g. at 50Hz rate or above) and thus we require a feedback controller
to stabilize the motion in between plan computations.
Bringing such approaches to real-time rates while enabling full-body optimization remains an open problem, likely to require the design of
dedicated numerical solvers and smart warm-start procedures.
Lastly, we note that the receding horizon control of whole-body motions ensuring stability, robustness and recursive feasibility, remains an open
and exciting research problem.
%
% They also let us see that it is not as fast as a quadratic program (QP) to be used directly for receding horizon control \cite{majidJournal, DBLP:conf/humanoids/Ramirez-Alpizar16,  conf/humanoids/TedrakeKDM15} and thus requires a feedback controller to correct for deviations from the nominal motion. 
%We note that our real-robot experiments cannot be attributed solely to none of them, but to the synergy between planner and controller, which are capable of generating accurate kino-dynamic motions and of robustly following them, despite the length duration of around 10[s]. In practice, our approach could be further sped up by running it in two phases: one to find an initial motion guess to warm-start the heuristics by for example using information from previous iterations or from predictive models as in \cite{yuchi}, and the second where the important factor is efficient convergence to high precision by monitoring for example sets of constraints and adapting the parameterization of the heuristics as required in order to focus computational power only on the required places. Lastly, we note that the receding horizon control of whole-body motions remains an open problem.}

%%%%%%%%%%%%%%
% Conclusion %
%%%%%%%%%%%%%%
\section{Conclusion} \label{sec:conclusion}
We have presented a structured and efficient algorithm for generating time-optimal motion plans for robots with arms and legs, as well as an approach to select a set of contact surfaces from a terrain description that supports such a motion. Finally, we have shown experimental evidence on a physical simulator and on a real quadruped robot that the algorithm is capable of efficiently generating dynamically feasible motion plans. Future work will include the extension of the algorithm to receding horizon control. The open source repository \cite{opensourcelink} offers fully functional kino-dynamic demos, examples of tasks descriptions and implementation details.

%%%%%%%%%%%%%%
% Bibliography %
%%%%%%%%%%%%%%

\bibliographystyle{ieeetr}
\bibliography{references}

%%%%%%%%%%%%%%
% Biography %
%%%%%%%%%%%%%%
	
 \begin{IEEEbiography}[{\includegraphics[width=1in,height=1.25in,clip,keepaspectratio]{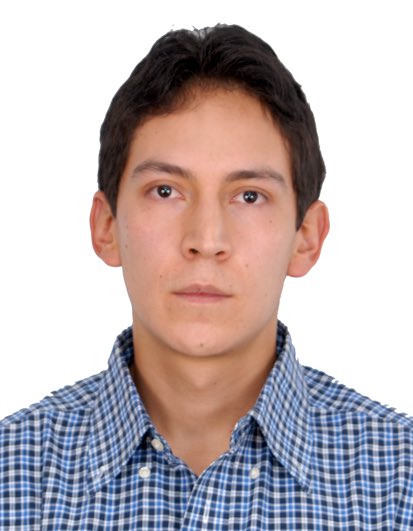}}]{Brahayam Ponton}
 received the B.Sc. degree in electronics and control engineering from the National Polytechnic University (EPN), Quito, Ecuador in 2011, the M.Sc degree in robotics from the Swiss Federal Institute of Technology Z\"urich (ETHZ), Z\"urich, Switzerland, in 2014 and Ph.D. degree in computer science from the Eberhard Karls Universit\"at T\"ubingen, T\"ubingen, Germany, in 2019.
 \end{IEEEbiography}

 \begin{IEEEbiography}[{\includegraphics[width=1in,height=1.25in,clip,keepaspectratio]{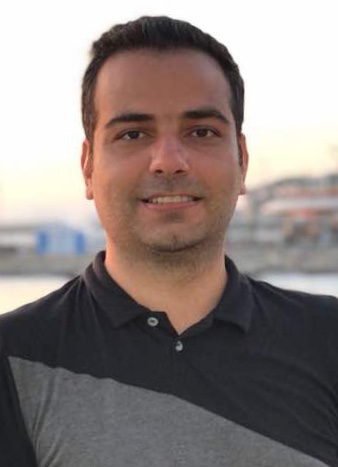}}]{Majid Khadiv}
 received the B.Sc. degree in mechan- ical engineering from the Isfahan University of Tech- nology (IUT), Isfahan, Iran, in 2010, and the M.Sc. and Ph.D. degrees in mechanical engineering from the K.N. Toosi University of Technology, Tehran, Iran, in 2012 and 2017, respectively.\\
 He is a Postdoctoral researcher with the Movement Generation and Control Group, Max-Planck Institute for Intelligent Systems, T\"ubingen, Germany. He joined the Iranian National Humanoid Project, Surena III, and worked as the Head of Dynamics and Control Group from 2012 to 2015. He also spent a one-year visiting scholarship under the supervision of Prof. L. Righetti at the Autonomous Motion Laboratory, Max-Planck Institute for Intelligent Systems. His main research interest is control of legged robots.
 \end{IEEEbiography}

 \begin{IEEEbiography}[{\includegraphics[width=1in,height=1.25in,clip,keepaspectratio]{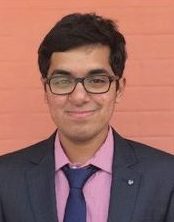}}]{Avadesh Meduri} received his B.E (hons) in Manufacturing Engineering from Birla Institute of Technology and Science Pilani (BITS Pilani), India in 2019. He is currently a PhD student in the Mechanical and Aerospace Engineering Department at Tandon School of Engineering, New York University, USA.\\
 He visited Movement Generation and Control Group at the Max-Planck Institute for Intelligent Systems to pursue his undergraduate thesis under the supervision of Prof. L. Righetti.  His main research interests are contact and motion planning for legged robots.
 \end{IEEEbiography}
	
 \begin{IEEEbiography}[{\includegraphics[width=1in,height=1.25in,clip,keepaspectratio]{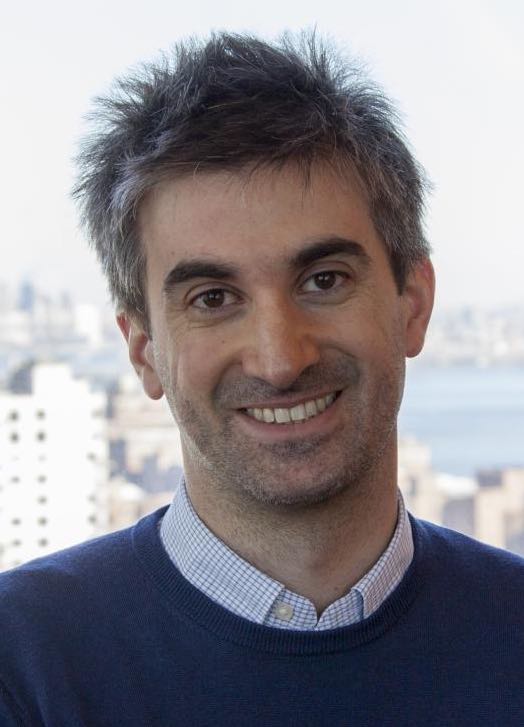}}]{Ludovic Righetti}
 (Senior Member, IEEE) received an engineering diploma in computer science and a doctorate in science from the Ecole Polytechnique Federale de Lausanne, Switzerland, in 2004 and 2008, respectively.\\
 He is an Associate Professor in the Electrical and Computer Engineering Department, the Mechanical and Aerospace Engineering Department and the Center for Urban Science And Progress at the Tandon School of Engineering, New York University. He is also a Senior Researcher at the Max-Planck Institute for Intelligent Systems in Germany. His research focuses on the planning and control of movements for autonomous robots, with a special emphasis on legged locomotion and manipulation.
 \end{IEEEbiography}

\end{document}